\documentclass{article}

\usepackage{iclr2027_conference,times}

\usepackage[utf8]{inputenc}
\usepackage[T1]{fontenc}
\usepackage{hyperref}
\usepackage{url}
\usepackage{xurl}
\usepackage{booktabs}
\usepackage{multirow}
\usepackage{amsmath,amssymb,amsfonts}
\usepackage{microtype}
\usepackage{xcolor}
\usepackage{enumitem}
\usepackage{graphicx}
\usepackage{wrapfig}
\usepackage{xspace}
\usepackage{fontawesome5}

\makeatletter
\renewenvironment{table}
  {\setlength{\abovecaptionskip}{\z@}%
   \setlength{\belowcaptionskip}{10\p@}%
   \@float{table}}
  {\end@float}
\renewcommand{\paragraph}{\@startsection{paragraph}{4}{\z@}{0.6ex plus
0.5ex minus .2ex}{-1em}{\normalsize\bf}}
\makeatother

\definecolor{darkblue}{rgb}{0, 0, 0.5}
\hypersetup{colorlinks=true, citecolor=darkblue, linkcolor=darkblue, urlcolor=darkblue}

\definecolor{ipcolor}{RGB}{15, 76, 129}      %
\definecolor{spcolor}{RGB}{139, 26, 26}      %
\newcommand{\ipstyle}[1]{\textcolor{ipcolor}{#1}}
\newcommand{\spstyle}[1]{\textcolor{spcolor}{#1}}
\newcommand{\IP}{\ipstyle{Isolated-Policy}\xspace}
\newcommand{\SP}{\spstyle{Shared-Policy}\xspace}
\newcommand{\IPa}{\ipstyle{IP}\xspace}
\newcommand{\SPa}{\spstyle{SP}\xspace}
\definecolor{answergray}{gray}{0.93}
\newcommand{\answerbox}[1]{\noindent\colorbox{answergray}{\parbox{\dimexpr\linewidth-2\fboxsep\relax}{#1}}\par\medskip}

\title{When Does Multi-Agent RL Improve LLM Workflows? Workflow, Scale, and Policy-Sharing Tradeoffs}

\author{%
  Yifan Zeng$^{1}$, Yiran Wu$^{2}$, Yaolun Zhang$^{1}$, \\
  \textbf{Wentian Zhao$^{3}$, Kun Wan$^{3}$, Qingyun Wu$^{2,4}$, Huazheng Wang$^{1,4}$} \\[3pt]
  $^{1}$Oregon State University \quad $^{2}$Pennsylvania State University \\
  $^{3}$Adobe Inc. \quad $^{4}$AG2AI, Inc. \\[3pt]
  \texttt{\small \{zengyif, zhanyaol, huazheng.wang\}@oregonstate.edu} \\
  \texttt{\small \{yiran.wu, qingyun.wu\}@psu.edu}, \texttt{\small \{wezhao, kuwan\}@adobe.com}
}

\iclrpreprint %

\begin{document}

\maketitle

\ificlrfinal
\begin{center}
\vspace{-10pt}
\href{https://xhmy.github.io/marl-llm-workflows/}{\faGlobe\ \texttt{Website}}
\quad
\href{https://github.com/XHMY/marl-llm-workflows}{\faGithub\ \texttt{XHMY/marl-llm-workflows}}
\end{center}
\fi

\begin{abstract}
Outcome-reward RL can train every role of a multi-agent LLM workflow. Its benefit is hard to predict because the reward assigns no credit to individual roles and a shared policy trades pooled data against specialization. We ask when this training improves a workflow over its base model and single-agent RL, and what each role learns. We train three workflows (Eval-Opt, Voting, Orch-Workers) with GRPO on math and code at three Qwen3 scales, with one policy for all roles (\SP, \SPa) or one per role type (\IP, \IPa). Every workflow gains by adapting to its structure; this does not make a role better than training it alone. Single-agent RL in the same structure matches or beats trained Voting and Orch-Workers, and within the grid only Eval-Opt under \IPa exceeds it. Eval-Opt's evaluator learns when to retry and its generator learns to complete the retried attempts, a joint strategy that a single policy trained alone does not acquire. Under \SPa, per-output token averaging sets each role's gradient share, a hidden hyperparameter absent from single-agent RL, and the solving roles learn slowly; reweighting it gives \SPa the speed and peak of \IPa, and its decline. Under \IPa, a parallel role, whose instances share one adapter, takes one optimizer step per instance per training step and can drift after the peak as its entropy rises and accuracy declines. Merging those steps delays the decline.
\end{abstract}

\section{Introduction}
\label{sec:intro}

Multi-agent LLM workflows improve on a single model call by splitting generation, evaluation, aggregation, or delegation across several calls~\citep{yang2026understanding,wu2023autogen,madaan2023selfrefine,wang2023selfconsistency,schluntz2024building}. Their roles are usually played by a model trained to answer on its own, so a generator never adapts to the evaluator that checks it. Reinforcement learning with verifiable rewards (RLVR) improves LLM reasoning on mathematics and code~\citep{shao2024deepseekmath,guo2025deepseek,yu2026dapo,zhao2025absolute} and trains multi-step agents from an end-of-episode outcome signal~\citep{jin2025searchr1,wei2025webagentr1,wang2025ragen}, so the same reward can train every role of a workflow.

\begin{figure}[t]
\centering
\includegraphics[width=\textwidth]{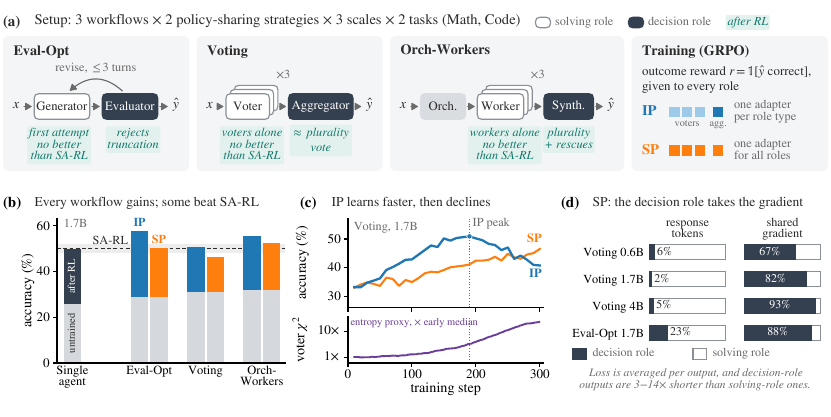}
\caption{\textbf{Setup and main findings.} (a) \IPa: one adapter per role type (the parallel role's instances share one); \SPa: one for all roles; tags: behavior after RL. (b) Accuracy at 1.7B before (gray) and after RL; band: within 2 points of one SA-RL response. (c) Voting at 1.7B: \IPa and \SPa validation accuracy, and the \IPa voter adapter's $\chi^2$ (an entropy proxy) over its early median. (d) \SPa decision role's share of response tokens and of the shared gradient in four cells.}
\label{fig:topology_routing}
\end{figure}

Whether this training helps is hard to predict. The reward indicates whether the final answer is correct, not which role produced it, yet every role receives it (Fig.~\ref{fig:topology_routing}(a)). In cooperative multi-agent RL, sharing parameters improves sample efficiency but can interfere with learning distinct roles, and its benefit depends on the environment~\citep{christianos2021scaling}. Multi-agent extensions of Group Relative Policy Optimization (GRPO)~\citep{shao2024deepseekmath} compare shared and per-role policies in systems with one agent per role (AT-GRPO~\citep{zhao2025stronger}, Dr.~MAS~\citep{feng2026drmas}; \S\ref{sec:related}), and in AT-GRPO the stronger strategy changes with task and model size. We ask: \textbf{when does RL training improve a multi-agent workflow over its base model, and what does each role learn?}

We train three workflows (Eval-Opt, Voting, Orch-Workers) at three scales (0.6B, 1.7B, 4B) on math and code under two policy-sharing strategies, one policy for all roles (\SP, \SPa) or one per role type (\IP, \IPa). \emph{Solving roles} write candidate solutions and a \emph{decision role} sets the final answer. Voting and Orch-Workers have a \emph{parallel role}, whose instances run in parallel within one episode and share one adapter under \IPa (Fig.~\ref{fig:topology_routing}(a)). Each of the 36 \emph{cells} is compared with its base workflow and with single-agent RL (SA-RL), alone and placed in the same structure.

We report four findings.
\begin{itemize}[leftmargin=*,itemsep=2pt,topsep=2pt]
\item \textbf{Which workflows gain, and where the gain comes from.} Every workflow gains by adapting to its structure (Fig.~\ref{fig:topology_routing}(b)), and each decision role learns the rule its structure needs (Fig.~\ref{fig:topology_routing}(a)). Adaptation does not make a role better than training it alone. SA-RL in the same structure matches or beats trained Voting and Orch-Workers, and within the grid only Eval-Opt under \IPa exceeds it (\S\ref{sec:results:landscape}).
\item \textbf{Why Eval-Opt gains most.} Its two roles learn a joint strategy. The evaluator learns when to retry, sending unfinished attempts back, and the generator learns to complete them in later turns (Fig.~\ref{fig:topology_routing}(a)). A single policy trained alone does not acquire this strategy. With one response at the same length cap, single-agent RL falls short of Eval-Opt under \IPa; the same retries close part of the gap, and a longer response can close most of it (\S\ref{sec:results:eval_opt_loop}).
\item \textbf{Why a shared policy learns more slowly.} \IPa learns faster and usually peaks higher (Fig.~\ref{fig:topology_routing}(c)). Under \SPa, averaging each output's loss over its own tokens gives a decision role with much shorter outputs well over its uniform share of the gradient, and the solving roles learn slowly (Fig.~\ref{fig:topology_routing}(d)). This implicit share is a hyperparameter single-agent RL lacks; reweighting it gives \SPa the speed and peak of \IPa, and also its decline (\S\ref{sec:results:ip_vs_sp}).
\item \textbf{Why isolated policies decline after their peak.} Under \IPa a parallel role takes one optimizer step per instance per training step, and its adapter can drift after the peak as its entropy rises (Fig.~\ref{fig:topology_routing}(c)). Merging those steps delays the decline (\S\ref{sec:results:ip_decline}).
\end{itemize}
Each strategy carries its own problem to watch: under \SPa the solving roles' gradient share, under \IPa the parallel role's drift.

\section{Related Work}
\label{sec:related}

\paragraph{Multi-agent RL training of LLM workflows.}
Several works extend GRPO to multi-agent LLM systems with structure-specific objectives: agent- and turn-wise grouping in AT-GRPO~\citep{zhao2025stronger}, centralized group-relative advantages in MAGRPO~\citep{liu2025magrpo}, alignment of variable sub-agent invocations in M-GRPO~\citep{hong2025mgrpo}, heterogeneous group advantages in MHGPO~\citep{chen2025mhgpo}, and per-agent LoRA adapters in MARFT~\citep{liao2025marft}. AT-GRPO and Dr.~MAS~\citep{feng2026drmas} compare shared and role-specialized policies, a choice long studied in cooperative multi-agent RL~\citep{terry2020revisiting,christianos2021scaling}. Other LLM systems use a separate model per agent~\citep{motwani2025malt,subramaniam2025multiagent} or shared parameters~\citep{wan2025rema}. Self-Refine~\citep{madaan2023selfrefine} and self-consistency~\citep{wang2023selfconsistency} are inference-time counterparts of Eval-Opt and Voting. We use the agent-wise advantage normalization of Dr.~MAS, which AT-GRPO further splits by turn (Appendix~\ref{sec:appendix:prior_methods}).

\paragraph{Parallel same-role agents and per-agent sample counts.}
Some works train only the aggregator or orchestrator: AggLM~\citep{zhao2025majority} trains it over solutions from a generator kept fixed in its main experiments, and Kimi K2.5~\citep{kimi2026k25} freezes its sub-agents to avoid ambiguous credit assignment and training instability. Others train parallel instances of one role jointly, with one adapter per agent in MAPoRL~\citep{park2025maporl}, one shared model in WideSeek-R1~\citep{xu2026wideseek} and Recursive Agent Optimization (RAO)~\citep{gandhi2026recursive}, and separate leader and sub-agent models in M-GRPO. Several correct for unequal sample counts: Dr.~MAS normalizes advantages per agent and M-GRPO duplicates or drops sub-agent trajectories to a fixed count. RAO and WideSeek-R1 reweight trajectories so sub-agents do not dominate the gradient, and SRPO~\citep{yang2026srpo} divides the summed log-ratio of parallel outputs by the square root of their number. These corrections reweight samples within one gradient; we study how many optimizer steps each role's adapter takes. When each role has its own mini-batches of a fixed number of sequences, as in verl~\citep{sheng2024hybridflow}, a role with more samples takes more optimizer steps per training step (\S\ref{sec:results:ip_decline}). Scaling each mini-batch to its role's sample count gives every role the same number of optimizer steps (Appendix~\ref{sec:appendix:prior_methods}).

\section{Experimental Setup}
\label{sec:setup}

Each of the 36 cells is trained once with a fixed seed (Fig.~\ref{fig:topology_routing}(a)). The controls are the base model alone, each workflow with the base model in every role (the \emph{base workflow}, or untrained workflow), and single-agent RL (SA-RL) at each task and scale. SA-RL is compared in two forms: as one response, and as the \emph{same-structure comparator}, the SA-RL adapter placed in every role of the untrained workflow (Appendix~\ref{sec:appendix:sa_baseline}). Training uses GRPO without an explicit KL penalty and samples $n=8$ full-workflow rollouts, or \emph{episodes}, per problem. Every role receives the episode's outcome reward, the \emph{broadcast reward}, normalized per problem and role identically under both strategies (Appendix~\ref{sec:appendix:rewards}).

\textbf{Data and models.} Math is DAPO-Math-17K~\citep{yu2026dapo} and Code is DeepCoder~\citep{luo2025deepcoder}; validation accuracy is pass@1 on the 1412-problem DAPO-Math and 1000-problem DeepCoder-PrimeIntellect validation splits. We use Qwen3~\citep{yang2025qwen3} at 0.6B, 1.7B, and 4B, and cap each Math response at 5{,}120 tokens (Appendix~\ref{sec:appendix:hyperparameters}).

\subsection{Multi-Agent Workflows}
\label{sec:setup:frameworks}

The workflows adapt the evaluator-optimizer, parallelization (voting), and orchestrator-workers patterns of \citet{schluntz2024building}. In \emph{Eval-Opt}, a generator answers and an evaluator returns a verdict and critique; any verdict other than correct sends the answer back for revision, up to three generator turns on Math. In \emph{Voting}, three generators (the voters) answer independently and an aggregator selects among or combines their answers. In \emph{Orch-Workers}, an orchestrator proposes a plan, three workers write candidate solutions, and a synthesizer writes the final answer. The \emph{decision roles}, whose output sets the outcome, are the evaluator, aggregator, and synthesizer. The voters, workers, and Eval-Opt generator are the \emph{solving roles}; the generator's first turn is its \emph{first attempt}. The voters and workers (three per episode) are \emph{parallel roles}, whose instances run in parallel within one episode and share one adapter under \IPa; Eval-Opt has none.

\subsection{Policy-Sharing Strategy}
\label{sec:setup:policy_sharing}

Under \SP (\SPa), all roles of a workflow use one policy. Under \IP (\IPa), each role type has its own adapter, shared by the role's instances. Both strategies use rank-64 LoRA adapters~\citep{hu2022lora} on a frozen base model. The strategy sets which adapter receives each role's loss. With fixed-size mini-batches and AdamW, it therefore also sets how a training step's optimizer steps are divided among the roles and whether the roles share the optimizer's running statistics (Appendix~\ref{sec:appendix:hyperparameters}).

\subsection{Diagnostics}
\label{sec:setup:diagnostics}

\paragraph{Training diagnostics.} For each adapter we log the training perplexity of its own sampled responses, the gradient norm, and $\chi^2=\mathbb{E}_t[(r_t-1)^2]$, where $r_t$ compares the trainer's token probability at the sampling temperature with the sampler's probability before temperature scaling. This $\chi^2$ grows with the role's per-token entropy. We also log the per-token entropy pooled over roles. Each is reported relative to its early median (Appendix~\ref{sec:appendix:measurements}). \emph{Training success} is the fraction of training episodes with a correct final answer, and the \emph{cap-hit rate} is the fraction of training responses that reach the length cap. Checkpoint probes keep a set of sampled responses fixed and measure two quantities per role: its per-token entropy at the sampling temperature, and its \emph{change between checkpoints}, the per-token KL divergence between nearby checkpoints on those responses (Appendix~\ref{sec:appendix:measurements}).

\paragraph{Validation and the selected peak.} The \emph{selected peak} of a run is its highest validation pass@1 among checkpoints at step $\leq 300$. Accuracy is reported there unless stated otherwise, and post-peak declines are measured over each run's full validated horizon. Peak differences below about 2 points, within checkpoint-to-checkpoint noise, are treated as ties. A run \emph{declines} when its last validated checkpoint is below its selected peak by more than checkpoint noise, judged on per-problem outcomes with the peak chosen on held-out problems, so that selecting the best checkpoint does not by itself create the drop.

\paragraph{Role-level decomposition.} Each role output is graded on its own answer, independently of the broadcast reward (Appendix~\ref{sec:appendix:trajectory_statistics}), and is \emph{truncated} if it reaches the length cap and \emph{finished} otherwise. The \emph{plurality vote} is the most common boxed answer of the voters or workers. A \emph{selection loss} is a Voting or Orch-Workers episode with a correct voter or worker and a wrong final answer; a \emph{rescue} is an Orch-Workers episode with no correct worker and a correct final answer. An Eval-Opt episode is \emph{recovered} when its first attempt is wrong and its final answer correct; the loop's \emph{net contribution} is final minus first-attempt accuracy. An output whose own answer is wrong in a rewarded episode is \emph{rewarded but wrong}.

\begin{table}[t]
\centering
\small
\caption{Accuracy (\%) at the selected peak, the best validated checkpoint at step $\leq 300$. Math: separate evaluation of each checkpoint (1412 problems). Code: re-evaluation (1000 problems) of the checkpoint selected on the validation series logged during training, SA-RL included. Base model: untrained, alone (single) or inside the workflow. Residual: the cell minus one SA-RL response, or minus SA-RL in the same structure (the SA-RL adapter placed in every role of the untrained workflow). Bold: $|\text{residual}| \geq 5$.}
\label{tab:main_results}
\newcommand{\posr}[1]{\textcolor{green!55!black}{#1}}
\newcommand{\negr}[1]{\textcolor{red!70!black}{#1}}
\setlength{\tabcolsep}{3pt}
\begin{tabular}{@{}lllcccccllll@{}}
\toprule
 & & & \multicolumn{2}{c}{Base model} & \multicolumn{3}{c}{Selected peak} & \multicolumn{4}{c}{Residual vs SA-RL (pp)} \\
\cmidrule(lr){4-5}\cmidrule(lr){6-8}\cmidrule(l){9-12}
 & & & & & & & & \multicolumn{2}{c}{one response} & \multicolumn{2}{c}{same structure} \\
\cmidrule(lr){9-10}\cmidrule(l){11-12}
Task & Scale & Workflow & Single & Workflow & SA-RL & \IPa & \SPa & \IPa & \SPa & \IPa & \SPa \\
\midrule
\multirow{9}{*}{\rotatebox[origin=c]{90}{Math}} & \multirow{3}{*}{0.6B} & Eval-Opt & \multirow{3}{*}{10.1} & 11.3 & \multirow{3}{*}{31.4} & 39.5 & 29.5 & \posr{$\mathbf{+8.2}$} & \negr{$-1.9$} & \posr{$\mathbf{+6.7}$} & \negr{$-3.3$} \\
 &  & Voting &  & 11.7 &  & 28.5 & 28.3 & \negr{$-2.8$} & \negr{$-3.1$} & \negr{$-3.6$} & \negr{$-3.9$} \\
 &  & Orch-Workers &  & 13.0 &  & 36.0 & 31.6 & \posr{$+4.7$} & \posr{$+0.2$} & \negr{$-1.6$} & \negr{$\mathbf{-6.1}$} \\
\addlinespace[2pt]
 & \multirow{3}{*}{1.7B} & Eval-Opt & \multirow{3}{*}{25.9} & 28.7 & \multirow{3}{*}{49.9} & 57.6 & 50.5 & \posr{$\mathbf{+7.6}$} & \posr{$+0.6$} & \posr{$\mathbf{+9.1}$} & \posr{$+2.1$} \\
 &  & Voting &  & 31.2 &  & 50.9 & 46.5 & \posr{$+1.0$} & \negr{$-3.4$} & \posr{$+0.7$} & \negr{$-3.7$} \\
 &  & Orch-Workers &  & 32.1 &  & 55.5 & 52.3 & \posr{$\mathbf{+5.5}$} & \posr{$+2.4$} & \negr{$-3.0$} & \negr{$\mathbf{-6.2}$} \\
\addlinespace[2pt]
 & \multirow{3}{*}{4B} & Eval-Opt & \multirow{3}{*}{46.1} & 48.2 & \multirow{3}{*}{66.6} & 75.2 & 65.8 & \posr{$\mathbf{+8.6}$} & \negr{$-0.8$} & \posr{$+3.3$} & \negr{$\mathbf{-6.1}$} \\
 &  & Voting &  & 53.0 &  & 68.1 & 59.4 & \posr{$+1.6$} & \negr{$\mathbf{-7.2}$} & \negr{$-2.8$} & \negr{$\mathbf{-11.6}$} \\
 &  & Orch-Workers &  & 54.8 &  & 67.9 & 69.6 & \posr{$+1.4$} & \posr{$+3.1$} & \negr{$\mathbf{-8.1}$} & \negr{$\mathbf{-6.4}$} \\
\midrule
\multirow{9}{*}{\rotatebox[origin=c]{90}{Code}} & \multirow{3}{*}{0.6B} & Eval-Opt & \multirow{3}{*}{5.5} & 6.5 & \multirow{3}{*}{11.8} & 15.3 & 13.2 & \posr{$+3.5$} & \posr{$+1.4$} & \posr{$+4.0$} & \posr{$+1.9$} \\
 &  & Voting &  & 6.9 &  & 13.1 & 11.7 & \posr{$+1.3$} & \negr{$-0.1$} & \posr{$+0.6$} & \negr{$-0.8$} \\
 &  & Orch-Workers &  & 6.7 &  & 15.9 & 14.6 & \posr{$+4.1$} & \posr{$+2.8$} & \posr{$+1.6$} & \posr{$+0.3$} \\
\addlinespace[2pt]
 & \multirow{3}{*}{1.7B} & Eval-Opt & \multirow{3}{*}{12.6} & 12.0 & \multirow{3}{*}{19.4} & 22.7 & 16.2 & \posr{$+3.3$} & \negr{$-3.2$} & \posr{$+3.1$} & \negr{$-3.4$} \\
 &  & Voting &  & 12.7 &  & 18.9 & 16.7 & \negr{$-0.5$} & \negr{$-2.7$} & \negr{$-0.6$} & \negr{$-2.8$} \\
 &  & Orch-Workers &  & 15.0 &  & 21.8 & 19.0 & \posr{$+2.4$} & \negr{$-0.4$} & \posr{$+0.1$} & \negr{$-2.7$} \\
\addlinespace[2pt]
 & \multirow{3}{*}{4B} & Eval-Opt & \multirow{3}{*}{22.7} & 25.5 & \multirow{3}{*}{33.3} & 30.3 & 38.1 & \negr{$-3.0$} & \posr{$+4.8$} & \negr{$\mathbf{-9.0}$} & \negr{$-1.2$} \\
 &  & Voting &  & 26.6 &  & 32.2 & 29.7 & \negr{$-1.1$} & \negr{$-3.6$} & \negr{$-3.6$} & \negr{$\mathbf{-6.1}$} \\
 &  & Orch-Workers &  & 27.8 &  & 33.6 & 31.3 & \posr{$+0.3$} & \negr{$-2.0$} & \negr{$\mathbf{-6.4}$} & \negr{$\mathbf{-8.7}$} \\
\bottomrule
\end{tabular}
\end{table}

\section{Results}
\label{sec:results}

RL adapts each workflow's roles to its structure, but this adaptation does not make a role better than training it alone, and within the grid only Eval-Opt under \IPa exceeds single-agent RL placed in the same structure (\S\ref{sec:results:landscape}, Table~\ref{tab:main_results}). Eval-Opt's evaluator and generator learn a joint strategy for the revision loop that a single policy trained alone does not acquire (\S\ref{sec:results:eval_opt_loop}). Under \SPa, averaging each output's loss over its own tokens sets each role's share of the gradient, a hyperparameter that single-agent RL does not have, and the solving roles learn slowly (\S\ref{sec:results:ip_vs_sp}). Under \IPa, the adapter of a parallel role can drift and accuracy can decline after the peak; merging its instances' optimizer steps into one per training step delays the decline (\S\ref{sec:results:ip_decline}).

\subsection{Which Workflows Gain From RL, and Where Does the Gain Come From?}
\label{sec:results:landscape}

\answerbox{Every workflow gains from end-to-end RL, and the gain is an adaptation to the workflow: each solving role improves at the subtask its structure assigns to it, and each decision role learns the rule its structure needs. This adaptation does not make a role better than training it alone. Graded on their own answers, the solving roles end below single-agent RL, and single-agent RL placed in the same structure matches or beats trained Voting and Orch-Workers. Within the grid, only Eval-Opt under isolated policies exceeds single-agent RL placed in its structure.}

\begin{figure}[t]
\centering
\includegraphics[width=\textwidth]{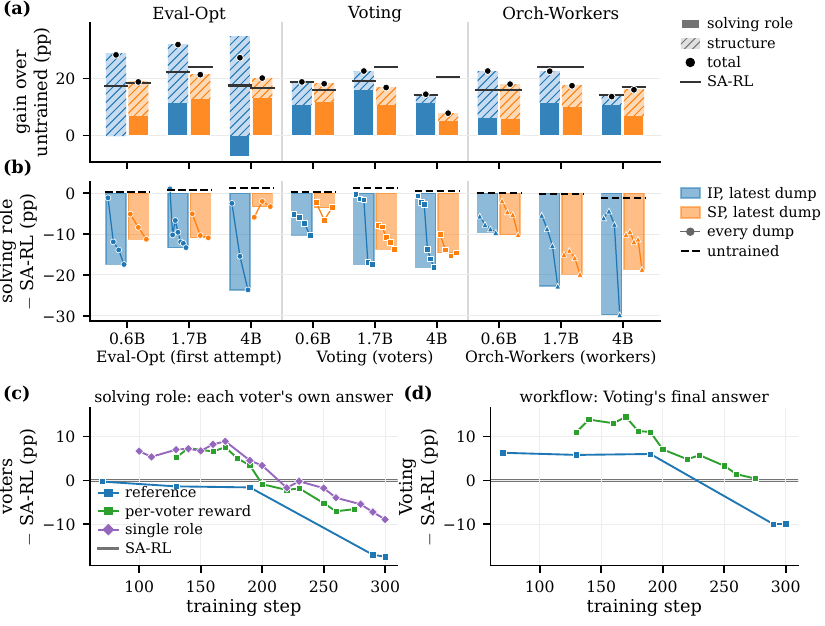}
\caption{RL gain and solving roles on Math, 1412 validation problems. (a)~Gain over the untrained workflow at the checkpoint nearest the selected peak: solving-role gain (solid) plus structural gain (hatched); dot: total; line: SA-RL's gain over the untrained single model. (b)~Solving role minus SA-RL at the nearest evaluated step, on the same problems; bars: latest sampled checkpoint; dots: all sampled checkpoints; dashed: untrained. (c)~Voter accuracy minus SA-RL, unpaired, for \texttt{Voting-IP-1.7B-Math} (reference) and its per-voter-reward and single-role controls. (d)~Voting accuracy minus SA-RL for the two Voting runs of (c) at the same checkpoints; (c) and (d) share the y axis. Values of (a) and (b): Table~\ref{tab:solving_role_vs_sa}.}
\label{fig:reward_broadcast}
\end{figure}

\paragraph{Result.}
RL raises every cell above its base workflow on both tasks and reorders the workflows (Table~\ref{tab:main_results}). On Math, Eval-Opt goes from the weakest untrained workflow to the strongest under \IPa. SA-RL placed in the same structure matches or beats trained Voting and Orch-Workers on both tasks. Only Eval-Opt under \IPa exceeds it, on Math at every scale and on Code at the two smaller scales (Table~\ref{tab:main_results}, same-structure columns; Table~\ref{tab:sa_same_structure}). The gain decomposes into what the solving role learns and what the structure adds on top of it (Fig.~\ref{fig:reward_broadcast}(a)).

\paragraph{Gain decomposition: only the structural part can exceed SA-RL.}
The structural gain is the change in what the plurality vote, the synthesizer or the revision loop adds on top of the solving role. In every Math cell the solving role gains less than the single model under SA-RL and ends below it (Fig.~\ref{fig:reward_broadcast}(b)). A workflow's margin over one SA-RL response is thus its structural gain minus that deficit (Table~\ref{tab:solving_role_vs_sa}). Under \IPa, Voting's gain is mostly its voters' own, and Eval-Opt has the largest structural gain at every scale, from its revision loop (\S\ref{sec:results:eval_opt_loop}). A workflow can therefore exceed SA-RL only through what its structure adds.

\paragraph{Solving roles: below SA-RL with or without the broadcast reward.}
Graded on their own answers, the voters, the workers and the Eval-Opt first attempt all end below SA-RL on Math (Fig.~\ref{fig:reward_broadcast}(b)). First-attempt accuracy is not a standalone measure of the generator, which finishes its first attempt mainly when it is likely correct (\S\ref{sec:results:eval_opt_loop}). The broadcast reward scores each output by its episode, so wrong and truncated outputs are rewarded inside correct episodes (Table~\ref{tab:solving_role_vs_sa}). Scoring each voter on its own answer, or training one generator alone on its own answer, still leaves the solving role below SA-RL after the peak (Fig.~\ref{fig:reward_broadcast}(c); settings in Table~\ref{tab:controls}). Voting's margin over SA-RL moves with its voters' deficit under both rewards (Fig.~\ref{fig:reward_broadcast}(d)). The broadcast reward widens the late gap but does not create it. A solving role trained inside a workflow thus ends no better than the same model trained alone.

\begin{table}[t]
\centering
\caption{Decision-role rules on Math at the latest sampled \IPa and \SPa checkpoints, all 1412 validation problems. $k/n$: count and denominator. Voting: workflow minus plurality-vote accuracy (points); lone correct voters picked against a wrong pair. Eval-Opt: evaluator's first verdict on truncated (T) or finished (F) wrong first attempts. Orch-Workers: share of answer-only synthesizer outputs (these return the workers' plurality answer) and of episodes with a selection loss or a rescue (\S\ref{sec:setup:diagnostics}). Steps, untrained workflow, other columns: Table~\ref{tab:decision_role_rules_full}; definitions: \S\ref{sec:appendix:sp_role_capture_signatures}.}
\label{tab:decision_role_rules}
\footnotesize
\setlength{\tabcolsep}{2pt}
\begin{tabular*}{\textwidth}{@{\extracolsep{\fill}}l cc cc cc cc cc cc cc@{}}
\toprule
 & \multicolumn{4}{c}{\textit{Voting aggregator}} & \multicolumn{4}{c}{\textit{Eval-Opt evaluator}} & \multicolumn{6}{c}{\textit{Orch-Workers synthesizer}} \\
\cmidrule(lr){2-5}\cmidrule(lr){6-9}\cmidrule(lr){10-15}
 & \multicolumn{2}{c}{$\Delta$ vs.\ plurality} & \multicolumn{2}{c}{Lone correct} & \multicolumn{2}{c}{Accepts T-wrong} & \multicolumn{2}{c}{Rejects F-wrong} & \multicolumn{2}{c}{Answer-only} & \multicolumn{2}{c}{Loss} & \multicolumn{2}{c}{Rescue} \\
\cmidrule(lr){2-3}\cmidrule(lr){4-5}\cmidrule(lr){6-7}\cmidrule(lr){8-9}\cmidrule(lr){10-11}\cmidrule(lr){12-13}\cmidrule(lr){14-15}
Scale & \IPa & \SPa & \IPa & \SPa & \IPa & \SPa & \IPa & \SPa & \IPa & \SPa & \IPa & \SPa & \IPa & \SPa \\
\midrule
0.6B & ${+}0.4$ & ${-}0.1$ & $0/15$ & $2/25$ & $0/1257$ & $24/1007$ & $3/16$ & $2/162$ & $0.06$ & $0.41$ & $0.02$ & $0.03$ & $0.10$ & $0.06$ \\
1.7B & ${+}0.5$ & ${+}0.4$ & $4/21$ & $3/25$ & $73/756$ & $28/714$ & $2/99$ & $12/145$ & $0.39$ & $0.35$ & $0.01$ & $0.04$ & $0.14$ & $0.07$ \\
4B & ${+}3.1$ & ${+}1.2$ & $5/24$ & $8/42$ & $0/837$ & $3/373$ & $1/11$ & $115/200$ & $0.41$ & $0.55$ & $0.01$ & $0.02$ & $0.16$ & $0.09$ \\
\bottomrule
\end{tabular*}
\end{table}

\paragraph{Decision roles: each learns the rule its structure needs.}
Each decision role learns the operation its structure needs (Table~\ref{tab:decision_role_rules}). The Voting aggregator, which receives three answers, scores close to a plurality vote over them. The Eval-Opt evaluator, whose loop recovers almost only unfinished attempts, rejects truncated first attempts (\S\ref{sec:results:eval_opt_loop}). When the Orch-Workers synthesizer writes only an answer, it returns the workers' most common answer; a copied answer cannot rescue an episode with no correct worker, so its rescues come from its longer outputs.

\paragraph{Code: a smaller deficit, since truncated outputs are rarely rewarded.}
On Code the solving roles end below SA-RL by a smaller margin than on Math. Their truncated outputs fall almost only on problems no instance solves, so a truncated output is rarely rewarded (Table~\ref{tab:code_reward_accounting}). At a common step cap, isolated and shared policies differ little on Code outside Eval-Opt (\S\ref{sec:results:ip_vs_sp}).

\subsection{Why Does Eval-Opt Gain Most?}
\label{sec:results:eval_opt_loop}

\answerbox{Its two roles learn a joint strategy. The evaluator learns the quality criterion that matters here, whether an attempt finished, and sends truncated attempts back; the generator learns to finish an attempt when it is likely correct and to complete an unfinished one on the next turn, so the loop raises the success rate of the retried attempts above the untrained loop and, at 1.7B, above a single-agent generator given the same retries. A single policy trained alone does not acquire this strategy. With one response at the same cap, single-agent RL falls short of Eval-Opt under isolated policies; given the same retries it closes part of the gap, and a longer response closes most of it at the larger scales.}

\begin{figure}[t]
\centering
\includegraphics[width=\textwidth]{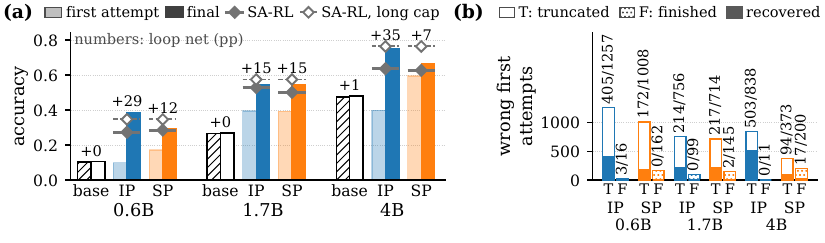}
\caption{Eval-Opt on Math, 1412 validation problems, each generator turn graded on its own answer. (a)~First-attempt (light) and final (dark) accuracy at the latest sampled checkpoint and untrained (hatched, open); gray diamonds: SA-RL at the nearest step; dashed open diamonds: SA-RL with one response capped at 15{,}360 tokens (10{,}240 at 1.7B); numbers: loop net contribution. (b)~Wrong first attempts, truncated (T) or finished (F); filled: recovered. Bars and dashed open diamonds: separate evaluations; gray diamonds: validation accuracy.}
\label{fig:eval_opt_loop}
\end{figure}

\paragraph{Result.}
Eval-Opt under \IPa is the only workflow above both one SA-RL response and SA-RL in its untrained structure at every Math scale (Table~\ref{tab:main_results}). Its loop recovers truncated first attempts and almost never loses a correct one (Fig.~\ref{fig:eval_opt_loop}), so the gain comes from the added turns.

\paragraph{Role outputs: the evaluator retries truncation, the generator learns when to finish.}
The evaluator sends back most truncated first attempts (Table~\ref{tab:decision_role_rules}). The \IPa generator learns to finish its first attempt mainly when it is likely correct (Fig.~\ref{fig:eval_opt_first_attempt}), and its retried attempts succeed far more often than in the untrained loop (Fig.~\ref{fig:eval_opt_loop}(b)).

\paragraph{Controls: a fixed retry rule reproduces most of the loop; the generator carries the rest.}
A fixed rule that retries every truncated attempt stays within noise of the trained evaluator at 1.7B and 4B but not at 0.6B, where the evaluator's text often states the answer (Table~\ref{tab:eval_opt_controllers}). Under that rule the \IPa generator does not fall below the SA-RL generator, clearly outperforms it at 1.7B, though with more turns, and leads the \SPa generator by about the trained \IPa minus \SPa gap. SA-RL gains from the same retries, and one longer SA-RL response closes most of Eval-Opt's lead at the two larger scales (Fig.~\ref{fig:eval_opt_loop}(a)).

\subsection{Why Does a Shared Policy Learn More Slowly?}
\label{sec:results:ip_vs_sp}

\answerbox{The two strategies differ in whether the roles share one gradient. Under a shared policy the roles' shares are set by an implicit weighting: each output's loss is averaged over its own tokens, so the decision role, with much shorter outputs, takes well over its uniform share, and the solving roles learn slowly. This share is a hyperparameter that single-agent RL does not have. Reweighting the loss by role length gives the shared policy the isolated policies' speed and peak, and with them the decline of \S\ref{sec:results:ip_decline}. Isolated policies have no share to set, since each role has its own update, and a different problem to watch.}

\begin{figure}[t]
\centering
\includegraphics[width=\textwidth]{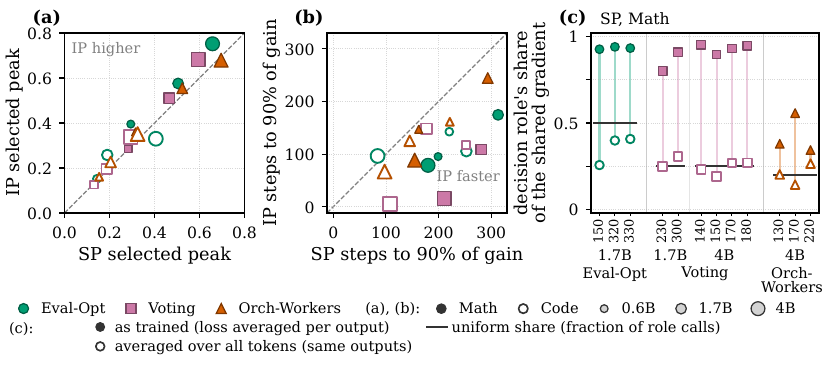}
\caption{\IPa against \SPa per matched pair. (a) Selected peak. (b) Training steps to 90\% of the pair's common gain. (c) \SPa decision role's gradient share at probed Math checkpoints, each output's loss averaged over its own tokens as in training (filled) or over all tokens (open); tick: uniform share, the role's fraction of calls. Color and marker: workflow; in (a, b) filled: Math, open: Code, size: scale.}
\label{fig:ip_vs_sp}
\end{figure}

\paragraph{Result.}
\IPa learns faster than \SPa in nearly every matched pair and peaks higher in most (Fig.~\ref{fig:ip_vs_sp}(a, b); every pair's series in Fig.~\ref{fig:ip_sp_curves}). Where \IPa declines, its lead shrinks or reverses (Table~\ref{tab:peak_last}, Panel~B). On Code, at a common step cap, \IPa leads clearly only for Eval-Opt at the two smaller scales (Table~\ref{tab:code_task}).

\paragraph{Gradients: the decision role holds most of the shared gradient.}
Under the per-output token mean, the decision role on Math holds well over its uniform share at every probed checkpoint, and about that share or less under an all-token mean (Fig.~\ref{fig:ip_vs_sp}(c); Appendix~\ref{sec:appendix:gradient_measurements}). The slow learning of \SPa is not due to smaller parameter changes; its adapter moves farther per step than the \IPa adapters in most pairs (Table~\ref{tab:gradient_evidence}, Panel~D2). On Code the shares stay near uniform.

\paragraph{Role outputs: the \SPa deficit sits in the solving roles.}
Early in training the \SPa solving roles trail in nearly every Math pair (Table~\ref{tab:ip_sp_decomposition}). At the peak the gap lies in the solving roles for Voting at the two larger scales and in the generator's later turns for Eval-Opt (\S\ref{sec:results:eval_opt_loop}). In Orch-Workers any lead of the \IPa workers does not carry over to episode accuracy.

\paragraph{Controls: length reweighting gives \SPa the \IPa speed and the \IPa decline.}
Scaling each role's loss by its mean output length in \texttt{Voting-SP-1.7B-Math} moves the shared gradient toward the voters. The shared policy then learns almost as fast as the isolated policies, peaks within noise of them and declines (Table~\ref{tab:controls}; Fig.~\ref{fig:ip_sp_curves}). Its voters close more than half of their early gap to the isolated voters (Appendix~\ref{sec:appendix:gradient_measurements}), so the share contributes to the solving roles' slow early learning.

\subsection{Why Do Isolated Policies Decline After Their Peak?}
\label{sec:results:ip_decline}

\answerbox{Under isolated policies a parallel role, one whose instances share one adapter, takes one optimizer step per instance in each training step. On Math at the two larger scales this adapter drifts: its parameter changes keep their direction more than noise would, and its outputs grow less stable, with per-token entropy and change between checkpoints already high at the peak and rising, while the decision roles' do not. The decline requires neither the workflow nor the broadcast reward, and it follows the gradient share: a shared policy reweighted toward its solving roles declines too. Merging the instances' steps into one per training step preserves the learning speed and delays the decline.}

\begin{figure}[t]
\centering
\includegraphics[width=\textwidth]{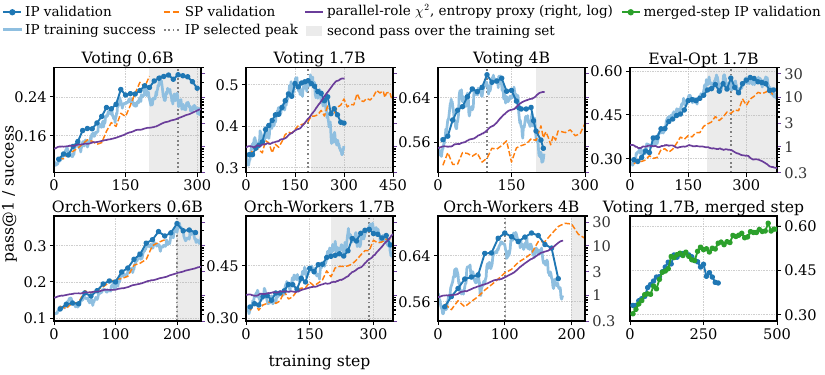}
\caption{Isolated-policy Math runs with a parallel role and \texttt{Eval-Opt-IP-1.7B-Math}: validation accuracy (dashed: matched \SPa run), training success and the parallel role's (Eval-Opt: generator's) logged $\chi^{2}$, an entropy proxy, over its early median. Dotted: selected peak; shaded: second training pass. The two 0.6B runs do not decline (Appendix~\ref{sec:appendix:peak_last}). Last panel: validation accuracy of \texttt{Voting-IP-1.7B-Math} and of its merged-step control, which takes one generator step per training step (Table~\ref{tab:controls}).}
\label{fig:ip_decline}
\end{figure}

\paragraph{Result.}
On Math, Voting and Orch-Workers under \IPa at 1.7B and 4B decline after their selected peak (Fig.~\ref{fig:ip_decline}; Table~\ref{tab:peak_last}). \texttt{Eval-Opt-IP-1.7B-Math} declines late, without a parallel role or a rise in its entropy proxy. Over the same steps \texttt{Voting-SP-1.7B-Math}, \texttt{Orch-Workers-SP-1.7B-Math} and \texttt{Eval-Opt-SP-1.7B-Math} stay at their peak (Fig.~\ref{fig:ip_sp_curves}; Appendix~\ref{sec:appendix:peak_last}), and SA-RL at 1.7B peaks later and ends near its peak (Table~\ref{tab:appendix:intervention_controls}). The robust declines are confined to isolated policies.

\paragraph{Training metrics: drift and instability are localized in the parallel role.}
On fixed responses, the parallel role's per-token entropy and change between checkpoints are already high at the peak and rise through the decline, while the decision roles change only in brief jumps. The \SPa and SA-RL solving roles change far less (Table~\ref{tab:checkpoint_divergence}). In the non-declining \texttt{Orch-Workers-IP-0.6B-Math} the workers' entropy also rises, but their change stays small. Training success falls together with validation accuracy, and at 4B the decline begins within the first epoch, so it is not overfitting (Table~\ref{tab:decline_parameters}). The parallel role's logged $\chi^{2}$, an entropy proxy, rises above its early level in the declining Voting and Orch-Workers runs, so the drift is visible in the training logs (Fig.~\ref{fig:ip_decline}). In most declining runs with a parallel role, its adapter's changes also keep their direction more than noise would (Appendix~\ref{sec:appendix:decline_parameters}). The instability and the drift are localized in the parallel role.

\paragraph{Role outputs: the lost accuracy comes from the parallel role.}
In the declining Voting and Orch-Workers runs, fewer problems have a correct voter or worker after the peak, selection losses rise far less than the voters' and workers' accuracy falls, and by the last sampled checkpoint these outputs reach the length cap more often (Table~\ref{tab:decline_windows}). The lost accuracy comes from the parallel role.

\paragraph{Controls: the decline needs neither the workflow nor the broadcast reward.}
The single-role control, which has no workflow, and the per-voter-reward control both decline, so the decline requires neither the workflow nor the broadcast reward. The shared policy reweighted toward its solving roles also declines, so the decline follows the gradient share. Merging the parallel role's steps into one per training step preserves the learning speed and delays the decline, which does not appear over a validated horizon longer than the reference's (Fig.~\ref{fig:ip_decline}, last panel; Table~\ref{tab:controls}; Appendix~\ref{sec:appendix:intervention}). Its accuracy keeps rising past the reference's peak and ends above single-agent RL (Table~\ref{tab:controls}).

\section{Conclusion}
\label{sec:conclusion}

We asked when end-to-end RL improves a multi-agent LLM workflow over its base model and what each role learns. Every workflow gains by adapting to its structure, but adaptation does not make a role better than training it alone. Single-agent RL in the same structure matches or beats trained Voting and Orch-Workers, so it should be the first baseline to run; within the grid only Eval-Opt under isolated policies exceeds it. Eval-Opt's evaluator learns when to retry and its generator learns to complete the retried attempts in later turns, a joint strategy that a single policy trained alone does not acquire. Joint training can help where roles respond to each other's outputs across turns. Under a shared policy, averaging each output's loss over its own tokens sets each role's share of the gradient, a hyperparameter single-agent RL lacks. Reweighting it gives the shared policy the isolated policies' speed and peak, and with them their decline. Under isolated policies, a parallel role takes one optimizer step per instance per training step and can drift after the peak as accuracy declines. Merging its steps delays the decline. Appendix~\ref{sec:discussion:limitations} discusses the limitations.

\ificlrpreprint\else

\subsection*{AI Use Statement}

LLMs are the object of study in this paper: we train and evaluate LLM-based agent workflows, described in \S\ref{sec:setup}. Separately, an LLM-based assistant (Claude Code) was used for implementation, experiment and analysis scripting, and for editing the text; it is not part of the core methodology, and all generated code was reviewed and validated by the authors.
We take responsibility for the final content of this work, including text, claims, and artifacts produced with the aid of generative AI.

\subsection*{Ethics Statement}

This work trains and evaluates publicly available base models (Qwen3~\citep{yang2025qwen3}) on publicly available math and code datasets (DAPO-Math-17K~\citep{yu2026dapo} and DeepCoder~\citep{luo2025deepcoder}), all openly licensed for research use. It involves no human subjects, crowdsourcing, or personal data, and it releases no new datasets or pretrained models.

\subsection*{Reproducibility Statement}

\S\ref{sec:setup} specifies the task, scale, policy-sharing, and workflow grid, the datasets, the rollout and reward-normalization scheme, and the diagnostics and role-level grading used in \S\ref{sec:results}. \S\ref{sec:appendix:hyperparameters} lists the training hyperparameters for every cell, \S\ref{sec:appendix:rewards} gives the reward functions, the grouping and normalization rule, and which adapter each role's loss updates, \S\ref{sec:appendix:sa_baseline} describes the single-agent baseline, \S\ref{sec:appendix:measurements} the role-level grading and the measurement definitions, and \S\ref{sec:appendix:compute} reports the hardware and compute. \ificlrfinal The training and evaluation code and the configuration files for every cell are available at \url{https://github.com/XHMY/marl-llm-workflows}.\else We will release the training and evaluation code together with the configuration files for every cell.\fi

\fi

\bibliography{colm2026_conference}
\bibliographystyle{iclr2027_conference}

\appendix

\section{Limitations}
\label{sec:discussion}

\label{sec:discussion:limitations}

\textbf{One seed and in-sample peak selection.} Each cell is trained once, so seed variance is not measured and cross-cell counts are descriptive. Each reported peak is selected on the validation set it is scored on, and each problem is sampled once, so small differences between cells lie within evaluation noise. Selecting the peak on held-out problems weakens some margins over the single-agent comparators (Table~\ref{tab:sa_same_structure}, held-out column).

\textbf{Models, tasks and workflows.} The grid uses one model family (Qwen3 at 0.6B, 1.7B and 4B), two tasks with one dataset each (DAPO-Math-17K and DeepCoder), three workflows and one multi-agent GRPO training rule (\S\ref{sec:appendix:prior_methods} relates it to AT-GRPO and Dr.~MAS). Agents solve non-interactive problems without tools or long horizons, and every role is trained on the final-answer reward. Results may differ for other or larger models, other datasets and topologies, agentic settings, or rewards that score intermediate role outputs.

\textbf{LoRA substrate.} Every cell trains LoRA adapters on a frozen base model with GRPO, an adaptive optimizer and no KL penalty~\citep{schulman2025lora}. The parameter-change analysis of \S\ref{sec:appendix:decline_parameters} is tied to this setup. We ran no LoRA-rank, full-parameter or alternative-optimizer ablations.

\textbf{Resource matching.} \IP, \SP and the single-agent baseline share the training configuration and the nominal training-step budget. They are not matched on generated tokens, role calls, parameter count, optimizer steps per adapter, wall-clock time or compute (\S\ref{sec:appendix:compute}, \S\ref{sec:appendix:hyperparameters}). Runs end at different horizons, so contrasts are made at the selected peak, or at a common step where stated. Comparisons hold the response cap fixed, not the token budget; Table~\ref{tab:sa_same_structure} adds single-agent comparators with the same structure, the same number of samples, and one longer response.

\section{Experimental Details}
\label{sec:appendix:details}

\subsection{Hyperparameters}
\label{sec:appendix:hyperparameters}

Table~\ref{tab:appendix:hyperparameters} lists the training hyperparameters. Each training batch is sorted by sequence length, without shuffling, before it is split into mini-batches. One optimizer step is taken per mini-batch of 512 sequences, so the number of optimizer steps per training step follows each adapter's sequence count. Under \IPa the adapter of a parallel role, whose three instances each contribute a full mini-batch, takes three and every other Voting or Orch-Workers role one, for totals of four (Voting) and five (Orch-Workers), the same as the shared adapter under \SPa. An Eval-Opt adapter receives one sequence per generator turn, so with $k$ generator turns per episode on average (at most three on Math and two on Code) each \IPa adapter takes $\lceil k\rceil$ optimizer steps and the shared adapter $\lceil 2k\rceil$. Because each adapter rounds its own partial mini-batch up, the two \IPa adapters together take more steps than the shared adapter: on average 5.3 against 3.8 at 0.6B on Math and 4.0 against 3.0 to 3.8 elsewhere. Each adapter keeps its own AdamW state, so under \SPa all roles share one set of moment estimates.

\paragraph{Learning-rate warmup.} Nine Code cells use learning-rate warmup: Eval-Opt at 4B, Voting at 0.6B and Orch-Workers at 0.6B and 4B under both strategies, and \texttt{Voting-SP-4B-Code}. So do the 0.6B single-agent Code baseline and the single-role control (Table~\ref{tab:controls}, column lr); every other run trains at the constant rate from the first step. The scheduler advances once per adapter updated in a training step, so the warmup covers the first 15 training steps under \SP and in the single-agent baseline and the first 4 to 8 under \IP. It ends before every compared peak.

\begin{table}[t]
\centering
\caption{Training hyperparameters, fixed across the grid unless a per-cell value is given. $^{\dagger}$Cells listed in \S\ref{sec:appendix:hyperparameters}.}
\label{tab:appendix:hyperparameters}
\resizebox{\textwidth}{!}{%
\begin{tabular}{@{}lll@{}}
\toprule
Group & Hyperparameter & Value \\
\midrule
\multirow{7}{*}{Optimization} & Optimizer & AdamW, weight decay 0.01 \\
 & Learning rate & $2 \times 10^{-5}$ \\
 & Learning-rate schedule & constant, no decay \\
 & Warmup, all Math cells and nine Code cells & none \\
 & Warmup, other nine Code cells$^{\dagger}$ & linear from 0 over 15 scheduler updates \\
 & Gradient clipping & 1.0 \\
 & PPO clip ratio (low / high) & 0.2 / 0.28 \\
\midrule
\multirow{4}{*}{LoRA adapter} & Rank $r$ & 64 \\
 & Alpha $\alpha$ & 32 \\
 & Dropout & 0.0 \\
 & Target modules & all linear modules \\
\midrule
\multirow{9}{*}{GRPO} & Algorithm & GRPO (no explicit KL) \\
 & Group size $n$ & 8 \\
 & Sampling unit & episode (independent full-workflow rollout) \\
 & Advantage grouping & per problem and role (\S\ref{sec:appendix:rewards}) \\
 & Advantage normalization & group-relative, before adapter assignment \\
 & Cross-role global baseline & none \\
 & PPO mini-batch size & 512 sequences (64 problems $\times$ $n{=}8$) \\
 & PPO epochs per update & 1 \\
 & Loss aggregation & mean over each sequence's tokens, then over sequences \\
\midrule
\multirow{3}{*}{Batching} & Train batch size (problems / step) & 64 \\
 & Validation batch size & 2048 \\
 & Rollout temperature & 0.7 \\
\midrule
\multirow{6}{*}{Sequence length}
 & Eval-Opt, Math (prompt / response) & 30720 / 5120 \\
 & Eval-Opt, Code (prompt / response) & 10240 / 2048 \\
 & Voting, Math (prompt / response) & 20480 / 5120 \\
 & Voting, Code (prompt / response) & 10240 / 2048 \\
 & Orch-Workers, Math (prompt / response) & 20480 / 5120 \\
 & Orch-Workers, Code (prompt / response) & 10240 / 2048 \\
\midrule
\multirow{4}{*}{Workflow caps}
 & Eval-Opt evaluation rounds per episode, Math & at most 3 (at most 2 revisions) \\
 & Eval-Opt evaluation rounds per episode, Code & at most 2 (at most 1 revision) \\
 & Voting generator calls per episode & 3 \\
 & Orch-Workers worker calls per episode & at most 3 \\
\midrule
\multirow{2}{*}{Training schedule}
 & Validation interval (steps) & 10 \\
 & Checkpoint interval (steps) & 5 or 10, by cell \\
\midrule
\multirow{2}{*}{Adapter assignment} & \IP & one LoRA adapter per role type \\
 & \SP & one shared LoRA adapter across all roles \\
\bottomrule
\end{tabular}%
}
\end{table}

\subsection{Rewards and Advantage Normalization}
\label{sec:appendix:rewards}

\paragraph{Math.} An episode's outcome reward is $1$ if the last \texttt{\textbackslash boxed\{\}} expression of its final answer parses and matches the ground-truth answer of DAPO-Math-17K, and $0$ otherwise, including when the boxed answer is missing or malformed.

\paragraph{Code.} An episode's outcome reward is $1$ if the last Python code block of its final answer passes every test in the problem's hidden test suite and $0$ otherwise, including code that fails to parse or compile or exceeds the per-test time limit.

\paragraph{Grouping, normalization, and adapter assignment.} For each problem, we sample $n=8$ independent full-workflow rollouts and compute one outcome reward per completed rollout, which every role's sequences in that rollout receive (the \emph{broadcast reward}); no per-step or per-role process reward is used. Each advantage is the reward minus the mean of its group, divided by the group's standard deviation, with groups keyed by problem and role. Under Voting on Math, each of the three generator instances forms one group and the aggregator another, each with 8 sequences; under Voting on Code, all generator instances of a problem form one group. Under Eval-Opt, all turns of a role across the problem's rollouts form one group. Under Orch-Workers, the groups are the orchestrator, the pooled workers, and the synthesizer. This is agent-wise normalization in the sense of Dr.~MAS~\citep{feng2026drmas}, not a global baseline pooling all roles (\S\ref{sec:appendix:prior_methods}). Gradients are accumulated only over tokens emitted by a role, with other roles' tokens masked from that role's loss. Under \SP all role losses update one adapter, whereas under \IP each loss updates its role-type adapter. The grouping rule is the same under \SP and \IP and is applied before each role's loss is assigned to an adapter, so normalization groups and parameter-sharing groups are distinct design choices.

\subsection{Single-Agent RL Baseline}
\label{sec:appendix:sa_baseline}

The SA-RL baseline (the SA-RL column of Table~\ref{tab:main_results}) trains one LoRA adapter on a single-role workflow whose prompt matches the multi-agent generator prompt. It uses the same base model, task, optimizer, LoRA configuration, GRPO implementation, and step budget as the multi-agent cells at that scale (resource matching: \S\ref{sec:discussion:limitations}).

\textbf{SA-RL in the same structure.} Table~\ref{tab:sa_same_structure} compares every trained cell with SA-RL comparators built from the SA-RL checkpoint at its selected peak. S is one SA-RL response, the SA-RL column of Table~\ref{tab:main_results}. M draws three samples per problem, each under the training length cap, and scores their plurality answer, with the same number of samples as a Voting episode. T$_{\mathrm{a}}$ places the SA-RL adapter in every role of the untrained workflow. R gives SA-RL up to three turns and continues only after a truncated attempt, the fixed rule~(b) of Appendix~\ref{sec:appendix:eval_opt_turns}. L$_{10}$ and L$_{15}$ give SA-RL one response at a 10240- or 15360-token cap, averaged over three samples. The same-structure columns of Table~\ref{tab:main_results} use T$_{\mathrm{a}}$ for every workflow and task; M and R are controls. The Code cells are the 1000-problem re-evaluation at the training cap.

Cell levels are validation accuracies. Paired tests run on each selected checkpoint's per-problem validation outcomes, or on its full-set trajectory dump where these were not stored; the accuracy of the tested outcomes is shown in parentheses where it differs. They use the exact McNemar test (\S\ref{sec:appendix:measurements}), or a paired $t$-test where a mean over samples enters. S at 0.6B and 1.7B has no stored per-problem outcomes, so its paired test uses the mean of three SA-RL samples at the training cap. The held-out column is the margin over the strongest comparator shown, with each side's selection optimism subtracted. Selection optimism is the selected-peak accuracy minus the mean held-out accuracy when the peak is chosen on the other half of the problems. Tokens per episode count every generated token of the episode; M matches the number of samples, not the token budget.

On Math every Voting cell falls below M and every Orch-Workers cell below T$_{\mathrm{a}}$ (Panel~B), and on Code no 4B cell exceeds T$_{\mathrm{a}}$ (Panel~C).

\begin{table}[t]
\centering
\small
\renewcommand{\arraystretch}{0.92}
\caption{Trained cells against SA-RL comparators from the SA-RL checkpoint at its selected peak: S one response, M a plurality vote over three samples, T$_{\mathrm{a}}$ the T of Table~\ref{tab:main_results} (the SA-RL adapter in every role), R up to three turns continuing after a truncated attempt, and L$_{10}$, L$_{15}$ one response at a 10240- or 15360-token cap. Panel~A: comparator accuracy (\%) and tokens per episode; Panels~B (Math), C (Code): cell minus comparator (points); parentheses: accuracy of the per-problem outcomes used in the tests, where it differs; held-out: margin over the strongest comparator after held-out peak selection. $^{\ast}$: paired $p<0.05$ on per-problem outcomes. $^{\ddagger}$Tokens from the nearest trajectory dump: Eval-Opt-IP-4B, step 220 (peak 230); Eval-Opt-SP-4B, step 215 (peak 200).}
\label{tab:sa_same_structure}
\textbf{Panel A.} Math comparators: accuracy and generated tokens per episode.
\smallskip

\setlength{\tabcolsep}{4pt}
\begin{tabular}{@{}llcccccc@{}}
\toprule
 & & \multicolumn{2}{c}{0.6B} & \multicolumn{2}{c}{1.7B} & \multicolumn{2}{c}{4B} \\
\cmidrule(lr){3-4}\cmidrule(lr){5-6}\cmidrule(l){7-8}
Comparator & Workflow & Acc. & Tokens & Acc. & Tokens & Acc. & Tokens \\
\midrule
S &  & 31.4 & 3.5k & 49.9 & 3.3k & 66.6 & 2.9k \\
M &  & 36.9 & 10.5k & 57.8 & 10.0k & 74.0 & 8.8k \\
R &  & 33.3 & 4.0k & 55.9 & 4.1k & 73.3 & 3.4k \\
L$_{10}$ &  & 33.8 & 3.9k & 57.4 & 4.0k & 74.6 & 3.5k \\
L$_{15}$ &  & 34.6 & 3.9k & 56.4 & 4.1k & 76.3 & 3.6k \\
\addlinespace[2pt]
T$_{\mathrm{a}}$ & Eval-Opt & 32.8 & — & 48.4 & — & 71.9 & — \\
T$_{\mathrm{a}}$ & Voting & 32.1 & — & 50.2 & — & 71.0 & — \\
T$_{\mathrm{a}}$ & Orch-Workers & 37.7 & — & 58.5 & — & 76.1 & — \\
\bottomrule
\end{tabular}

\medskip
\textbf{Panel B.} Math cells minus comparator (points).
\smallskip

\resizebox{\textwidth}{!}{%
\begin{tabular}{@{}lllccccccccc@{}}
\toprule
Scale & Workflow & Policy & Acc. & Tokens & S & M & T$_{\mathrm{a}}$ & R & L$_{10}$ & L$_{15}$ & Held-out \\
\midrule
0.6B & Eval-Opt & \IPa & 39.5 & 14.1k & $+8.2^{\ast}$ & $+2.6^{\ast}$ & $+6.7^{\ast}$ & $+6.2^{\ast}$ & $+5.7^{\ast}$ & $+4.9^{\ast}$ & $+3.2$ (M) \\
 & Eval-Opt & \SPa & 29.5 & 11.5k & $-1.9$ & $-7.4^{\ast}$ & $-3.3^{\ast}$ & $-3.8^{\ast}$ & $-4.3^{\ast}$ & $-5.2^{\ast}$ & $-6.5$ (M) \\
 & Voting & \IPa & 28.5 & 13.0k & $-2.8^{\ast}$ & $-8.4^{\ast}$ & $-3.6^{\ast}$ & $-4.8^{\ast}$ & $-5.2^{\ast}$ & $-6.1^{\ast}$ & $-8.3$ (M) \\
 & Voting & \SPa & 28.3 & 11.6k & $-3.1^{\ast}$ & $-8.6^{\ast}$ & $-3.9^{\ast}$ & $-5.0^{\ast}$ & $-5.5^{\ast}$ & $-6.4^{\ast}$ & $-7.8$ (M) \\
 & Orch-Workers & \IPa & 36.0 (34.4) & 15.0k & $+4.7^{\ast}$ & $-0.8^{\ast}$ & $-1.6^{\ast}$ & $+2.8$ & $+2.3$ & $+1.4$ & $-4.7$ (T$_{\mathrm{a}}$) \\
 & Orch-Workers & \SPa & 31.6 & 15.0k & $+0.2$ & $-5.3^{\ast}$ & $-6.1^{\ast}$ & $-1.7$ & $-2.2^{\ast}$ & $-3.0^{\ast}$ & $-5.4$ (T$_{\mathrm{a}}$) \\
\addlinespace[2pt]
1.7B & Eval-Opt & \IPa & 57.6 (58.8) & 7.5k & $+7.6^{\ast}$ & $-0.2$ & $+9.1^{\ast}$ & $+1.7^{\ast}$ & $+0.1$ & $+1.2^{\ast}$ & $+2.1$ (M) \\
 & Eval-Opt & \SPa & 50.5 (48.2) & 7.0k & $+0.6$ & $-7.3^{\ast}$ & $+2.1$ & $-5.4^{\ast}$ & $-6.9^{\ast}$ & $-5.9^{\ast}$ & $-9.1$ (M) \\
 & Voting & \IPa & 50.9 (51.0) & 10.0k & $+1.0$ & $-6.9^{\ast}$ & $+0.7$ & $-5.0^{\ast}$ & $-6.5^{\ast}$ & $-5.5^{\ast}$ & $-8.2$ (M) \\
 & Voting & \SPa & 46.5 (45.2) & 7.7k & $-3.4^{\ast}$ & $-11.2^{\ast}$ & $-3.7^{\ast}$ & $-9.4^{\ast}$ & $-10.9^{\ast}$ & $-9.9^{\ast}$ & $-14.1$ (M) \\
 & Orch-Workers & \IPa & 55.5 (56.1) & 12.4k & $+5.5^{\ast}$ & $-2.3$ & $-3.0^{\ast}$ & $-0.4$ & $-2.0$ & $-1.0$ & $-2.2$ (T$_{\mathrm{a}}$) \\
 & Orch-Workers & \SPa & 52.3 & 11.1k & $+2.4^{\ast}$ & $-5.4^{\ast}$ & $-6.2^{\ast}$ & $-3.5^{\ast}$ & $-5.1^{\ast}$ & $-4.1^{\ast}$ & $-6.2$ (T$_{\mathrm{a}}$) \\
\addlinespace[2pt]
4B & Eval-Opt & \IPa & 75.2 & 7.6k$^{\ddagger}$ & $+8.6^{\ast}$ & $+1.2$ & $+3.3^{\ast}$ & $+1.9$ & $+0.6$ & $-1.1$ & $-1.1$ (L$_{15}$) \\
 & Eval-Opt & \SPa & 65.8 & 4.8k$^{\ddagger}$ & $-0.8$ & $-8.2^{\ast}$ & $-6.1^{\ast}$ & $-7.5^{\ast}$ & $-8.8^{\ast}$ & $-10.5^{\ast}$ & $-10.5$ (L$_{15}$) \\
 & Voting & \IPa & 68.1 & 8.1k & $+1.6$ & $-5.9^{\ast}$ & $-2.8^{\ast}$ & $-5.2^{\ast}$ & $-6.5^{\ast}$ & $-8.2^{\ast}$ & $-8.2$ (L$_{15}$) \\
 & Voting & \SPa & 59.4 & 6.5k & $-7.2^{\ast}$ & $-14.7^{\ast}$ & $-11.6^{\ast}$ & $-13.9^{\ast}$ & $-15.3^{\ast}$ & $-17.0^{\ast}$ & $-17.4$ (L$_{15}$) \\
 & Orch-Workers & \IPa & 67.9 & 9.2k & $+1.4$ & $-6.1^{\ast}$ & $-8.1^{\ast}$ & $-5.4^{\ast}$ & $-6.7^{\ast}$ & $-8.4^{\ast}$ & $-8.6$ (L$_{15}$) \\
 & Orch-Workers & \SPa & 69.6 & 10.6k & $+3.1^{\ast}$ & $-4.4^{\ast}$ & $-6.4^{\ast}$ & $-3.7^{\ast}$ & $-5.0^{\ast}$ & $-6.7^{\ast}$ & $-6.6$ (L$_{15}$) \\
\bottomrule
\end{tabular}}

\medskip
\textbf{Panel C.} Code cells (1000 problems, re-evaluated) minus comparator (points).
\smallskip

\setlength{\tabcolsep}{4pt}
\begin{tabular}{@{}lllccccc@{}}
\toprule
 & & & & \multicolumn{2}{c}{S} & \multicolumn{2}{c}{T$_{\mathrm{a}}$} \\
\cmidrule(lr){5-6}\cmidrule(l){7-8}
Scale & Workflow & Policy & Acc. & Acc. & Cell $-$ S & Acc. & Cell $-$ T$_{\mathrm{a}}$ \\
\midrule
0.6B & Eval-Opt & \IPa & 15.3 & 11.8 & $+3.5^{\ast}$ & 11.3 & $+4.0^{\ast}$ \\
 & Eval-Opt & \SPa & 13.2 & 11.8 & $+1.4$ & 11.3 & $+1.9^{\ast}$ \\
 & Voting & \IPa & 13.1 & 11.8 & $+1.3$ & 12.5 & $+0.6$ \\
 & Voting & \SPa & 11.7 & 11.8 & $-0.1$ & 12.5 & $-0.8$ \\
 & Orch-Workers & \IPa & 15.9 & 11.8 & $+4.1^{\ast}$ & 14.3 & $+1.6$ \\
 & Orch-Workers & \SPa & 14.6 & 11.8 & $+2.8^{\ast}$ & 14.3 & $+0.3$ \\
\addlinespace[2pt]
1.7B & Eval-Opt & \IPa & 22.7 & 19.4 & $+3.3^{\ast}$ & 19.6 & $+3.1^{\ast}$ \\
 & Eval-Opt & \SPa & 16.2 & 19.4 & $-3.2^{\ast}$ & 19.6 & $-3.4^{\ast}$ \\
 & Voting & \IPa & 18.9 & 19.4 & $-0.5$ & 19.5 & $-0.6$ \\
 & Voting & \SPa & 16.7 & 19.4 & $-2.7^{\ast}$ & 19.5 & $-2.8^{\ast}$ \\
 & Orch-Workers & \IPa & 21.8 & 19.4 & $+2.4$ & 21.7 & $+0.1$ \\
 & Orch-Workers & \SPa & 19.0 & 19.4 & $-0.4$ & 21.7 & $-2.7^{\ast}$ \\
\addlinespace[2pt]
4B & Eval-Opt & \IPa & 30.3 & 33.3 & $-3.0^{\ast}$ & 39.3 & $-9.0^{\ast}$ \\
 & Eval-Opt & \SPa & 38.1 & 33.3 & $+4.8^{\ast}$ & 39.3 & $-1.2$ \\
 & Voting & \IPa & 32.2 & 33.3 & $-1.1$ & 35.8 & $-3.6^{\ast}$ \\
 & Voting & \SPa & 29.7 & 33.3 & $-3.6^{\ast}$ & 35.8 & $-6.1^{\ast}$ \\
 & Orch-Workers & \IPa & 33.6 & 33.3 & $+0.3$ & 40.0 & $-6.4^{\ast}$ \\
 & Orch-Workers & \SPa & 31.3 & 33.3 & $-2.0$ & 40.0 & $-8.7^{\ast}$ \\
\bottomrule
\end{tabular}
\end{table}

\subsection{Measurement Definitions}
\label{sec:appendix:measurements}

The appendix displays share these definitions and the diagnostics of \S\ref{sec:setup:diagnostics}. Accuracy differences are in points, and a dash marks a value that is not defined or not available.

\paragraph{Role-level displays.}
\label{sec:appendix:trajectory_statistics}
Role-level displays grade one episode per problem on the full validation sets of \S\ref{sec:setup}, all 1412 Math and all 1000 Code problems. Every role output is graded on its own answer (\S\ref{sec:setup:diagnostics}), the last boxed answer on Math and the last code block on Code, independently of the episode's reward. SA-RL values are taken at the evaluated or sampled SA-RL step nearest the compared checkpoint, on the same problems. Token counts are completion tokens.

\paragraph{Roles.} Under \IPa the three instances of a parallel role train one adapter and are measured as one role.

\paragraph{Graded outcomes.} The \emph{oracle} holds when at least one solving-role output of an episode is correct. The three instances of an episode \emph{disagree} when one or two of them are correct. An Eval-Opt episode is \emph{lost} when its first attempt is correct and its final answer wrong. A wrong first attempt is \emph{T-wrong} if truncated and \emph{F-wrong} if finished. A role output is \emph{rewarded} when its episode's final answer is correct. On a problem, \emph{no instance solves} when none of the episode's three voters or workers is correct, and \emph{SA-RL solves} when the SA-RL checkpoint of the same scale at the nearest sampled step is correct. Plurality-vote ties are broken at random.

\paragraph{Training diagnostics.} Each logged diagnostic is divided by its median over training steps 1--30; the \emph{late ratio} takes the mean over the last 30 training steps. The logged $\chi^2$, perplexity and gradient norm are measured per adapter (\SPa: the shared adapter); the gradient norm is that of one optimizer step, averaged over a training step. The logged $\chi^2$ grows with the role's per-token entropy, and the logged training perplexity of the adapter's own sampled responses is the same signal (\S\ref{sec:appendix:gradient_amplification_signatures}), so both are treated as one entropy proxy. The \emph{mismatch ratio} (Mism.) is the late ratio of the temperature mismatch behind $\chi^2$, measured before the update and pooled over the batch.

\paragraph{Checkpoint probes.} A probe at step $s$ samples the step-$s$ workflow on the first 200 validation problems at the sampling temperature and keeps these responses fixed. A role's \emph{per-token entropy} is that of its step-$s$ adapter (\SPa: the shared adapter) on its own responses. Its \emph{checkpoint KL}, the change between checkpoints of \S\ref{sec:setup:diagnostics}, is the per-token KL from the step-$s$ adapter to the step-$(s{+}10)$ adapter on the same responses. Both are ratios to the first probe (step 20; Code, step 205).

\paragraph{Parameter change.} $N$ is the number of sequences an adapter receives per training step divided by the fixed mini-batch size. When the division is exact it is the adapter's optimizer steps per training step (\S\ref{sec:appendix:hyperparameters}). Under Eval-Opt, $N$ is the mean number $k$ of generator turns per episode for each \IPa adapter and $2k$ for the shared adapter. The \emph{rate} of an adapter is the distance between its checkpoints 50 training steps apart divided by 50, in units of the learning rate times the square root of its parameter count, averaged over windows that start every 10 training steps from step 20 and end by step 300. The noise model is AdamW driven by independent Gaussian gradients with the same $N$; it has no free parameter and predicts both the rate and the cosine between consecutive 50-step changes.

\paragraph{Tests and horizons.} Paired comparisons of two checkpoints use the two-sided exact McNemar test on per-problem outcomes; bootstrap intervals resample problems. A run \emph{declines} (\S\ref{sec:setup}) when its last checkpoint is below its selected peak with $p<0.05$ and, on Math, the drop holds under \emph{held-out peak selection}: it is significant in at least 95\% of 200 random splits of the 1412 problems into halves, each half scored at the peak step the other half selects. A drop that is significant in sample but fails this test is \emph{in-sample only} and does not count as a decline; every other run is non-declining. Displays of \texttt{Eval-Opt-IP-1.7B-Math} end at step 379 (last sampled checkpoint: step 370), because its later training rewards were computed with an earlier version of the math grader.

\subsection{Compute}
\label{sec:appendix:compute}

All training runs use a single compute node, a frozen base model, LoRA adapters, and a vLLM rollout engine. The frozen model and rollout engine dominate total device memory, while adapter and optimizer states differ with the number of trainable adapters.

Under \IP all role adapters stay in GPU memory, and each rollout request and gradient update uses its role's adapter (resource matching: \S\ref{sec:discussion:limitations}).

Most runs train on two GPUs and some on four. Most 1.7B and 4B cells use H100 GPUs (80\,GB) and most 0.6B and some 1.7B cells use L40S GPUs (48\,GB). The summed wall-clock time of all runs is about 235 node-days.

\section{Additional Results}
\label{sec:appendix:results}

\subsection{Selected Peak and Last Checkpoint per Cell}
\label{sec:appendix:peak_last}

Panel~A of Table~\ref{tab:peak_last} lists every cell's selected peak and last checkpoint at step $\leq 300$. Its After column counts the training steps validated after the selected peak, over the whole validated run on Math and up to step 300 on Code. Runs end at different steps, so the gap between peak and last is compared within a run, not across cells.

\begin{figure}[t]
\centering
\includegraphics[width=\textwidth]{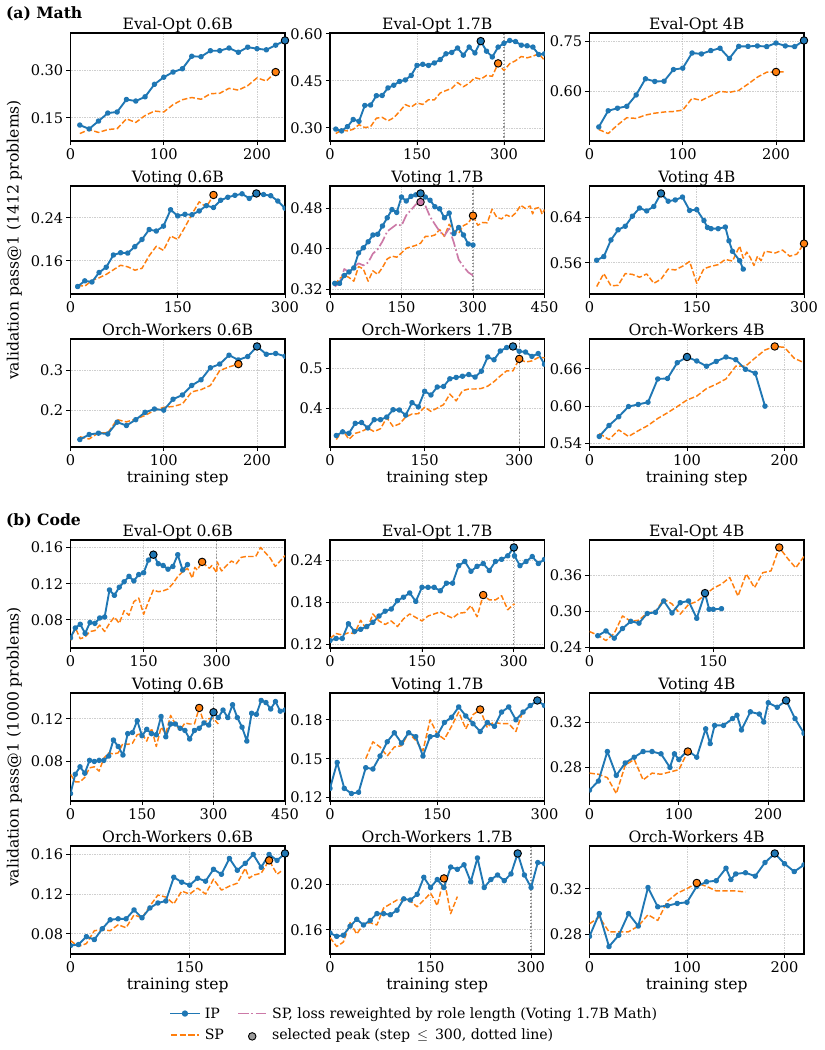}
\caption{Validation accuracy of every matched \IPa/\SPa pair over the full logged horizon: (a) Math, 1412 problems; (b) Code, 1000 problems, the series logged during training (re-evaluated values at the marked steps in Table~\ref{tab:peak_last}). Rows: workflow; columns: scale. Solid: \IPa; dashed: \SPa; dash-dot (Voting 1.7B Math): \SPa with each output's loss reweighted by its role's mean length (Table~\ref{tab:controls}). Filled circles: selected peak (step $\leq 300$, dotted line). Series start at the first validated step.}
\label{fig:ip_sp_curves}
\end{figure}

Panel~B compares \IPa with \SPa at the selected peaks of Panel~A and at the last step validated in both runs, and each strategy with SA-RL at the selected peaks. Where the \IPa run declines, \IPa minus \SPa shifts in favor of \SPa by the last common step (\S\ref{sec:results:ip_vs_sp}). On Code, \IPa and \SPa are also compared at a common step cap (Table~\ref{tab:code_task}).
Fig.~\ref{fig:ip_sp_curves} shows each pair's validation series over its full horizon, with the steps of the selected peaks of Panel~A marked; most \SPa Math runs end at or just after their selected peak (Panel~A of Table~\ref{tab:peak_last}, After column).

In Fig.~\ref{fig:ip_vs_sp}(b), a pair's common gain is the smaller of the two runs' validation gains within their shared horizon. Gains are measured on 3-point smoothed validation series, from the untrained workflow on Math and from the run's step-0 validation on Code (for \texttt{Eval-Opt-IP-4B-Code}, that of the \SPa run, which starts from the same untrained workflow).

\begin{table}[t]
\centering
\small
\caption{Validation accuracy (\%) at the selected peak and the last validated checkpoint at step $\leq 300$ (step in parentheses), with the training steps validated after the peak (After) (Panel~A), and differences in points under the horizon rules of the text (Panel~B). Math: off-policy validator (1412 problems); Code: re-evaluation (1000 problems) of the checkpoints selected on the validation series logged during training, SA-RL included; \textemdash: no re-evaluated checkpoint.}
\label{tab:peak_last}
\textbf{Panel A.} Selected peak and last checkpoint at step $\leq 300$; training steps validated after the selected peak.
\smallskip

\resizebox{\textwidth}{!}{%
\begin{tabular}{@{}lllcccccc@{}}
\toprule
 & & & \multicolumn{3}{c}{\IPa} & \multicolumn{3}{c}{\SPa} \\
\cmidrule(lr){4-6}\cmidrule(l){7-9}
Task & Scale & Workflow & Peak (step) & Last (step) & After & Peak (step) & Last (step) & After \\
\midrule
\multirow{9}{*}{Math} & \multirow{3}{*}{0.6B} & Eval-Opt & 39.5 (230) & 39.5 (230) & 0 & 29.5 (220) & 29.5 (220) & 0 \\
 &  & Voting & 28.5 (260) & 25.8 (300) & 40 & 28.3 (200) & 28.3 (200) & 0 \\
 &  & Orch-Workers & 36.0 (200) & 33.6 (230) & 30 & 31.6 (180) & 31.6 (180) & 0 \\
\addlinespace[2pt]
 & \multirow{3}{*}{1.7B} & Eval-Opt & 57.6 (260) & 56.9 (300) & 110 & 50.5 (290) & 48.2 (300) & 80 \\
 &  & Voting & 50.9 (190) & 40.7 (300) & 110 & 46.5 (300) & 46.5 (300) & 150 \\
 &  & Orch-Workers & 55.5 (290) & 54.2 (300) & 50 & 52.3 (300) & 52.3 (300) & 40 \\
\addlinespace[2pt]
 & \multirow{3}{*}{4B} & Eval-Opt & 75.2 (230) & 75.2 (230) & 0 & 65.8 (200) & 65.8 (210) & 10 \\
 &  & Voting & 68.1 (100) & 54.9 (215) & 115 & 59.4 (300) & 59.4 (300) & 0 \\
 &  & Orch-Workers & 67.9 (100) & 60.0 (180) & 80 & 69.6 (190) & 67.0 (220) & 30 \\
\midrule
\multirow{9}{*}{Code} & \multirow{3}{*}{0.6B} & Eval-Opt & 15.3 (170) & 14.6 (240) & 70 & 13.2 (270) & 13.4 (300) & 30 \\
 &  & Voting & 13.1 (300) & 13.1 (300) & 0 & 11.7 (270) & 12.9 (300) & 30 \\
 &  & Orch-Workers & 15.9 (270) & 15.9 (270) & 0 & 14.6 (250) & 13.7 (265) & 15 \\
\addlinespace[2pt]
 & \multirow{3}{*}{1.7B} & Eval-Opt & 22.7 (300) & 22.7 (300) & 0 & 16.2 (250) & 18.5 (300) & 50 \\
 &  & Voting & 18.9 (290) & 18.9 (300) & 10 & 16.7 (210) & 18.0 (270) & 60 \\
 &  & Orch-Workers & 21.8 (280) & 20.1 (300) & 20 & 19.0 (170) & 18.9 (190) & 20 \\
\addlinespace[2pt]
 & \multirow{3}{*}{4B} & Eval-Opt & 30.3 (140) & 28.4 (160) & 20 & 38.1 (230) & 39.4 (260) & 30 \\
 &  & Voting & 32.2 (220) & 33.4 (240) & 20 & 29.7 (110) & 29.7 (110) & 0 \\
 &  & Orch-Workers & 33.6 (190) & 33.8 (220) & 30 & 31.3 (110) & 32.0 (160) & 50 \\
\bottomrule
\end{tabular}%
}

\bigskip

\textbf{Panel B.} \IPa minus \SPa at the selected peak and at the last step validated in both runs, and each strategy minus SA-RL at the selected peak (points).
\smallskip

\begin{tabular}{@{}lllcccc@{}}
\toprule
 & & & \multicolumn{2}{c}{\IPa $-$ \SPa} & \IPa $-$ SA-RL & \SPa $-$ SA-RL \\
\cmidrule(lr){4-5}\cmidrule(lr){6-6}\cmidrule(l){7-7}
Task & Scale & Workflow & Peak & Last common & Peak & Peak \\
\midrule
\multirow{9}{*}{Math} & \multirow{3}{*}{0.6B} & Eval-Opt & $+10.1$ & $+8.6$ (220) & $+8.2$ & $-1.9$ \\
 &  & Voting & $+0.3$ & $-2.3$ (200) & $-2.8$ & $-3.1$ \\
 &  & Orch-Workers & $+4.5$ & $+1.1$ (180) & $+4.7$ & $+0.2$ \\
\addlinespace[2pt]
 & \multirow{3}{*}{1.7B} & Eval-Opt & $+7.1$ & $+2.0$ (370) & $+7.6$ & $+0.6$ \\
 &  & Voting & $+4.4$ & $-5.8$ (300) & $+1.0$ & $-3.4$ \\
 &  & Orch-Workers & $+3.1$ & $-1.6$ (340) & $+5.5$ & $+2.4$ \\
\addlinespace[2pt]
 & \multirow{3}{*}{4B} & Eval-Opt & $+9.4$ & $+7.8$ (210) & $+8.6$ & $-0.8$ \\
 &  & Voting & $+8.8$ & $-1.9$ (215) & $+1.6$ & $-7.2$ \\
 &  & Orch-Workers & $-1.7$ & $-9.1$ (180) & $+1.4$ & $+3.1$ \\
\midrule
\multirow{9}{*}{Code} & \multirow{3}{*}{0.6B} & Eval-Opt & $+2.1$ & \textemdash & $+3.5$ & $+1.4$ \\
 &  & Voting & $+1.4$ & \textemdash & $+1.3$ & $-0.1$ \\
 &  & Orch-Workers & $+1.3$ & \textemdash & $+4.1$ & $+2.8$ \\
\addlinespace[2pt]
 & \multirow{3}{*}{1.7B} & Eval-Opt & $+6.5$ & \textemdash & $+3.3$ & $-3.2$ \\
 &  & Voting & $+2.2$ & \textemdash & $-0.5$ & $-2.7$ \\
 &  & Orch-Workers & $+2.8$ & \textemdash & $+2.4$ & $-0.4$ \\
\addlinespace[2pt]
 & \multirow{3}{*}{4B} & Eval-Opt & $-7.8$ & \textemdash & $-3.0$ & $+4.8$ \\
 &  & Voting & $+2.5$ & \textemdash & $-1.1$ & $-3.6$ \\
 &  & Orch-Workers & $+2.3$ & \textemdash & $+0.3$ & $-2.0$ \\
\bottomrule
\end{tabular}%
\end{table}

\paragraph{Post-peak declines.} The runs of Fig.~\ref{fig:ip_decline} are tested from the selected peak to the last checkpoint of the validated run (\S\ref{sec:appendix:measurements}). Voting and Orch-Workers under \IPa at 1.7B and 4B, and \texttt{Eval-Opt-IP-1.7B-Math} from step 260 to 370, decline with $p<10^{-4}$, also under held-out peak selection. \texttt{Voting-IP-0.6B-Math} falls with $p=0.009$ in sample only, so it does not count as a decline, and \texttt{Orch-Workers-IP-0.6B-Math} does not fall ($p=0.52$).

\paragraph{Post-peak windows of the shared-policy runs.} At matched offsets of 10, 30, 50 and 80 training steps after the selected peak, the five \IPa runs that decline under held-out peak selection fall significantly (the test of \S\ref{sec:appendix:measurements}; Fig.~\ref{fig:ip_sp_curves}). \texttt{Voting-IP-1.7B-Math} and \texttt{Orch-Workers-IP-1.7B-Math} fall from the 30-step offset on, \texttt{Voting-IP-4B-Math} from the 50-step offset on, \texttt{Orch-Workers-IP-4B-Math} at the 80-step offset, and \texttt{Eval-Opt-IP-1.7B-Math} at the 10-, 30- and 80-step offsets. Among the \SPa runs, only \texttt{Orch-Workers-SP-4B-Math} falls significantly, by 2.6 points at the 30-step offset, and the three 1.7B \SPa runs fall at no offset. Only two \SPa runs, \texttt{Eval-Opt-SP-1.7B-Math} and \texttt{Voting-SP-1.7B-Math}, are validated at the longer offsets, so most \SPa runs are observed too briefly after their peak for their post-peak change to be compared with that of the \IPa runs.

\subsection{Solving Roles Against Single-Agent RL}
\label{sec:appendix:solving_role_vs_sa}

Table~\ref{tab:solving_role_vs_sa} gives the per-checkpoint values for \S\ref{sec:results:landscape}. Under Eval-Opt the oracle counts a correct answer at any generator turn. The rewarded share of truncated outputs is shown only for rows with at least five truncated outputs. In the SA-RL rows the reward is the grade of the policy's own answer. Since every role receives the episode's outcome, Rew.\ wrong equals Struct.\ plus the share of correct solving-role outputs in unrewarded episodes, so it measures no separate effect.

Fig.~\ref{fig:reward_broadcast}(a) uses the checkpoints marked $\dagger$. Against the untrained row on the same problems, the change in Solver is the solving-role gain and the change in Struct.\ the structural gain; the two sum to the change in Workflow.

In Fig.~\ref{fig:reward_broadcast}(c), each point is the mean voter accuracy of \texttt{Voting-IP-1.7B-Math} or of a control (Table~\ref{tab:controls}) at a sampled checkpoint on all 1412 problems, minus SA-RL's validation accuracy at 1.7B at the nearest evaluated step, without pairing; for the single-role control it is the accuracy of its one output. In Fig.~\ref{fig:reward_broadcast}(d), each point is the Voting accuracy of the same checkpoint on the same problems, minus the same SA-RL value.

\begin{table}[p]
\centering
\caption{Solving roles trained inside each workflow against SA-RL on Math, per sampled checkpoint, on all 1412 problems; fractions. Solver: accuracy on the role's own answer; SA-RL: single-agent accuracy on the same problems (step); Diff.: Solver minus SA-RL; Struct.: Workflow minus Solver; Rew.\ wrong, Rew.\,$\mid$\,wrong: rewarded-but-wrong outputs over all and over wrong outputs; Trunc.: truncated share; Rew.\,$\mid$\,trunc.: rewarded share of truncated outputs. Base: untrained workflow; $\dagger$: checkpoint of Fig.~\ref{fig:reward_broadcast}(a).}
\label{tab:solving_role_vs_sa}
\setlength{\tabcolsep}{3pt}
\resizebox{\textwidth}{!}{%
\begin{tabular}{@{}c@{\hspace{10pt}}c@{}}
\begin{tabular}[t]{@{}llrcccccccccc@{}}
\toprule
Scale & Strategy & Step & Solver & \begin{tabular}[b]{@{}c@{}}SA-RL\\(step)\end{tabular} & \begin{tabular}[b]{@{}c@{}}Diff.\\(pp)\end{tabular} & Oracle & Workflow & Struct. & \begin{tabular}[b]{@{}c@{}}Rew.\\wrong\end{tabular} & \begin{tabular}[b]{@{}c@{}}Rew.\,$\mid$\\wrong\end{tabular} & Trunc. & \begin{tabular}[b]{@{}c@{}}Rew.\,$\mid$\\trunc.\end{tabular} \\
\midrule
\multicolumn{13}{@{}l}{\textit{Voting: mean voter}} \\
0.6B & Base & 0 & $0.10$ & -- & -- & $0.20$ & $0.10$ & $-0.00$ & $0.04$ & $0.04$ & $0.00$ & $0.00$ \\
 & \IPa & 100 & $0.18$ & $0.23$\,(100) & $-5$ & $0.33$ & $0.19$ & $0.01$ & $0.07$ & $0.08$ & $0.16$ & $0.10$ \\
 & \IPa & 200 & $0.20$ & $0.26$\,(200) & $-6$ & $0.31$ & $0.25$ & $0.05$ & $0.08$ & $0.09$ & $0.63$ & $0.08$ \\
 & \IPa & 260$^{\dagger}$ & $0.21$ & $0.29$\,(260) & $-7$ & $0.33$ & $0.29$ & $0.08$ & $0.09$ & $0.12$ & $0.59$ & $0.11$ \\
 & \IPa & 300 & $0.20$ & $0.30$\,(300) & $-10$ & $0.32$ & $0.26$ & $0.07$ & $0.09$ & $0.11$ & $0.44$ & $0.12$ \\
 & \SPa & 50 & $0.12$ & $0.14$\,(50) & $-2$ & $0.24$ & $0.13$ & $0.01$ & $0.05$ & $0.06$ & $0.00$ & $0.00$ \\
 & \SPa & 90 & $0.14$ & $0.21$\,(90) & $-7$ & $0.28$ & $0.16$ & $0.01$ & $0.06$ & $0.07$ & $0.02$ & $0.06$ \\
 & \SPa & 200$^{\dagger}$ & $0.22$ & $0.26$\,(200) & $-4$ & $0.36$ & $0.28$ & $0.06$ & $0.09$ & $0.11$ & $0.47$ & $0.09$ \\
\addlinespace[2pt]
1.7B & Base & 0 & $0.27$ & -- & -- & $0.44$ & $0.28$ & $0.01$ & $0.08$ & $0.10$ & $0.03$ & $0.10$ \\
 & \IPa & 70 & $0.33$ & $0.33$\,(70) & $0$ & $0.50$ & $0.40$ & $0.07$ & $0.11$ & $0.16$ & $0.09$ & $0.19$ \\
 & \IPa & 130 & $0.39$ & $0.41$\,(130) & $-1$ & $0.57$ & $0.46$ & $0.07$ & $0.11$ & $0.19$ & $0.21$ & $0.17$ \\
 & \IPa & 190$^{\dagger}$ & $0.43$ & $0.45$\,(190) & $-2$ & $0.59$ & $0.51$ & $0.08$ & $0.11$ & $0.19$ & $0.32$ & $0.19$ \\
 & \IPa & 290 & $0.33$ & $0.50$\,(290) & $-17$ & $0.48$ & $0.40$ & $0.07$ & $0.10$ & $0.15$ & $0.32$ & $0.15$ \\
 & \IPa & 300 & $0.32$ & $0.50$\,(300) & $-17$ & $0.46$ & $0.40$ & $0.08$ & $0.10$ & $0.15$ & $0.47$ & $0.14$ \\
 & \SPa & 120 & $0.32$ & $0.40$\,(120) & $-8$ & $0.49$ & $0.39$ & $0.07$ & $0.11$ & $0.16$ & $0.07$ & $0.16$ \\
 & \SPa & 230 & $0.37$ & $0.45$\,(230) & $-8$ & $0.55$ & $0.43$ & $0.06$ & $0.10$ & $0.16$ & $0.11$ & $0.19$ \\
 & \SPa & 250 & $0.36$ & $0.47$\,(250) & $-11$ & $0.52$ & $0.42$ & $0.06$ & $0.10$ & $0.15$ & $0.13$ & $0.15$ \\
 & \SPa & 300$^{\dagger}$ & $0.38$ & $0.50$\,(300) & $-12$ & $0.55$ & $0.45$ & $0.07$ & $0.11$ & $0.17$ & $0.15$ & $0.17$ \\
 & \SPa & 450 & $0.40$ & $0.53$\,(430) & $-14$ & $0.55$ & $0.48$ & $0.08$ & $0.11$ & $0.18$ & $0.23$ & $0.19$ \\
\addlinespace[2pt]
4B & Base & 0 & $0.47$ & -- & -- & $0.64$ & $0.52$ & $0.05$ & $0.10$ & $0.18$ & $0.04$ & $0.21$ \\
 & \IPa & 70 & $0.56$ & $0.57$\,(70) & $-1$ & $0.72$ & $0.65$ & $0.09$ & $0.11$ & $0.26$ & $0.15$ & $0.28$ \\
 & \IPa & 100$^{\dagger}$ & $0.58$ & $0.60$\,(100) & $-2$ & $0.73$ & $0.66$ & $0.08$ & $0.11$ & $0.25$ & $0.20$ & $0.27$ \\
 & \IPa & 130 & $0.59$ & $0.62$\,(130) & $-3$ & $0.73$ & $0.68$ & $0.09$ & $0.11$ & $0.27$ & $0.21$ & $0.25$ \\
 & \IPa & 195 & $0.50$ & $0.63$\,(190) & $-14$ & $0.66$ & $0.60$ & $0.10$ & $0.12$ & $0.25$ & $0.22$ & $0.25$ \\
 & \IPa & 200 & $0.49$ & $0.65$\,(200) & $-16$ & $0.65$ & $0.58$ & $0.09$ & $0.11$ & $0.22$ & $0.19$ & $0.20$ \\
 & \IPa & 215 & $0.45$ & $0.63$\,(210) & $-18$ & $0.61$ & $0.55$ & $0.10$ & $0.12$ & $0.22$ & $0.29$ & $0.21$ \\
 & \SPa & 70 & $0.47$ & $0.57$\,(70) & $-10$ & $0.64$ & $0.53$ & $0.06$ & $0.11$ & $0.20$ & $0.04$ & $0.22$ \\
 & \SPa & 140 & $0.48$ & $0.62$\,(140) & $-14$ & $0.65$ & $0.56$ & $0.08$ & $0.11$ & $0.21$ & $0.05$ & $0.22$ \\
 & \SPa & 170 & $0.49$ & $0.64$\,(170) & $-15$ & $0.65$ & $0.55$ & $0.06$ & $0.10$ & $0.20$ & $0.05$ & $0.19$ \\
 & \SPa & 300$^{\dagger}$ & $0.52$ & $0.67$\,(300) & $-15$ & $0.69$ & $0.59$ & $0.07$ & $0.11$ & $0.23$ & $0.07$ & $0.26$ \\
\midrule
\multicolumn{13}{@{}l}{\textit{Orch-Workers: mean worker}} \\
0.6B & Base & 0 & $0.10$ & -- & -- & $0.22$ & $0.12$ & $0.02$ & $0.06$ & $0.06$ & $0.00$ & $0.13$ \\
 & \IPa & 90 & $0.15$ & $0.21$\,(90) & $-6$ & $0.29$ & $0.19$ & $0.04$ & $0.08$ & $0.09$ & $0.06$ & $0.08$ \\
 & \IPa & 180 & $0.18$ & $0.26$\,(180) & $-8$ & $0.31$ & $0.32$ & $0.13$ & $0.15$ & $0.18$ & $0.62$ & $0.18$ \\
 & \IPa & 190 & $0.17$ & $0.26$\,(190) & $-9$ & $0.28$ & $0.33$ & $0.16$ & $0.16$ & $0.20$ & $0.66$ & $0.21$ \\
 & \IPa & 200$^{\dagger}$ & $0.16$ & $0.26$\,(200) & $-10$ & $0.26$ & $0.34$ & $0.18$ & $0.19$ & $0.22$ & $0.73$ & $0.22$ \\
 & \SPa & 40 & $0.12$ & $0.14$\,(40) & $-2$ & $0.25$ & $0.17$ & $0.05$ & $0.08$ & $0.09$ & $0.02$ & $0.04$ \\
 & \SPa & 70 & $0.14$ & $0.18$\,(70) & $-5$ & $0.27$ & $0.17$ & $0.04$ & $0.07$ & $0.09$ & $0.04$ & $0.13$ \\
 & \SPa & 110 & $0.17$ & $0.22$\,(110) & $-5$ & $0.32$ & $0.22$ & $0.05$ & $0.09$ & $0.11$ & $0.10$ & $0.16$ \\
 & \SPa & 180$^{\dagger}$ & $0.16$ & $0.26$\,(180) & $-10$ & $0.26$ & $0.30$ & $0.14$ & $0.15$ & $0.18$ & $0.70$ & $0.18$ \\
\addlinespace[2pt]
1.7B & Base & 0 & $0.26$ & -- & -- & $0.44$ & $0.34$ & $0.08$ & $0.12$ & $0.16$ & $0.03$ & $0.17$ \\
 & \IPa & 140 & $0.34$ & $0.40$\,(140) & $-5$ & $0.53$ & $0.42$ & $0.08$ & $0.12$ & $0.18$ & $0.11$ & $0.19$ \\
 & \IPa & 280 & $0.40$ & $0.49$\,(280) & $-9$ & $0.55$ & $0.55$ & $0.16$ & $0.17$ & $0.28$ & $0.42$ & $0.31$ \\
 & \IPa & 290$^{\dagger}$ & $0.37$ & $0.50$\,(290) & $-13$ & $0.51$ & $0.56$ & $0.19$ & $0.20$ & $0.32$ & $0.49$ & $0.34$ \\
 & \IPa & 330 & $0.29$ & $0.52$\,(330) & $-23$ & $0.40$ & $0.53$ & $0.24$ & $0.24$ & $0.34$ & $0.64$ & $0.35$ \\
 & \SPa & 210 & $0.32$ & $0.47$\,(210) & $-15$ & $0.49$ & $0.42$ & $0.11$ & $0.13$ & $0.19$ & $0.16$ & $0.17$ \\
 & \SPa & 300$^{\dagger}$ & $0.36$ & $0.50$\,(300) & $-14$ & $0.53$ & $0.51$ & $0.15$ & $0.17$ & $0.27$ & $0.40$ & $0.28$ \\
 & \SPa & 330 & $0.36$ & $0.52$\,(330) & $-16$ & $0.53$ & $0.53$ & $0.17$ & $0.18$ & $0.29$ & $0.45$ & $0.30$ \\
 & \SPa & 340 & $0.34$ & $0.54$\,(340) & $-20$ & $0.50$ & $0.53$ & $0.19$ & $0.20$ & $0.30$ & $0.50$ & $0.33$ \\
\bottomrule
\end{tabular}
&
\begin{tabular}[t]{@{}llrcccccccccc@{}}
\toprule
Scale & Strategy & Step & Solver & \begin{tabular}[b]{@{}c@{}}SA-RL\\(step)\end{tabular} & \begin{tabular}[b]{@{}c@{}}Diff.\\(pp)\end{tabular} & Oracle & Workflow & Struct. & \begin{tabular}[b]{@{}c@{}}Rew.\\wrong\end{tabular} & \begin{tabular}[b]{@{}c@{}}Rew.\,$\mid$\\wrong\end{tabular} & Trunc. & \begin{tabular}[b]{@{}c@{}}Rew.\,$\mid$\\trunc.\end{tabular} \\
\midrule
\multicolumn{13}{@{}l}{\textit{Orch-Workers: mean worker}} \\
4B & Base & 0 & $0.45$ & -- & -- & $0.63$ & $0.54$ & $0.09$ & $0.13$ & $0.23$ & $0.04$ & $0.23$ \\
 & \IPa & 60 & $0.51$ & $0.57$\,(60) & $-6$ & $0.69$ & $0.61$ & $0.10$ & $0.13$ & $0.27$ & $0.08$ & $0.26$ \\
 & \IPa & 100$^{\dagger}$ & $0.56$ & $0.60$\,(100) & $-4$ & $0.73$ & $0.68$ & $0.12$ & $0.14$ & $0.32$ & $0.20$ & $0.36$ \\
 & \IPa & 120 & $0.54$ & $0.62$\,(120) & $-8$ & $0.69$ & $0.66$ & $0.12$ & $0.14$ & $0.31$ & $0.29$ & $0.34$ \\
 & \IPa & 180 & $0.34$ & $0.64$\,(180) & $-30$ & $0.47$ & $0.61$ & $0.27$ & $0.27$ & $0.42$ & $0.62$ & $0.43$ \\
 & \SPa & 70 & $0.47$ & $0.57$\,(70) & $-10$ & $0.66$ & $0.57$ & $0.10$ & $0.13$ & $0.25$ & $0.06$ & $0.27$ \\
 & \SPa & 130 & $0.52$ & $0.62$\,(130) & $-9$ & $0.71$ & $0.64$ & $0.12$ & $0.15$ & $0.30$ & $0.13$ & $0.30$ \\
 & \SPa & 170 & $0.52$ & $0.64$\,(170) & $-12$ & $0.70$ & $0.67$ & $0.15$ & $0.17$ & $0.35$ & $0.25$ & $0.34$ \\
 & \SPa & 190$^{\dagger}$ & $0.52$ & $0.63$\,(190) & $-11$ & $0.68$ & $0.70$ & $0.18$ & $0.19$ & $0.40$ & $0.33$ & $0.40$ \\
 & \SPa & 220 & $0.45$ & $0.64$\,(220) & $-19$ & $0.60$ & $0.68$ & $0.23$ & $0.23$ & $0.42$ & $0.47$ & $0.43$ \\
\midrule
\multicolumn{13}{@{}l}{\textit{Eval-Opt: first attempt}} \\
0.6B & Base & 0 & $0.10$ & -- & -- & $0.11$ & $0.10$ & $0.00$ & $0.00$ & $0.00$ & $0.00$ & $0.00$ \\
 & \IPa & 100 & $0.21$ & $0.23$\,(100) & $-1$ & $0.30$ & $0.29$ & $0.08$ & $0.08$ & $0.10$ & $0.51$ & $0.14$ \\
 & \IPa & 190 & $0.14$ & $0.26$\,(190) & $-12$ & $0.37$ & $0.36$ & $0.22$ & $0.22$ & $0.26$ & $0.82$ & $0.27$ \\
 & \IPa & 200 & $0.12$ & $0.26$\,(200) & $-14$ & $0.37$ & $0.37$ & $0.25$ & $0.25$ & $0.28$ & $0.86$ & $0.30$ \\
 & \IPa & 230$^{\dagger}$ & $0.10$ & $0.27$\,(230) & $-17$ & $0.39$ & $0.39$ & $0.29$ & $0.29$ & $0.32$ & $0.90$ & $0.33$ \\
 & \SPa & 110 & $0.17$ & $0.22$\,(110) & $-5$ & $0.19$ & $0.18$ & $0.01$ & $0.01$ & $0.02$ & $0.14$ & $0.06$ \\
 & \SPa & 210 & $0.18$ & $0.27$\,(210) & $-8$ & $0.29$ & $0.29$ & $0.10$ & $0.10$ & $0.13$ & $0.65$ & $0.16$ \\
 & \SPa & 220$^{\dagger}$ & $0.17$ & $0.28$\,(220) & $-11$ & $0.29$ & $0.29$ & $0.12$ & $0.12$ & $0.15$ & $0.72$ & $0.18$ \\
\addlinespace[2pt]
1.7B & Base & 0 & $0.27$ & -- & -- & $0.27$ & $0.27$ & $0.00$ & $0.00$ & $0.00$ & $0.03$ & $0.00$ \\
 & \IPa & 160 & $0.43$ & $0.42$\,(160) & $+1$ & $0.51$ & $0.50$ & $0.07$ & $0.07$ & $0.13$ & $0.38$ & $0.19$ \\
 & \IPa & 260$^{\dagger}$ & $0.38$ & $0.48$\,(260) & $-10$ & $0.59$ & $0.59$ & $0.21$ & $0.21$ & $0.33$ & $0.58$ & $0.37$ \\
 & \IPa & 300 & $0.43$ & $0.50$\,(300) & $-7$ & $0.58$ & $0.58$ & $0.14$ & $0.14$ & $0.25$ & $0.49$ & $0.30$ \\
 & \IPa & 320 & $0.41$ & $0.50$\,(320) & $-10$ & $0.56$ & $0.55$ & $0.15$ & $0.15$ & $0.25$ & $0.52$ & $0.29$ \\
 & \IPa & 340 & $0.42$ & $0.54$\,(340) & $-12$ & $0.57$ & $0.57$ & $0.15$ & $0.15$ & $0.25$ & $0.52$ & $0.30$ \\
 & \IPa & 360 & $0.39$ & $0.51$\,(360) & $-12$ & $0.53$ & $0.53$ & $0.13$ & $0.13$ & $0.22$ & $0.56$ & $0.25$ \\
 & \IPa & 370 & $0.39$ & $0.53$\,(370) & $-13$ & $0.55$ & $0.55$ & $0.15$ & $0.15$ & $0.25$ & $0.54$ & $0.29$ \\
 & \SPa & 150 & $0.36$ & $0.41$\,(150) & $-5$ & $0.40$ & $0.38$ & $0.02$ & $0.03$ & $0.05$ & $0.20$ & $0.09$ \\
 & \SPa & 290$^{\dagger}$ & $0.40$ & $0.50$\,(290) & $-10$ & $0.49$ & $0.48$ & $0.09$ & $0.09$ & $0.14$ & $0.42$ & $0.20$ \\
 & \SPa & 320 & $0.39$ & $0.50$\,(320) & $-11$ & $0.56$ & $0.54$ & $0.15$ & $0.16$ & $0.25$ & $0.51$ & $0.31$ \\
\addlinespace[2pt]
4B & Base & 0 & $0.47$ & -- & -- & $0.52$ & $0.48$ & $0.01$ & $0.03$ & $0.06$ & $0.04$ & $0.10$ \\
 & \IPa & 90 & $0.58$ & $0.61$\,(90) & $-2$ & $0.65$ & $0.64$ & $0.06$ & $0.06$ & $0.14$ & $0.26$ & $0.23$ \\
 & \IPa & 180 & $0.49$ & $0.64$\,(180) & $-15$ & $0.75$ & $0.74$ & $0.26$ & $0.26$ & $0.50$ & $0.49$ & $0.53$ \\
 & \IPa & 220$^{\dagger}$ & $0.40$ & $0.64$\,(220) & $-24$ & $0.76$ & $0.75$ & $0.35$ & $0.36$ & $0.59$ & $0.60$ & $0.60$ \\
 & \SPa & 105 & $0.54$ & $0.60$\,(100) & $-6$ & $0.60$ & $0.56$ & $0.02$ & $0.04$ & $0.09$ & $0.12$ & $0.13$ \\
 & \SPa & 210$^{\dagger}$ & $0.61$ & $0.63$\,(210) & $-2$ & $0.69$ & $0.68$ & $0.08$ & $0.08$ & $0.21$ & $0.26$ & $0.29$ \\
 & \SPa & 215 & $0.59$ & $0.63$\,(210) & $-3$ & $0.68$ & $0.66$ & $0.07$ & $0.08$ & $0.19$ & $0.27$ & $0.26$ \\
\midrule
\multicolumn{13}{@{}l}{\textit{SA-RL: the single output}} \\
0.6B & SA-RL & 100 & $0.21$ & -- & -- & -- & -- & -- & $0.00$ & $0.00$ & $0.45$ & $0.00$ \\
 & SA-RL & 200 & $0.27$ & -- & -- & -- & -- & -- & $0.00$ & $0.00$ & $0.46$ & $0.00$ \\
 & SA-RL & 300 & $0.31$ & -- & -- & -- & -- & -- & $0.00$ & $0.00$ & $0.32$ & $0.01$ \\
\addlinespace[2pt]
1.7B & SA-RL & 100 & $0.38$ & -- & -- & -- & -- & -- & $0.00$ & $0.00$ & $0.26$ & $0.00$ \\
 & SA-RL & 200 & $0.45$ & -- & -- & -- & -- & -- & $0.00$ & $0.00$ & $0.35$ & $0.00$ \\
 & SA-RL & 300 & $0.50$ & -- & -- & -- & -- & -- & $0.00$ & $0.00$ & $0.29$ & $0.01$ \\
 & SA-RL & 430 & $0.52$ & -- & -- & -- & -- & -- & $0.00$ & $0.00$ & $0.16$ & $0.00$ \\
\addlinespace[2pt]
4B & SA-RL & 100 & $0.62$ & -- & -- & -- & -- & -- & $0.00$ & $0.00$ & $0.24$ & $0.01$ \\
 & SA-RL & 200 & $0.63$ & -- & -- & -- & -- & -- & $0.00$ & $0.00$ & $0.24$ & $0.00$ \\
 & SA-RL & 300 & $0.67$ & -- & -- & -- & -- & -- & $0.00$ & $0.00$ & $0.22$ & $0.00$ \\
\bottomrule
\end{tabular}
\end{tabular}%
}
\end{table}

\subsection{Decision-Role Rules in Full}
\label{sec:appendix:sp_role_capture_signatures}

Table~\ref{tab:decision_role_rules_full} adds the untrained workflow and the remaining columns to Table~\ref{tab:decision_role_rules}. $\Delta$ vs.\ plurality is workflow accuracy minus the expected plurality-vote accuracy over the voters' boxed answers. Lone correct picked counts the episodes with one correct voter and two voters sharing a wrong answer in which the aggregator selects the correct one; Follows 2--1 majority counts the episodes with a two-to-one split in which it selects a majority voter. Answer-only is the share of synthesizer outputs of at most 40 tokens, and Equals plurality counts, among answer-only outputs whose workers have a unique most common answer, those that return it. Loss and Rescue are the shares of episodes with a selection loss and with a rescue (\S\ref{sec:setup:diagnostics}). Against the untrained workflow, the trained aggregator follows a two-to-one majority more often, and the synthesizer loses fewer worker answers and rescues more episodes.

\begin{table}[t]
\centering
\caption{Decision-role behavior on Math, full version of Table~\ref{tab:decision_role_rules}: the untrained workflow (Base) and the latest sampled \IPa and \SPa checkpoints (steps in the last column), on all 1412 validation problems. $k/n$: count and denominator; --: empty denominator. Voting: workflow minus plurality-vote accuracy and the aggregator's selections. Eval-Opt: the evaluator's first verdict. Orch-Workers: answer-only synthesizer outputs and episode shares. Column definitions: \S\ref{sec:appendix:sp_role_capture_signatures}.}
\label{tab:decision_role_rules_full}
\footnotesize
\setlength{\tabcolsep}{2.5pt}
\begin{tabular*}{\textwidth}{@{\extracolsep{\fill}}l ccc ccc ccc r@{}}
\toprule
\multicolumn{11}{@{}l}{\textit{Voting aggregator}} \\
 & \multicolumn{3}{c}{$\Delta$ vs.\ plurality (pp)} & \multicolumn{3}{c}{Lone correct picked} & \multicolumn{3}{c}{Follows 2--1 majority} & \multicolumn{1}{r@{}}{Steps} \\
\cmidrule(lr){2-4}\cmidrule(lr){5-7}\cmidrule(lr){8-10}
Scale & Base & \IPa & \SPa & Base & \IPa & \SPa & Base & \IPa & \SPa & \IPa{} / \SPa \\
\midrule
0.6B & ${-}1.5$ & ${+}0.4$ & ${-}0.1$ & $11/28$ & $0/15$ & $2/25$ & $228/335$ & $153/161$ & $169/179$ & 300 / 200 \\
1.7B & ${-}2.4$ & ${+}0.5$ & ${+}0.4$ & $16/56$ & $4/21$ & $3/25$ & $304/414$ & $177/199$ & $235/263$ & 300 / 450 \\
4B & ${+}0.3$ & ${+}3.1$ & ${+}1.2$ & $14/52$ & $5/24$ & $8/42$ & $337/403$ & $294/317$ & $346/394$ & 215 / 300 \\
\bottomrule
\end{tabular*}
\medskip
\begin{tabular*}{\textwidth}{@{\extracolsep{\fill}}l ccc ccc ccc r@{}}
\toprule
\multicolumn{11}{@{}l}{\textit{Eval-Opt evaluator}} \\
 & \multicolumn{3}{c}{Accepts truncated wrong} & \multicolumn{3}{c}{Rejects finished wrong} & \multicolumn{3}{c}{Accepts finished correct} & \multicolumn{1}{r@{}}{Steps} \\
\cmidrule(lr){2-4}\cmidrule(lr){5-7}\cmidrule(lr){8-10}
Scale & Base & \IPa & \SPa & Base & \IPa & \SPa & Base & \IPa & \SPa & \IPa{} / \SPa \\
\midrule
0.6B & $5/7$ & $0/1257$ & $24/1007$ & $114/1259$ & $3/16$ & $2/162$ & $136/146$ & $108/120$ & $234/235$ & 230 / 220 \\
1.7B & $49/49$ & $73/756$ & $28/714$ & $22/987$ & $2/99$ & $12/145$ & $376/376$ & $550/550$ & $515/544$ & 370 / 320 \\
4B & $0/52$ & $0/837$ & $3/373$ & $464/692$ & $1/11$ & $115/200$ & $531/668$ & $543/548$ & $362/836$ & 220 / 215 \\
\bottomrule
\end{tabular*}
\medskip
\begin{tabular*}{\textwidth}{@{\extracolsep{\fill}}l ccc ccc ccc ccc r@{}}
\toprule
\multicolumn{14}{@{}l}{\textit{Orch-Workers synthesizer}} \\
 & \multicolumn{3}{c}{Answer-only} & \multicolumn{3}{c}{Equals plurality} & \multicolumn{3}{c}{Loss} & \multicolumn{3}{c}{Rescue} & \multicolumn{1}{r@{}}{Steps} \\
\cmidrule(lr){2-4}\cmidrule(lr){5-7}\cmidrule(lr){8-10}\cmidrule(lr){11-13}
Scale & Base & \IPa & \SPa & Base & \IPa & \SPa & Base & \IPa & \SPa & Base & \IPa & \SPa & \IPa{} / \SPa \\
\midrule
0.6B & $0.03$ & $0.06$ & $0.41$ & $16/17$ & $71/73$ & $425/431$ & $0.10$ & $0.02$ & $0.03$ & $0.00$ & $0.10$ & $0.06$ & 200 / 180 \\
1.7B & $0.00$ & $0.39$ & $0.35$ & $2/2$ & $536/538$ & $484/485$ & $0.11$ & $0.01$ & $0.04$ & $0.01$ & $0.14$ & $0.07$ & 330 / 340 \\
4B & $0.00$ & $0.41$ & $0.55$ & -- & $519/523$ & $709/717$ & $0.09$ & $0.01$ & $0.02$ & $0.00$ & $0.16$ & $0.09$ & 180 / 220 \\
\bottomrule
\end{tabular*}
\end{table}

\subsection{Code Tasks}
\label{sec:appendix:code_task}

\paragraph{Columns.} Code responses are capped at 2{,}048 tokens. Table~\ref{tab:code_task} follows each \IPa run from its selected peak to its last checkpoint at step $\leq 300$ on the 1000 re-evaluated problems, with the exact McNemar $p$, and compares \IPa with \SPa at the selected peaks and at the common step cap, the earlier of the two runs' last validated steps. In Table~\ref{tab:code_reward_accounting}, each value is a mean over the checkpoints of a task and workflow. The difference from SA-RL is a checkpoint's solving-role accuracy minus that of the SA-RL checkpoint at the nearest step on the same problems, averaged over all checkpoints or over those of one scale. \emph{Rewarded solving-role outputs: Wrong} is the share of wrong outputs among the solving-role outputs of correct episodes. \emph{Disagreement over independence} divides the share of problems on which the instances disagree by the share expected if each instance solved each problem independently with that problem's accuracy at the cell's other checkpoints. Under Eval-Opt and SA-RL a dash marks a row that needs several graded solving-role outputs per episode, and in the SA-RL column also a comparison of SA-RL with itself.

\paragraph{Solving roles.} Table~\ref{tab:code_reward_accounting} gives the values for the Code analysis of \S\ref{sec:results:landscape}. Truncated Code outputs fall almost only on problems that no instance of the episode solves, so they are rarely rewarded. The share of wrong outputs among the rewarded ones is about the same on both tasks, so the lower rewarded-but-wrong share on Code follows from its lower accuracy.

\paragraph{Isolated against shared policies.} No pair is equivalent at $\pm 2$ points by TOST, at the common step cap or at the selected peaks. The Eval-Opt-4B-Code pair favors \SPa at the selected peaks but is within noise at the common step cap (Table~\ref{tab:code_task}), because the \IPa run ends early and the \SPa run peaks after the cap.

\paragraph{Unclosed code blocks.} In \texttt{Orch-Workers-IP-0.6B-Code} the workers leave their code blocks unclosed, which the grader scores as wrong, while the synthesizer closes its block, so the episode reward does not penalize the omission. Graded with the closing fence appended, the workers reach 7.8 to 8.6\% at steps 230, 270 and 275, against 11.8 to 11.9\% for SA-RL at the nearest step. The cell is excluded from Table~\ref{tab:code_reward_accounting} and from the relative statements about the Code solving roles.

\begin{table}[t]
\centering
\caption{Code \IPa runs after their selected peak at step $\le 300$ (steps in Table~\ref{tab:peak_last}), and \IPa minus \SPa at the common step cap and at the selected peaks. Validations after: validated checkpoints after the peak. Accuracy differences in points on the 1000 problems, paired exact McNemar $p$ in parentheses; \textemdash: the peak is the last validated checkpoint.}
\label{tab:code_task}
\resizebox{\textwidth}{!}{%
\begin{tabular}{@{}lccccc@{}}
\toprule
& \multicolumn{2}{c}{\IPa run, steps $\le 300$} & \multicolumn{2}{c}{Common step cap} & Selected peaks \\
\cmidrule(lr){2-3}\cmidrule(lr){4-5}\cmidrule(l){6-6}
Workflow and scale & Validations after & Last $-$ peak ($p$) & Steps IP/SP & IP $-$ SP ($p$) & IP $-$ SP ($p$) \\
\midrule
Voting-0.6B & $0$ & \textemdash & $300/270$ & $+1.4$ ($0.12$) & $+1.4$ ($0.12$) \\
Voting-1.7B & $1$ & $0.0$ ($1.00$) & $250/210$ & $+1.7$ ($0.09$) & $+2.2$ ($0.03$) \\
Voting-4B & $2$ & $+1.2$ ($0.30$) & $110/110$ & $+0.6$ ($0.66$) & $+2.5$ ($0.06$) \\
\addlinespace[2pt]
Orch-Workers-0.6B & $0$ & \textemdash & $230/250$ & $+0.7$ ($0.55$) & $+1.3$ ($0.22$) \\
Orch-Workers-1.7B & $2$ & $-1.7$ ($0.13$) & $180/170$ & $+2.0$ ($0.09$) & $+2.8$ ($0.02$) \\
Orch-Workers-4B & $3$ & $+0.2$ ($0.93$) & $140/110$ & $+2.5$ ($0.06$) & $+2.3$ ($0.08$) \\
\addlinespace[2pt]
Eval-Opt-0.6B & $7$ & $-0.7$ ($0.52$) & $170/240$ & $+3.0$ ($3{\times}10^{-3}$) & $+2.1$ ($0.04$) \\
Eval-Opt-1.7B & $0$ & \textemdash & $300/250$ & $+6.5$ ($3{\times}10^{-8}$) & $+6.5$ ($3{\times}10^{-8}$) \\
Eval-Opt-4B & $3$ & $-1.9$ ($0.14$) & $140/160$ & $-1.8$ ($0.18$) & $-7.8$ ($1{\times}10^{-8}$) \\
\bottomrule
\end{tabular}%
}
\end{table}

\begin{table}[t]
\centering
\footnotesize
\caption{Reward accounting of the solving roles (voters, workers, Eval-Opt first attempts) and SA-RL on all 1000 Code and 1412 Math validation problems, each output graded on its own answer: shares of outputs, of rewarded and truncated outputs and of problems, and differences from SA-RL in points (within and below count checkpoints), averaged over checkpoints. Orch.: Orch-Workers; SE: paired per-problem standard error. The \texttt{Orch-Workers-IP-0.6B-Code} cell is excluded.}
\label{tab:code_reward_accounting}
\setlength{\tabcolsep}{3.5pt}
\resizebox{\ifdim\width>\textwidth\textwidth\else\width\fi}{!}{%
\begin{tabular}{@{}lcccccccc@{}}
\toprule
& \multicolumn{4}{c}{Code (1000 problems)} & \multicolumn{4}{c}{Math (1412 problems)} \\
\cmidrule(lr){2-5}\cmidrule(l){6-9}
& Voting & Orch. & Eval-Opt & SA-RL & Voting & Orch. & Eval-Opt & SA-RL \\
\midrule
Checkpoints & $20$ & $15$ & $19$ & $14$ & $27$ & $25$ & $23$ & $10$ \\
\addlinespace[2pt]
\multicolumn{9}{@{}l}{\textit{All solving-role outputs}} \\
\quad At the length cap & $39.7$ & $28.1$ & $59.2$ & $37.3$ & $21.8$ & $34.1$ & $49.2$ & $30.0$ \\
\quad Truncated and rewarded & $1.0$ & $1.4$ & $4.6$ & $0.0$ & $3.4$ & $9.9$ & $14.0$ & $0.1$ \\
\quad Rewarded but wrong & $3.9$ & $8.9$ & $6.7$ & $0.0$ & $9.9$ & $15.8$ & $14.1$ & $0.0$ \\
\addlinespace[2pt]
\multicolumn{9}{@{}l}{\textit{Rewarded solving-role outputs}} \\
\quad Wrong & $21.5$ & $36.2$ & $30.8$ & $0.0$ & $24.8$ & $35.6$ & $29.5$ & $0.0$ \\
\addlinespace[2pt]
\multicolumn{9}{@{}l}{\textit{Truncated solving-role outputs}} \\
\quad Rewarded & $2.9$ & $5.5$ & $7.4$ & $0.0$ & $17.1$ & $26.4$ & $26.4$ & $0.3$ \\
\quad No instance solves & $96.1$ & $95.1$ & \textemdash & \textemdash & $78.8$ & $80.0$ & \textemdash & \textemdash \\
\quad SA-RL solves & $3.1$ & $4.3$ & $5.9$ & \textemdash & $22.2$ & $22.9$ & $22.4$ & \textemdash \\
\addlinespace[2pt]
\multicolumn{9}{@{}l}{\textit{Problems}} \\
\quad Instances disagree & $12.4$ & $19.1$ & \textemdash & \textemdash & $30.2$ & $29.4$ & \textemdash & \textemdash \\
\quad Disagreement over independence & $1.15$ & $1.26$ & \textemdash & \textemdash & $1.04$ & $1.04$ & \textemdash & \textemdash \\
\addlinespace[2pt]
\multicolumn{9}{@{}l}{\textit{Solving role minus SA-RL}} \\
\quad Mean (points) & $-1.6$ & $-3.5$ & $-3.4$ & \textemdash & $-9.7$ & $-11.8$ & $-9.2$ & \textemdash \\
\quad Mean at 0.6B & $-1.0$ & $-2.8$ & $-3.7$ & \textemdash & $-7.4$ & $-8.2$ & $-9.4$ & \textemdash \\
\quad Mean at 1.7B & $-1.8$ & $-2.1$ & $-2.0$ & \textemdash & $-9.0$ & $-13.6$ & $-9.0$ & \textemdash \\
\quad Mean at 4B & $-2.1$ & $-5.4$ & $-3.9$ & \textemdash & $-11.9$ & $-13.3$ & $-9.3$ & \textemdash \\
\quad Within 3 points & $18/20$ & $7/15$ & $10/19$ & \textemdash & $2/27$ & $0/25$ & $4/23$ & \textemdash \\
\quad More than 2 SE below & $9/20$ & $12/15$ & $14/19$ & \textemdash & $26/27$ & $25/25$ & $21/23$ & \textemdash \\
\bottomrule
\end{tabular}%
}
\end{table}

\subsection{Per-Turn Decomposition of Eval-Opt}
\label{sec:appendix:eval_opt_turns}

\begin{figure}[t]
\centering
\includegraphics[width=0.9\textwidth]{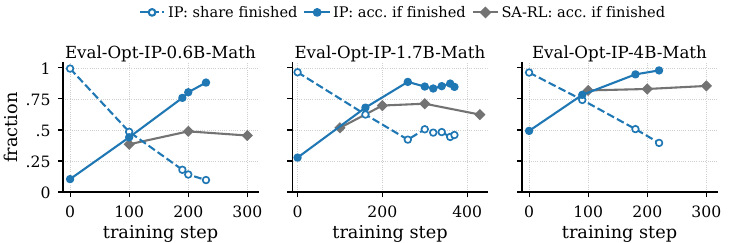}
\caption{\IPa first attempts of Eval-Opt on Math over training, all 1412 validation problems; step~0 is the untrained workflow. Dashed: share of first attempts that finish under the length cap; solid: their accuracy; gray: accuracy of SA-RL's finished outputs at its sampled checkpoints. Values: Table~\ref{tab:eval_opt_turns}.}
\label{fig:eval_opt_first_attempt}
\end{figure}

Table~\ref{tab:eval_opt_turns} gives the per-checkpoint values behind Figs.~\ref{fig:eval_opt_loop} and~\ref{fig:eval_opt_first_attempt}. The Val.\ column is the validation accuracy, and the per-turn quantities come from the trajectory dumps of a separate evaluation of the same checkpoint. The eval.\ column counts the recovered episodes in which the evaluator stated the correct answer before the generator did. The evaluator columns describe its first verdict, and an unparsable verdict counts as a rejection, as in the workflow. Tokens are the generator's, summed over all turns of an episode.

As training proceeds the \IPa generator finishes its first attempt less often, and the first attempts it finishes become more accurate than SA-RL's finished outputs (Fig.~\ref{fig:eval_opt_first_attempt}).

\begin{table}[p]
\centering
\caption{Per-turn decomposition of Eval-Opt episodes on Math, each generator turn graded on its own answer. Val.: validation accuracy of the checkpoint; all other columns come from the trajectory dumps of a separate evaluation of the same checkpoint (1{,}412 problems). SA-RL: validation accuracy\,/\,accuracy of finished outputs (step). Definitions in \S\ref{sec:appendix:measurements}.}
\label{tab:eval_opt_turns}
\setlength{\tabcolsep}{3.5pt}
\resizebox{\textwidth}{!}{%
\begin{tabular}{@{}llrcccccccccccccccrc@{}}
\toprule
 & & & \multicolumn{2}{c}{Final} & \multicolumn{3}{c}{First attempt} & Loop & \multicolumn{3}{c}{Recovered} & & \multicolumn{4}{c}{Evaluator, first verdict} & \multicolumn{2}{c}{Per episode} & SA-RL \\
\cmidrule(lr){4-5}\cmidrule(lr){6-8}\cmidrule(lr){10-12}\cmidrule(lr){14-17}\cmidrule(lr){18-19}
Scale & Strategy & Step & val. & dump & acc. & trunc. & fin.\ acc. & net (pp) & T-wrong & F-wrong & eval. & Lost & Accept & TPR & Rej. F-wrong & Rej. T-wrong & Rounds & Tokens & acc.\,/\,fin.\ (step) \\
\midrule
\multicolumn{20}{@{}l}{\textit{Math: 5{,}120 tokens per turn, up to three generator turns}} \\
0.6B & Base & 0 & $0.11$ & $0.10$ & $0.10$ & $0.00$ & $0.10$ & $+0.1$ & $0/7$ & $2/1259$ & $0$ & $0/146$ & $0.91$ & $0.93$ & $0.09$ & $0.29$ & $0.11$ & $976$ & -- \\
 & \IPa & 100 & $0.28$ & $0.29$ & $0.21$ & $0.51$ & $0.44$ & $+7.8$ & $101/726$ & $15/383$ & $29$ & $6/303$ & $0.36$ & $0.84$ & $0.37$ & $0.99$ & $1.12$ & $7{,}457$ & $0.23$\,/\,$0.38$\,(100) \\
 & \IPa & 190 & $0.37$ & $0.36$ & $0.14$ & $0.82$ & $0.76$ & $+22.3$ & $309/1156$ & $6/61$ & $235$ & $0/195$ & $0.14$ & $0.81$ & $0.38$ & $1.00$ & $1.50$ & $9{,}889$ & $0.26$\,/\,$0.49$\,(190\,/\,200) \\
 & \IPa & 200 & $0.37$ & $0.37$ & $0.12$ & $0.86$ & $0.80$ & $+25.0$ & $353/1204$ & $0/39$ & $262$ & $0/169$ & $0.14$ & $0.94$ & $0.05$ & $1.00$ & $1.31$ & $9{,}569$ & $0.26$\,/\,$0.49$\,(200) \\
 & \IPa & 230 & $0.40$ & $0.39$ & $0.10$ & $0.90$ & $0.88$ & $+28.9$ & $405/1257$ & $3/16$ & $344$ & $0/139$ & $0.09$ & $0.78$ & $0.19$ & $1.00$ & $1.53$ & $10{,}195$ & $0.27$\,/\,$0.49$\,(230\,/\,200) \\
 & \SPa & 110 & $0.19$ & $0.18$ & $0.17$ & $0.14$ & $0.20$ & $+1.0$ & $11/197$ & $9/972$ & $0$ & $6/243$ & $0.69$ & $0.90$ & $0.23$ & $0.96$ & $0.60$ & $3{,}081$ & $0.22$\,/\,$0.38$\,(110\,/\,100) \\
 & \SPa & 210 & $0.27$ & $0.29$ & $0.18$ & $0.65$ & $0.52$ & $+10.3$ & $146/917$ & $1/238$ & $67$ & $1/257$ & $0.36$ & $0.99$ & $0.02$ & $0.97$ & $0.99$ & $7{,}607$ & $0.27$\,/\,$0.49$\,(210\,/\,200) \\
 & \SPa & 220 & $0.29$ & $0.29$ & $0.17$ & $0.72$ & $0.59$ & $+12.1$ & $172/1008$ & $0/162$ & $109$ & $1/242$ & $0.30$ & $0.97$ & $0.01$ & $0.98$ & $1.13$ & $8{,}396$ & $0.28$\,/\,$0.49$\,(220\,/\,200) \\
\addlinespace[2pt]
1.7B & Base & 0 & $0.29$ & $0.27$ & $0.27$ & $0.03$ & $0.28$ & $+0.1$ & $0/49$ & $2/985$ & $0$ & $0/378$ & $0.98$ & $1.00$ & $0.02$ & $0.00$ & $0.03$ & $1{,}574$ & -- \\
 & \IPa & 160 & $0.50$ & $0.50$ & $0.43$ & $0.38$ & $0.68$ & $+7.0$ & $99/528$ & $6/282$ & $18$ & $6/602$ & $0.54$ & $0.92$ & $0.29$ & $0.97$ & $0.85$ & $5{,}357$ & $0.42$\,/\,$0.70$\,(160\,/\,200) \\
 & \IPa & 260 & $0.58$ & $0.59$ & $0.38$ & $0.58$ & $0.89$ & $+20.8$ & $293/808$ & $0/67$ & $88$ & $0/537$ & $0.41$ & $0.95$ & $0.15$ & $0.99$ & $0.95$ & $6{,}896$ & $0.48$\,/\,$0.71$\,(260\,/\,300) \\
 & \IPa & 300 & $0.57$ & $0.58$ & $0.43$ & $0.49$ & $0.85$ & $+14.4$ & $204/696$ & $0/107$ & $43$ & $0/609$ & $0.52$ & $0.99$ & $0.08$ & $0.96$ & $0.80$ & $6{,}345$ & $0.50$\,/\,$0.71$\,(300) \\
 & \IPa & 320 & $0.57$ & $0.55$ & $0.41$ & $0.52$ & $0.83$ & $+14.8$ & $209/728$ & $0/112$ & $50$ & $0/572$ & $0.50$ & $0.99$ & $0.05$ & $0.95$ & $0.82$ & $6{,}465$ & $0.50$\,/\,$0.71$\,(320\,/\,300) \\
 & \IPa & 340 & $0.56$ & $0.57$ & $0.42$ & $0.52$ & $0.85$ & $+14.9$ & $210/724$ & $0/100$ & $52$ & $0/588$ & $0.51$ & $0.99$ & $0.04$ & $0.94$ & $0.78$ & $6{,}564$ & $0.54$\,/\,$0.71$\,(340\,/\,300) \\
 & \IPa & 360 & $0.53$ & $0.53$ & $0.39$ & $0.56$ & $0.87$ & $+13.5$ & $190/779$ & $0/79$ & $64$ & $0/554$ & $0.50$ & $1.00$ & $0.01$ & $0.90$ & $0.72$ & $6{,}812$ & $0.51$\,/\,$0.71$\,(360\,/\,300) \\
 & \IPa & 370 & $0.54$ & $0.55$ & $0.39$ & $0.54$ & $0.85$ & $+15.2$ & $214/756$ & $0/99$ & $73$ & $0/557$ & $0.51$ & $1.00$ & $0.02$ & $0.90$ & $0.70$ & $6{,}727$ & $0.53$\,/\,$0.62$\,(370\,/\,430) \\
 & \SPa & 150 & $0.38$ & $0.38$ & $0.36$ & $0.20$ & $0.45$ & $+2.3$ & $27/284$ & $21/620$ & $9$ & $16/508$ & $0.57$ & $0.87$ & $0.44$ & $0.96$ & $0.79$ & $4{,}081$ & $0.41$\,/\,$0.52$\,(150\,/\,100) \\
 & \SPa & 290 & $0.51$ & $0.48$ & $0.40$ & $0.42$ & $0.68$ & $+8.6$ & $119/592$ & $4/261$ & $73$ & $1/559$ & $0.60$ & $0.98$ & $0.06$ & $0.91$ & $0.67$ & $5{,}476$ & $0.50$\,/\,$0.71$\,(290\,/\,300) \\
 & \SPa & 320 & $0.51$ & $0.54$ & $0.39$ & $0.51$ & $0.79$ & $+15.3$ & $217/714$ & $2/145$ & $138$ & $3/553$ & $0.48$ & $0.93$ & $0.08$ & $0.96$ & $0.87$ & $6{,}638$ & $0.50$\,/\,$0.71$\,(320\,/\,300) \\
\addlinespace[2pt]
4B & Base & 0 & $0.48$ & $0.48$ & $0.47$ & $0.04$ & $0.49$ & $+0.6$ & $5/52$ & $41/690$ & $2$ & $37/670$ & $0.54$ & $0.80$ & $0.67$ & $1.00$ & $0.84$ & $2{,}655$ & -- \\
 & \IPa & 90 & $0.67$ & $0.64$ & $0.58$ & $0.26$ & $0.78$ & $+5.7$ & $81/363$ & $4/227$ & $4$ & $4/822$ & $0.72$ & $0.96$ & $0.15$ & $0.93$ & $0.44$ & $3{,}987$ & $0.61$\,/\,$0.82$\,(90\,/\,100) \\
 & \IPa & 180 & $0.74$ & $0.74$ & $0.49$ & $0.49$ & $0.95$ & $+25.7$ & $363/687$ & $1/37$ & $20$ & $1/688$ & $0.51$ & $0.99$ & $0.05$ & $1.00$ & $0.69$ & $6{,}266$ & $0.64$\,/\,$0.83$\,(180\,/\,200) \\
 & \IPa & 220 & $0.73$ & $0.75$ & $0.40$ & $0.60$ & $0.98$ & $+35.5$ & $503/838$ & $0/11$ & $34$ & $2/563$ & $0.39$ & $0.96$ & $0.09$ & $1.00$ & $0.89$ & $7{,}401$ & $0.64$\,/\,$0.83$\,(220\,/\,200) \\
 & \SPa & 105 & -- & $0.56$ & $0.54$ & $0.12$ & $0.61$ & $+2.0$ & $20/164$ & $39/481$ & $0$ & $31/767$ & $0.48$ & $0.68$ & $0.69$ & $0.96$ & $0.91$ & $3{,}379$ & $0.60$\,/\,$0.82$\,(100) \\
 & \SPa & 210 & $0.66$ & $0.68$ & $0.61$ & $0.26$ & $0.82$ & $+7.5$ & $105/363$ & $10/192$ & $9$ & $9/857$ & $0.49$ & $0.66$ & $0.38$ & $0.99$ & $0.77$ & $4{,}207$ & $0.63$\,/\,$0.83$\,(210\,/\,200) \\
 & \SPa & 215 & -- & $0.66$ & $0.59$ & $0.27$ & $0.81$ & $+6.9$ & $94/373$ & $17/200$ & $12$ & $14/839$ & $0.32$ & $0.43$ & $0.57$ & $0.99$ & $1.02$ & $4{,}453$ & $0.63$\,/\,$0.83$\,(210\,/\,200) \\
\bottomrule
\end{tabular}%
}
\end{table}

\paragraph{Fixed controllers.}
Table~\ref{tab:eval_opt_controllers} holds the controller fixed. Each variant is a fresh evaluation on the 1412 validation problems, one sample per problem, with the 5120-token cap per turn and up to three generator turns, at the selected peak of the Eval-Opt cell and of SA-RL. Rule~(a), \emph{retry if cut off, evaluator text kept}, calls the evaluator and passes its text to the generator but continues only after a truncated attempt. Rule~(b), \emph{retry if cut off, no evaluator call}, continues a truncated attempt with a fixed continuation message. \emph{Always accept} stops after the first attempt, and \emph{always reject} runs all three attempts. SA-RL under rule~(b) and one SA-RL response at a 10240- or 15360-token cap are the comparators R, L$_{10}$ and L$_{15}$ of Appendix~\ref{sec:appendix:sa_baseline}; the longer-response comparisons carry no McNemar test.

The margin of trained Eval-Opt under \IPa over one SA-RL response splits into three terms. A, the retry structure, is SA-RL with rule~(b) minus one SA-RL response. B, the generator, is the Eval-Opt \IPa generator minus the SA-RL generator, both under rule~(b); it compares the generators at a matched controller, not a matched budget. C, the evaluator, is the trained evaluator minus rule~(b) with the Eval-Opt \IPa generator. A paired value marked $^{\dagger}$ is computed on a separate evaluation of the checkpoint and can therefore differ from the level difference next to it.

\begin{table}[t]
\centering
\caption{Eval-Opt on Math with the controller fixed, 1{,}412 validation problems; group rows give the checkpoints. Top: accuracy (\%), first-attempt truncation (\%), generator attempts, model calls and completion tokens (thousands) per episode, per sample for one SA-RL response. Accuracy of trained workflows and of one SA-RL response at cap 5{,}120 is the validation accuracy. Bottom: differences in points; level difference, paired 95\% CI, McNemar $p$. $^{\dagger}$: paired on a separate evaluation of the checkpoint; $^{\ddagger}$: consumption at step 220.}
\label{tab:eval_opt_controllers}
\setlength{\tabcolsep}{3pt}
\resizebox{\textwidth}{!}{%
\begin{tabular}{@{}lrrrrrrrrrrrrrrr@{}}
\toprule
 & \multicolumn{5}{c}{0.6B} & \multicolumn{5}{c}{1.7B} & \multicolumn{5}{c}{4B} \\
\cmidrule(lr){2-6}\cmidrule(lr){7-11}\cmidrule(l){12-16}
System & Acc. & Trunc. & Att. & Calls & Tok. & Acc. & Trunc. & Att. & Calls & Tok. & Acc. & Trunc. & Att. & Calls & Tok. \\
\midrule
\multicolumn{16}{@{}l}{\textit{Eval-Opt-\IPa generator, steps 230\,/\,260\,/\,230}} \\
Trained evaluator & 39.5 & 90 & 2.53 & 5.07 & 14.1 & 57.6 & 58 & 1.95 & 3.89 & 7.5 & 75.2 & 60$^{\ddagger}$ & 1.89$^{\ddagger}$ & 3.78$^{\ddagger}$ & 7.6$^{\ddagger}$ \\
Retry if cut off, evaluator text kept & 37.4 & 89 & 2.25 & 4.51 & 13.3 & 57.9 & 58 & 1.74 & 3.48 & 7.1 & 74.7 & 59 & 1.81 & 3.63 & 7.2 \\
Retry if cut off, no evaluator call & 34.3 & 91 & 2.51 & 2.51 & 11.2 & 60.4 & 57 & 1.75 & 1.75 & 6.7 & 74.9 & 60 & 1.82 & 1.82 & 7.1 \\
Always accept & 9.6 & 90 & 1.00 & 1.00 & 4.9 & 38.7 & 56 & 1.00 & 1.00 & 4.0 & 40.0 & 60 & 1.00 & 1.00 & 4.1 \\
Always reject & 38.3 & 90 & 3.00 & 6.00 & 14.9 & 58.3 & 57 & 3.00 & 6.00 & 8.8 & 75.6 & 59 & 3.00 & 6.00 & 9.1 \\
\addlinespace[2pt]
\multicolumn{16}{@{}l}{\textit{Eval-Opt-\SPa generator, steps 220\,/\,290\,/\,200}} \\
Trained evaluator & 29.5 & 72 & 2.13 & 4.27 & 11.5 & 50.5 & 42 & 1.67 & 3.34 & 7.0 & 65.8 & 23 & 1.64 & 3.28 & 4.0 \\
Retry if cut off, evaluator text kept & 29.9 & 72 & 2.08 & 4.15 & 11.3 & 50.1 & 42 & 1.50 & 3.00 & 6.5 & 66.6 & 24 & 1.26 & 2.53 & 3.6 \\
Retry if cut off, no evaluator call & 26.6 & 73 & 2.22 & 2.22 & 9.5 & 50.6 & 41 & 1.59 & 1.59 & 5.8 & 65.5 & 25 & 1.28 & 1.28 & 3.5 \\
\addlinespace[2pt]
\multicolumn{16}{@{}l}{\textit{SA-RL generator, steps 290\,/\,290\,/\,300}} \\
One response, cap 5{,}120 & 31.4 & 24 & 1.00 & 1.00 & 3.5 & 49.9 & 29 & 1.00 & 1.00 & 3.3 & 66.6 & 23 & 1.00 & 1.00 & 2.9 \\
Retry if cut off, no evaluator call & 33.3 & 25 & 1.25 & 1.25 & 4.0 & 55.9 & 29 & 1.31 & 1.31 & 4.1 & 73.3 & 23 & 1.24 & 1.24 & 3.4 \\
One response, cap 10{,}240 & 33.8 & 1 & 1.00 & 1.00 & 3.9 & 57.4 & 4 & 1.00 & 1.00 & 4.0 & 74.6 & 4 & 1.00 & 1.00 & 3.5 \\
One response, cap 15{,}360 & 34.6 & 0 & 1.00 & 1.00 & 3.9 & 56.4 & 1 & 1.00 & 1.00 & 4.1 & 76.3 & 1 & 1.00 & 1.00 & 3.6 \\
\bottomrule
\end{tabular}%
}
\par\vspace{6pt}
\resizebox{\textwidth}{!}{%
\begin{tabular}{@{}lrcrrcrrcr@{}}
\toprule
 & \multicolumn{3}{c}{0.6B} & \multicolumn{3}{c}{1.7B} & \multicolumn{3}{c}{4B} \\
\cmidrule(lr){2-4}\cmidrule(lr){5-7}\cmidrule(l){8-10}
Difference (points) & Level & Paired 95\% CI & $p$ & Level & Paired 95\% CI & $p$ & Level & Paired 95\% CI & $p$ \\
\midrule
\multicolumn{10}{@{}l}{\textit{Fixed controller minus trained evaluator}} \\
\IPa: retry if cut off, evaluator text kept & $-2.1$ & [$-4.5$, $0.3$] & 0.08 & $+0.3$ & $-0.9^{\dagger}$\,[$-3.2$, $1.3$] & 0.46 & $-0.5$ & [$-2.3$, $1.3$] & 0.64 \\
\IPa: retry if cut off, no evaluator call & $-5.2$ & [$-7.5$, $-2.7$] & $<$0.001 & $+2.8$ & $+1.6^{\dagger}$\,[$-0.6$, $4.0$] & 0.18 & $-0.3$ & [$-2.1$, $1.5$] & 0.82 \\
\IPa: always accept & $-29.9$ & [$-32.3$, $-27.5$] & $<$0.001 & $-18.8$ & $-20.0^{\dagger}$\,[$-22.5$, $-17.5$] & $<$0.001 & $-35.2$ & [$-37.8$, $-32.9$] & $<$0.001 \\
\IPa: always reject & $-1.2$ & [$-3.5$, $1.2$] & 0.36 & $+0.7$ & $-0.5^{\dagger}$\,[$-2.8$, $1.8$] & 0.72 & $+0.4$ & [$-1.3$, $2.2$] & 0.69 \\
\SPa: retry if cut off, evaluator text kept & $+0.4$ & [$-2.0$, $2.9$] & 0.78 & $-0.4$ & $+1.8^{\dagger}$\,[$-0.6$, $4.3$] & 0.16 & $+0.9$ & [$-1.6$, $3.2$] & 0.51 \\
\SPa: retry if cut off, no evaluator call & $-2.8$ & [$-5.2$, $-0.4$] & 0.03 & $+0.1$ & $+2.4^{\dagger}$\,[$-0.1$, $5.0$] & 0.08 & $-0.3$ & [$-2.5$, $2.1$] & 0.86 \\
\addlinespace[2pt]
\multicolumn{10}{@{}l}{\textit{Margin of trained Eval-Opt-\IPa over one SA-RL response}} \\
Total ($=$ A $+$ B $+$ C) & $+8.2$ & $+8.7^{\dagger}$\,[$6.8$, $10.7$] & -- & $+7.6$ & $+9.2^{\dagger}$\,[$7.1$, $11.3$] & -- & $+8.6$ & [$6.4$, $10.8$] & $<$0.001 \\
A: retry structure, SA-RL generator & $+1.9$ & $+2.5^{\dagger}$\,[$0.6$, $4.4$] & -- & $+6.0$ & $+6.3^{\dagger}$\,[$4.4$, $8.3$] & -- & $+6.7$ & [$4.8$, $8.8$] & $<$0.001 \\
B: Eval-Opt-\IPa vs SA-RL generator, same retry & $+1.1$ & [$-1.2$, $3.3$] & 0.40 & $+4.5$ & [$2.3$, $6.8$] & $<$0.001 & $+1.6$ & [$-0.3$, $3.5$] & 0.11 \\
C: trained evaluator vs the same retry & $+5.2$ & [$2.8$, $7.5$] & $<$0.001 & $-2.8$ & $-1.6^{\dagger}$\,[$-4.0$, $0.6$] & 0.18 & $+0.3$ & [$-1.6$, $2.1$] & 0.82 \\
\addlinespace[2pt]
\multicolumn{10}{@{}l}{\textit{Other differences}} \\
Eval-Opt-\SPa vs SA-RL generator, same retry & $-6.7$ & [$-9.1$, $-4.2$] & $<$0.001 & $-5.2$ & [$-7.7$, $-2.7$] & $<$0.001 & $-7.8$ & [$-10.0$, $-5.5$] & $<$0.001 \\
Trained Eval-Opt-\IPa minus SA-RL with the same retry & $+6.2$ & [$3.8$, $8.7$] & $<$0.001 & $+1.7$ & $+2.9^{\dagger}$\,[$0.5$, $5.2$] & 0.02 & $+1.9$ & [$0.0$, $3.8$] & 0.07 \\
Trained Eval-Opt-\IPa minus SA-RL, one response, cap 10{,}240 & $+5.7$ & [$3.7$, $7.7$] & -- & $+0.1$ & $+1.3^{\dagger}$\,[$-0.6$, $3.4$] & -- & $+0.6$ & [$-1.1$, $2.3$] & -- \\
Trained Eval-Opt-\IPa minus SA-RL, one response, cap 15{,}360 & $+4.9$ & [$2.9$, $6.9$] & -- & $+1.2$ & $+2.4^{\dagger}$\,[$0.3$, $4.4$] & -- & $-1.1$ & [$-2.8$, $0.6$] & -- \\
\IPa minus \SPa, trained & $+10.1$ & [$7.5$, $12.6$] & $<$0.001 & $+7.1$ & $+10.6^{\dagger}$\,[$8.0$, $13.1$] & $<$0.001 & $+9.4$ & [$7.2$, $11.7$] & $<$0.001 \\
\IPa minus \SPa, retry if cut off, no evaluator call & $+7.7$ & [$5.3$, $10.1$] & $<$0.001 & $+9.8$ & [$7.2$, $12.3$] & $<$0.001 & $+9.4$ & [$7.2$, $11.6$] & $<$0.001 \\
\bottomrule
\end{tabular}%
}
\end{table}

\section{Additional Analysis of Policy Sharing and Post-Peak Decline}
\label{sec:appendix:mechanism_evidence}

\subsection{Per-Role Training Diagnostics}
\label{sec:appendix:gradient_amplification_signatures}

These measurements place the instability of \S\ref{sec:results:ip_decline} in the parallel role. Table~\ref{tab:checkpoint_divergence} gives per-token entropy and checkpoint KL, the change between checkpoints, both measured on fixed responses (\S\ref{sec:appendix:measurements}). The brief checkpoint-KL jumps of the decision roles also occur in non-declining shared-policy runs.

\begin{table}[t]
\centering
\caption{Per-role per-token entropy $H$ and checkpoint KL at probes before the selected peak, at it and at the last probe; the selected peak step is in parentheses. Values are ratios to the first probe; brackets: 95\% bootstrap interval. $\times 3$: parallel role; $^{\uparrow}$: rises at every probe; Max: largest KL ratio (step). Definitions: \S\ref{sec:appendix:measurements}.}
\label{tab:checkpoint_divergence}
\resizebox{\textwidth}{!}{%
\begin{tabular}{@{}llrllllrlll@{}}
\toprule
& & \multicolumn{3}{c}{Before peak} & \multicolumn{2}{c}{At peak} & \multicolumn{3}{c}{Last probe} & \\
\cmidrule(lr){3-5}\cmidrule(lr){6-7}\cmidrule(lr){8-10}
Run (peak) & Role & Step & $H$ & KL & $H$ & KL & Step & $H$ & KL & Max KL \\
\midrule
\multicolumn{11}{@{}l}{\emph{Isolated policy, declining}} \\
\multirow{2}{*}{\shortstack[l]{Voting-IP-1.7B\\(190)}} & generator ($\times 3$) & 180 & $2.67$ & $3.84$ {\scriptsize[3.65, 4.05]} & $3.11$ & $5.42$ {\scriptsize[5.15, 5.72]} & 290 & $12.3^{\uparrow}$ & $10.8^{\uparrow}$ {\scriptsize[10.3, 11.3]} & $10.8$ (290) \\
 & aggregator & 180 & $0.91$ & $0.18$ {\scriptsize[0.15, 0.21]} & $0.81$ & $0.25$ {\scriptsize[0.18, 0.32]} & 290 & $0.80$ & $0.17$ {\scriptsize[0.15, 0.20]} & $0.49$ (100) \\
\addlinespace[3pt]
\multirow{2}{*}{\shortstack[l]{Voting-IP-4B\\(100)}} & generator ($\times 3$) & 90 & $1.71$ & $4.05$ {\scriptsize[3.88, 4.23]} & $2.01$ & $3.83$ {\scriptsize[3.67, 4.01]} & 205 & $8.85^{\uparrow}$ & $14.1$ {\scriptsize[13.6, 14.7]} & $14.1$ (205) \\
 & aggregator & 90 & $1.33$ & $0.97$ {\scriptsize[0.48, 1.76]} & $1.28$ & $0.72$ {\scriptsize[0.42, 1.22]} & 205 & $0.46$ & $0.15$ {\scriptsize[0.03, 0.33]} & $10.8$ (130) \\
\addlinespace[3pt]
\multirow{3}{*}{\shortstack[l]{Orch-Workers-IP-1.7B\\(290)}} & worker ($\times 3$) & 280 & $3.57$ & $2.89$ {\scriptsize[2.75, 3.03]} & $3.98$ & $3.38$ {\scriptsize[3.23, 3.55]} & 335 & $7.77^{\uparrow}$ & $8.41$ {\scriptsize[8.06, 8.79]} & $8.41$ (335) \\
 & orchestrator & 280 & $1.15$ & $8.17$ {\scriptsize[7.37, 9.01]} & $1.14$ & $12.5$ {\scriptsize[11.3, 13.8]} & 335 & $1.18$ & $2.03$ {\scriptsize[1.90, 2.17]} & $12.5$ (290) \\
 & synthesizer & 280 & $0.79$ & $1.28$ {\scriptsize[1.02, 1.65]} & $0.76$ & $0.91$ {\scriptsize[0.75, 1.12]} & 335 & $0.71$ & $2.81$ {\scriptsize[2.33, 3.44]} & $5.79$ (100) \\
\addlinespace[3pt]
\multirow{3}{*}{\shortstack[l]{Orch-Workers-IP-4B\\(100)}} & worker ($\times 3$) & 90 & $1.90$ & $2.81$ {\scriptsize[2.69, 2.92]} & $2.15$ & $4.13$ {\scriptsize[3.96, 4.31]} & 175 & $7.51^{\uparrow}$ & $19.3^{\uparrow}$ {\scriptsize[18.6, 20.1]} & $19.3$ (175) \\
 & orchestrator & 90 & $1.05$ & $0.59$ {\scriptsize[0.55, 0.64]} & $1.04$ & $0.62$ {\scriptsize[0.57, 0.67]} & 175 & $1.05$ & $1.17$ {\scriptsize[1.07, 1.29]} & $1.44$ (50) \\
 & synthesizer & 90 & $1.43$ & $0.42$ {\scriptsize[0.37, 0.47]} & $1.33$ & $0.74$ {\scriptsize[0.63, 0.88]} & 175 & $0.58$ & $0.42$ {\scriptsize[0.34, 0.53]} & $5.10$ (150) \\
\midrule
\multicolumn{11}{@{}l}{\emph{Isolated policy, not declining}} \\
\multirow{3}{*}{\shortstack[l]{Orch-Workers-IP-0.6B\\(200)}} & worker ($\times 3$) & 190 & $2.24$ & $1.49$ {\scriptsize[1.44, 1.56]} & $2.45$ & $1.40$ {\scriptsize[1.35, 1.46]} & 220 & $2.86^{\uparrow}$ & $1.43$ {\scriptsize[1.37, 1.49]} & $1.49$ (190) \\
 & orchestrator & 190 & $0.07$ & $7.52$ {\scriptsize[6.29, 8.91]} & $0.05$ & $17.3$ {\scriptsize[14.4, 20.2]} & 220 & $0.02$ & $1.37$ {\scriptsize[0.98, 1.82]} & $17.3$ (200) \\
 & synthesizer & 190 & $1.09$ & $3.05$ {\scriptsize[2.58, 3.69]} & $1.05$ & $1.45$ {\scriptsize[1.22, 1.75]} & 220 & $1.04$ & $1.72$ {\scriptsize[1.47, 2.04]} & $3.05$ (190) \\
\midrule
\multicolumn{11}{@{}l}{\emph{Shared policy and single-agent RL}} \\
\multirow{3}{*}{\shortstack[l]{Orch-Workers-SP-1.7B\\(300)}} & worker ($\times 3$) & 290 & $1.52$ & $1.94$ {\scriptsize[1.85, 2.04]} & $1.68$ & $1.56$ {\scriptsize[1.43, 1.77]} & 335 & $1.95$ & $1.73$ {\scriptsize[1.65, 1.82]} & $1.94$ (290) \\
 & orchestrator & 290 & $0.67$ & $2.80$ {\scriptsize[2.56, 3.08]} & $0.71$ & $2.06$ {\scriptsize[1.89, 2.24]} & 335 & $0.68$ & $4.68$ {\scriptsize[4.21, 5.19]} & $4.68$ (335) \\
 & synthesizer & 290 & $1.23$ & $3.15$ {\scriptsize[2.68, 3.76]} & $1.44$ & $2.69$ {\scriptsize[2.29, 3.22]} & 335 & $1.66$ & $2.22$ {\scriptsize[1.99, 2.48]} & $3.15$ (290) \\
\addlinespace[3pt]
\multirow{2}{*}{\shortstack[l]{Voting-SP-1.7B\\(300)}} & generator ($\times 3$) & 290 & $1.40^{\uparrow}$ & $1.14$ {\scriptsize[1.08, 1.20]} & \multicolumn{1}{c}{\textemdash} & \multicolumn{1}{c}{\textemdash} & \multicolumn{1}{c}{\textemdash} & \multicolumn{1}{c}{\textemdash} & \multicolumn{1}{c}{\textemdash} & $1.41$ (150) \\
 & aggregator & 290 & $0.53$ & $4.67$ {\scriptsize[3.57, 6.10]} & \multicolumn{1}{c}{\textemdash} & \multicolumn{1}{c}{\textemdash} & \multicolumn{1}{c}{\textemdash} & \multicolumn{1}{c}{\textemdash} & \multicolumn{1}{c}{\textemdash} & $4.67$ (290) \\
\addlinespace[3pt]
SA-RL-1.7B (290) & generator & 280 & $1.58$ & $1.76$ {\scriptsize[1.64, 1.90]} & $1.58$ & $2.60$ {\scriptsize[2.42, 2.80]} & 420 & $1.32$ & $2.18$ {\scriptsize[2.02, 2.35]} & $2.60$ (290) \\
\midrule
\multicolumn{11}{@{}l}{\emph{Code}} \\
\multirow{2}{*}{\shortstack[l]{Voting-IP-1.7B-Code\\(290)}} & generator ($\times 3$) & 280 & $0.91$ & $0.97$ {\scriptsize[0.90, 1.05]} & $0.95$ & $1.63$ {\scriptsize[1.51, 1.78]} & \multicolumn{1}{c}{\textemdash} & \multicolumn{1}{c}{\textemdash} & \multicolumn{1}{c}{\textemdash} & $1.63$ (290) \\
 & aggregator & 280 & $2.27$ & $5.01$ {\scriptsize[2.09, 15.4]} & $2.51$ & $4.29$ {\scriptsize[1.74, 14.8]} & \multicolumn{1}{c}{\textemdash} & \multicolumn{1}{c}{\textemdash} & \multicolumn{1}{c}{\textemdash} & $5.01$ (280) \\
\bottomrule
\end{tabular}%
}
\end{table}

The logged $\chi^{2}$ is treated as an entropy proxy because, on the probe checkpoints, its log ratio follows the log ratio of the probe entropy with $r=1.00$ on solving roles and $r=0.92$ on decision roles. The logged log training perplexity follows the probe entropy with $r=1.00$ and $0.99$, so it and the logged $\chi^{2}$ are one entropy signal. For near-deterministic decision roles the trainer and the sampler differ even at equal temperature by an amount comparable to the temperature mismatch, so the proxy does not carry over to them.

\subsection{Per-Role Gradient Measurements}
\label{sec:appendix:gradient_measurements}

These measurements support the gradient analysis of \S\ref{sec:results:ip_vs_sp}. They show how the per-sequence loss mean divides a shared policy's gradient among the roles and what that division costs the solving roles. Gradients are decomposed by role at sampled checkpoints, on rollouts that the checkpoint generates on a fixed set of training problems. Each role's gradient is accumulated over one training step, with the three instances of a parallel role summed, so a share describes one training step's gradient, not the update applied between checkpoints. Fig.~\ref{fig:ip_vs_sp}(c) comes from a second measurement on four \SPa Math cells that keeps each sequence's gradient, so the loss of the same outputs can also be averaged over all tokens.

\paragraph{Panels.} Panel~D1 of Table~\ref{tab:gradient_evidence} covers seven \SPa cells and, in its last row, \texttt{Voting-IP-1.7B-Math}, where each role has its own adapter. The projection is a role's component along the summed gradient as a fraction of that gradient's squared norm; projections sum to one over roles and net out gradient that other roles cancel. Panel~D2 divides, window by window, the rate of each \SPa adapter by the largest rate among the adapters of the matched \IPa run, over the windows that both runs' checkpoints cover.

\paragraph{Gradient shares.} In \texttt{Voting-IP-1.7B-Math} the aggregator holds a share of the summed per-role gradient comparable to that in \texttt{Voting-SP-1.7B-Math} (Panel~D1, last row), so the concentration comes from the role and the per-sequence averaging, not from sharing. Sharing determines which parameters that gradient updates. On both Code cells the two roles' lengths are closer and the shares are near uniform.

\paragraph{Reweighted shared loss.} The control of \S\ref{sec:results:ip_vs_sp} trains \texttt{Voting-SP-1.7B-Math} with each sequence's loss scaled by its role's mean response length over the overall mean, the weights normalized to a mean of one over the sequences. Paired with the shared reference, it reaches 90\% of the pair's common gain at step 136, against 281 for the shared reference; the isolated policies reach 90\% of their pair's common gain at step 109. Graded on their own answers, its voters lead the shared reference's by 4.1 points [2.6, 5.5] early and by 2.1 [0.6, 3.5] at the two runs' peaks. The isolated voters lead the shared reference's by 7.3 and 4.9 at these two points, so the reweighting closes 56\% of the voter gap early and 43\% at the peak. The early comparison uses step 120 for the shared runs and step 130 for the isolated run.

\paragraph{Decomposition of the peak gap.} Table~\ref{tab:ip_sp_decomposition} splits each Math pair's \IPa{}$-$\SPa gap at the dumps nearest the selected peaks into a solving-role part, the difference in mean solving-role accuracy graded on each role's own answer, and a structure part, the remainder. It also gives the solving-role lead early, at a step fixed in advance.

\paragraph{Parameter change.} The shared adapter trains on every role's sequences, so it takes more optimizer steps per training step than any \IPa adapter (\S\ref{sec:appendix:hyperparameters}). Averaged over the compared windows, it moves farther in all but one pair (Panel~D2).

\begin{table}[t]
\centering
\caption{Per-role gradient measurements and parameter change (definitions in \S\ref{sec:appendix:measurements}). Panel~D1: \SPa cells and, below the rule, \texttt{Voting-IP-1.7B-Math} for comparison. Share: smallest and largest fraction of the summed per-role gradient norm over the sampled checkpoints; uniform: the share if every role call contributed equally. Projection: the role's share along the summed gradient (\SPa only). Optimizer steps are per training step (\IPa row: parallel / decision role); var.: varies with the revision rounds per episode. Panel~D2: parameter change per training step of the shared adapter over that of the fastest adapter of the matched \IPa run; mean over windows (smallest window ratio).}
\label{tab:gradient_evidence}

\textbf{Panel D1.} Shared policy, one adapter serving every role; last row: isolated policies.
\smallskip

\resizebox{\textwidth}{!}{%
\begin{tabular}{@{}llcccccccccc@{}}
\toprule
& & \multicolumn{3}{c}{Decision role's share} & \multicolumn{4}{c}{Solving role's share} & \multicolumn{2}{c}{Mean length (tokens)} & \multicolumn{1}{c}{Per training step} \\
\cmidrule(lr){3-5}\cmidrule(lr){6-9}\cmidrule(lr){10-11}\cmidrule(l){12-12}
Cell & Decision role & range & projection & uniform & solving role & range & projection & uniform & decision & solving & optimizer steps \\
\midrule
\texttt{Voting-SP-0.6B-Math} & aggregator & $0.542$--$0.782$ & $0.583$--$0.922$ & $0.25$ & generator & $0.218$--$0.458$ & $0.078$--$0.417$ & $0.75$ & $465$ & $2675$ & $4$ \\
\texttt{Voting-SP-1.7B-Math} & aggregator & $0.764$--$0.903$ & $0.915$--$0.991$ & $0.25$ & generator & $0.097$--$0.236$ & $0.009$--$0.085$ & $0.75$ & $163$ & $2241$ & $4$ \\
\texttt{Voting-SP-4B-Math} & aggregator & $0.909$--$0.964$ & $0.990$--$0.999$ & $0.25$ & generator & $0.036$--$0.091$ & $0.001$--$0.010$ & $0.75$ & $289$ & $1795$ & $4$ \\
\texttt{Eval-Opt-SP-1.7B-Math} & evaluator & $0.791$--$0.945$ & $0.929$--$0.995$ & $0.50$ & generator & $0.055$--$0.209$ & $0.005$--$0.071$ & $0.50$ & $1080$ & $3331$ & var. \\
\texttt{Orch-Workers-SP-4B-Math} & synthesizer & $0.299$--$0.760$ & $0.217$--$0.935$ & $0.20$ & worker & $0.048$--$0.151$ & $0.004$--$0.057$ & $0.60$ & $1021$ & $3200$ & $5$ \\
\texttt{Voting-SP-1.7B-Code} & aggregator & $0.146$--$0.333$ & $0.029$--$0.182$ & $0.25$ & generator & $0.667$--$0.854$ & $0.818$--$0.971$ & $0.75$ & $832$ & $875$ & $4$ \\
\texttt{Eval-Opt-SP-0.6B-Code} & evaluator & $0.520$--$0.620$ & $0.540$--$0.727$ & $0.50$ & generator & $0.380$--$0.480$ & $0.273$--$0.460$ & $0.50$ & $873$ & $1300$ & var. \\
\midrule
\multicolumn{12}{@{}l}{\textit{Isolated policies}} \\
\texttt{Voting-IP-1.7B-Math} & aggregator & $0.781$--$0.885$ & \textemdash & $0.25$ & generator & $0.115$--$0.219$ & \textemdash & $0.75$ & $205$ & $3140$ & $3$\,/\,$1$ \\
\bottomrule
\end{tabular}%
}

\bigskip

\textbf{Panel D2.} Parameter change per training step, shared adapter over the fastest \IPa adapter of the matched run.
\smallskip

\begin{tabular}{@{}llccc@{}}
\toprule
Workflow & Task & 0.6B & 1.7B & 4B \\
\midrule
Eval-Opt & Math & $1.13$ ($0.95$) & $1.23$ ($0.97$) & $1.32$ ($1.04$) \\
Eval-Opt & Code & $1.06$ ($0.68$) & $1.34$ ($1.20$) & $1.22$ ($1.10$) \\
Voting & Math & $1.39$ ($1.04$) & $1.27$ ($1.11$) & $1.23$ ($1.01$) \\
Voting & Code & $1.24$ ($1.04$) & $1.09$ ($1.07$) & $0.84$ ($0.79$) \\
Orch-Workers & Math & $1.64$ ($1.38$) & $1.52$ ($1.42$) & $1.55$ ($1.50$) \\
Orch-Workers & Code & $1.35$ ($1.17$) & $1.36$ ($1.29$) & $1.42$ ($1.36$) \\
\bottomrule
\end{tabular}
\end{table}

\begin{table}[t]
\centering
\small
\caption{\IPa{} minus \SPa{} on Math (points). Peak gap: at the selected peaks (Table~\ref{tab:peak_last}, Panel~B). At the peak dumps, each run's checkpoint nearest its peak (steps IP/SP): workflow gap, its solving-role part (mean voter, mean worker, Eval-Opt first attempt, each graded on its own answer) with 95\% interval, and the structure part, the difference. Early, rule fixed in advance: the solving-role lead at each run's checkpoint nearest the step at which \IPa{} reaches half the pair's common gain (Fig.~\ref{fig:ip_vs_sp}(b) definition; ties to the earlier step).}
\label{tab:ip_sp_decomposition}
\resizebox{\textwidth}{!}{%
\begin{tabular}{@{}llccccccc@{}}
\toprule
 & & & \multicolumn{4}{c}{At the peak dumps} & \multicolumn{2}{c}{Early} \\
\cmidrule(lr){4-7}\cmidrule(l){8-9}
Scale & Workflow & Peak gap & Steps & Workflow & Solving role [95\% CI] & Structure & Steps & Solving role [95\% CI] \\
\midrule
\multirow{3}{*}{0.6B} & Eval-Opt & $+10.1$ & 230/220 & $+9.5$ & $-7.4$ $[-9.3, -5.6]$ & $+16.9$ & 100/110 & $+4.2$ $[+2.0, +6.7]$ \\
 & Voting & $+0.3$ & 260/200 & $+0.6$ & $-1.0$ $[-2.3, +0.4]$ & $+1.6$ & 100/90 & $+3.3$ $[+1.9, +4.6]$ \\
 & Orch-Workers & $+4.5$ & 200/180 & $+4.6$ & $+0.4$ $[-0.9, +1.7]$ & $+4.2$ & 90/110 & $-1.9$ $[-3.2, -0.6]$ \\
\addlinespace[2pt]
\multirow{3}{*}{1.7B} & Eval-Opt & $+7.1$ & 260/290 & $+10.6$ & $-1.6$ $[-3.8, +0.8]$ & $+12.1$ & 160/150 & $+6.7$ $[+3.9, +9.2]$ \\
 & Voting & $+4.4$ & 190/300 & $+5.8$ & $+5.5$ $[+3.8, +7.2]$ & $+0.3$ & 70/120 & $+1.2$ $[-0.2, +2.6]$ \\
 & Orch-Workers & $+3.1$ & 290/300 & $+5.1$ & $+1.3$ $[-0.2, +2.8]$ & $+3.8$ & 140/210 & $+2.4$ $[+1.0, +3.9]$ \\
\addlinespace[2pt]
\multirow{3}{*}{4B} & Eval-Opt & $+9.4$ & 220/210 & $+7.2$ & $-20.8$ $[-23.4, -18.3]$ & $+27.9$ & 90/105 & $+3.9$ $[+1.3, +6.4]$ \\
 & Voting & $+8.8$ & 100/300 & $+6.7$ & $+6.1$ $[+4.5, +7.5]$ & $+0.6$ & 70/70 & $+9.4$ $[+7.9, +11.1]$ \\
 & Orch-Workers & $-1.7$ & 100/190 & $-2.4$ & $+3.8$ $[+2.4, +5.3]$ & $-6.2$ & 60/70 & $+4.1$ $[+2.5, +5.6]$ \\
\bottomrule
\end{tabular}%
}
\end{table}

\subsection{Decline Windows and Parameter Change of the Parallel Role}
\label{sec:appendix:decline_parameters}

Tables~\ref{tab:decline_windows} and~\ref{tab:decline_parameters} hold the values behind the role-output, training-metrics and controls analyses of \S\ref{sec:results:ip_decline}: where the \IPa decline loses accuracy and how the parallel role's adapter moves.

\textbf{Decline windows.} Table~\ref{tab:decline_windows} covers the \IPa Math runs of Voting and Orch-Workers whose sampled checkpoints span the validation peak. Each \SPa window ends at the sampled checkpoint nearest the end of the \IPa window and starts at the earlier checkpoint nearest the start of the \IPa window. A problem is lost when the reference checkpoint answers it correctly and the row's checkpoint does not, and gained in the reverse case.

\textbf{Training metrics.} In Table~\ref{tab:decline_parameters}, each validation drop is the peak minus the last checkpoint within its horizon. The training-success drop compares the 31 training steps centered on the full-run validation peak with the last 30. Training success is graded like validation, on the episode's final answer, so its fall reflects lost training problems, not reward exploitation. The $\chi^{2}$ and Mism.\ columns carry one entropy signal (\S\ref{sec:appendix:gradient_amplification_signatures}).

\textbf{Direction persistence.} $\Delta\cos$ is the mean cosine between an adapter's consecutive 50-step changes minus its value under the AdamW noise model at the adapter's optimizer-step count $N$ (\S\ref{sec:appendix:measurements}). It sets the parallel adapters of three of the four declining runs with a parallel role apart from every other adapter (Table~\ref{tab:decline_parameters}); this persistence is the drift named in \S\ref{sec:results:ip_decline}. Across \IPa parallel adapters it correlates with the validation drop, but it averages a few pairs of changes on checkpoints that span the decline, so it is a correlate of the decline and does not show when it begins.

\begin{table}[t]
\centering
\caption{Voters and workers at sampled checkpoints before and after the validation peak of the \IPa Math runs and over matched \SPa windows, one block per run, on all 1412 validation problems. Peak: selected peak (step). Val.: validation accuracy. Acc.: workflow accuracy on the row's problems. Inst.: mean instance accuracy. Loss, Rescue: selection losses, rescues. Trunc.: truncated instance outputs. Lost/gained: problems lost and gained relative to the reference checkpoint Ref., with the McNemar $p$.}
\label{tab:decline_windows}
\small
\setlength{\tabcolsep}{3pt}
\resizebox{\textwidth}{!}{%
\begin{tabular}{@{}lrrrrrrrrrcccc@{}}
\toprule
Run & Step & Val. & Acc. & Oracle & Inst. & Loss & Rescue & Trunc. & Ref. & \multicolumn{2}{c}{Final answer} & \multicolumn{2}{c}{Oracle} \\
\cmidrule(lr){11-12}\cmidrule(l){13-14}
 & & & & & & & & & & Lost/gained & $p$ & Lost/gained & $p$ \\
\midrule
\texttt{Voting-IP-0.6B-Math} & 200 & 25.9 & 25.2 & 31.4 & 19.9 & 6.2 & 0.0 & 62.6 &  &  &  &  &  \\
Peak 28.5 (260) & 300 & 25.8 & 26.5 & 32.2 & 19.8 & 5.7 & 0.0 & 44.3 & 200 & 98/116 & 0.25 & 104/115 & 0.50 \\
\addlinespace[3pt]
\texttt{Voting-SP-0.6B-Math} & 90 & 14.2 & 15.6 & 28.0 & 14.3 & 12.5 & 0.1 & 1.8 &  &  &  &  &  \\
Peak 28.3 (200) & 200 & 28.3 & 28.5 & 36.2 & 22.3 & 7.7 & 0.0 & 46.6 & 90 & 66/248 & $5.9{\times}10^{-26}$ & 97/212 & $5.4{\times}10^{-11}$ \\
\addlinespace[3pt]
\texttt{Voting-IP-1.7B-Math} & 130 & 47.7 & 46.5 & 57.1 & 39.3 & 10.6 & 0.0 & 21.5 &  &  &  &  &  \\
Peak 50.9 (190) & 290 & 40.9 & 39.9 & 48.0 & 33.0 & 8.1 & 0.0 & 32.0 & 130 & 217/125 & $7.4{\times}10^{-7}$ & 220/92 & $3.0{\times}10^{-13}$ \\
 & 300 & 40.7 & 39.9 & 46.3 & 32.4 & 6.4 & 0.0 & 46.6 & 130 & 225/132 & $9.8{\times}10^{-7}$ & 233/81 & $3.4{\times}10^{-18}$ \\
\addlinespace[3pt]
\texttt{Voting-SP-1.7B-Math} & 120 & 38.0 & 38.9 & 48.9 & 31.9 & 10.1 & 0.1 & 6.8 &  &  &  &  &  \\
Peak 46.5 (300) & 300 & 46.5 & 45.2 & 54.9 & 37.9 & 9.7 & 0.0 & 15.4 & 120 & 111/200 & $5.1{\times}10^{-7}$ & 92/176 & $3.2{\times}10^{-7}$ \\
\addlinespace[3pt]
\texttt{Voting-IP-4B-Math} & 70 & 65.6 & 64.9 & 72.4 & 56.2 & 7.6 & 0.1 & 15.4 &  &  &  &  &  \\
Peak 68.1 (100) & 130 & 67.5 & 68.1 & 72.9 & 58.9 & 5.0 & 0.1 & 21.3 & 70 & 94/138 & 0.005 & 78/86 & 0.58 \\
 & 200 & 58.0 & 57.5 & 64.8 & 48.6 & 7.3 & 0.0 & 19.1 & 130 & 219/70 & $4.9{\times}10^{-19}$ & 185/70 & $3.7{\times}10^{-13}$ \\
 & 195 & 59.8 & 59.8 & 65.9 & 49.5 & 6.1 & 0.1 & 21.6 &  &  &  &  &  \\
 & 215 & 54.9 & 54.9 & 60.6 & 44.5 & 5.8 & 0.1 & 29.4 & 195 & 152/82 & $5.6{\times}10^{-6}$ & 134/60 & $1.1{\times}10^{-7}$ \\
\addlinespace[3pt]
\texttt{Voting-SP-4B-Math} & 140 & 56.2 & 55.8 & 65.0 & 48.3 & 9.5 & 0.3 & 4.8 &  &  &  &  &  \\
Peak 59.4 (300) & 170 & 55.4 & 55.2 & 65.4 & 48.8 & 10.2 & 0.0 & 5.5 & 140 & 140/132 & 0.67 & 109/115 & 0.74 \\
\addlinespace[3pt]
\texttt{Orch-Workers-IP-1.7B-Math} & 280 & 55.2 & 55.5 & 54.9 & 39.8 & 4.2 & 4.7 & 41.7 &  &  &  &  &  \\
Peak 55.5 (290) & 330 & 53.7 & 52.9 & 40.3 & 29.2 & 1.1 & 13.7 & 63.7 & 280 & 157/121 & 0.04 & 242/36 & $1.2{\times}10^{-38}$ \\
\addlinespace[3pt]
\texttt{Orch-Workers-SP-1.7B-Math} & 300 & 52.3 & 51.0 & 53.2 & 35.8 & 5.7 & 3.5 & 40.3 &  &  &  &  &  \\
Peak 52.3 (300) & 330 & 52.7 & 53.0 & 52.7 & 36.1 & 4.4 & 4.7 & 44.5 & 300 & 114/143 & 0.08 & 116/109 & 0.69 \\
\addlinespace[3pt]
\texttt{Orch-Workers-IP-4B-Math} & 60 & 60.6 & 61.4 & 69.1 & 50.9 & 8.1 & 0.5 & 8.2 &  &  &  &  &  \\
Peak 67.9 (100) & 120 & 66.4 & 66.4 & 69.3 & 54.2 & 5.4 & 2.5 & 29.4 & 60 & 92/163 & $1.0{\times}10^{-5}$ & 96/100 & 0.83 \\
 & 180 & 60.0 & 61.3 & 46.9 & 34.4 & 1.2 & 15.6 & 62.3 & 120 & 162/89 & $4.8{\times}10^{-6}$ & 335/18 & $8.4{\times}10^{-77}$ \\
\addlinespace[3pt]
\texttt{Orch-Workers-SP-4B-Math} & 130 & 63.5 & 64.2 & 70.7 & 52.3 & 7.1 & 0.6 & 13.0 &  &  &  &  &  \\
Peak 69.6 (190) & 170 & 67.8 & 67.0 & 70.3 & 52.2 & 5.5 & 2.2 & 25.0 & 130 & 95/134 & 0.01 & 106/101 & 0.78 \\
\bottomrule
\end{tabular}%
}
\end{table}

\begin{table}[t]
\centering
\caption{Parameter change and diagnostics of the \IPa parallel-role adapter (Voting generator, Orch-Workers worker), the matched \SPa adapter and the SA-RL generators; $N$, Rate and the noise model: \S\ref{sec:appendix:measurements}. Pred.: noise-model rate. vs.\ single: Rate over the mean rate of the run's single-instance adapters. $\Delta\cos$: excess cosine of consecutive 50-step changes over the noise model (pairs in parentheses). $\chi^{2}$, Mism.: late ratios of the adapter's $\chi^{2}$ and of the pre-update temperature mismatch (an entropy proxy, one signal with $\chi^{2}$); drop columns: \S\ref{sec:appendix:decline_parameters}.}
\label{tab:decline_parameters}
\small
\setlength{\tabcolsep}{3pt}
\resizebox{\textwidth}{!}{%
\begin{tabular}{@{}llrrrrrrrrrrr@{}}
\toprule
 & & & & & & & & & & \multicolumn{3}{c}{Drop after peak (pp)} \\
\cmidrule(l){11-13}
Setting & Adapter & $N$ & Rate & Pred. & Obs./pred. & vs.\ single & $\Delta\cos$ & $\chi^{2}$ & Mism. & Val.\ $\leq 300$ & Val.\ full & Train. \\
\midrule
\multicolumn{13}{@{}l}{\textit{Math}} \\
Voting 0.6B & \IPa{} parallel & 3 & 0.191 & 0.236 & 0.81 & 1.22 & $+0.11$ (4) & 5.1 & 2.3 & 2.8 & 2.8 & $+1.3$ \\
 & \SPa{} shared & 4 & 0.274 & 0.276 & 0.99 & 1.88 & $+0.03$ (2) & 1.8 & 1.5 & 0.0 & 0.0 & $+1.5$ \\
\addlinespace[2pt]
Voting 1.7B & \IPa{} parallel & 3 & 0.212 & 0.236 & 0.90 & 2.19 & $+0.26$ (4) & 22.4 & 3.5 & 10.2 & 10.2 & $+15.0$ \\
 & \SPa{} shared & 4 & 0.268 & 0.276 & 0.97 & 2.77 & $+0.01$ (4) & 2.7 & 1.9 & 0.0 & 0.9 & $+0.2$ \\
\addlinespace[2pt]
Voting 4B & \IPa{} parallel & 3 & 0.189 & 0.236 & 0.80 & 1.66 & $+0.24$ (2) & 11.1 & 3.1 & 13.2 & 13.2 & $+9.4$ \\
 & \SPa{} shared & 4 & 0.229 & 0.276 & 0.83 & 2.00 & $+0.02$ (1) & 2.8 & 1.8 & 0.0 & 0.0 & $+1.5$ \\
\addlinespace[2pt]
Orch-Workers 0.6B & \IPa{} parallel & 3 & 0.201 & 0.236 & 0.85 & 1.33 & $+0.10$ (2) & 3.4 & 1.9 & 2.5 & 2.5 & $+1.1$ \\
 & \SPa{} shared & 5 & 0.345 & 0.310 & 1.11 & 2.25 & $+0.04$ (1) & 2.6 & 1.9 & 0.0 & 0.0 & $+0.3$ \\
\addlinespace[2pt]
Orch-Workers 1.7B & \IPa{} parallel & 3 & 0.205 & 0.236 & 0.87 & 1.34 & $+0.12$ (4) & 12.9 & 3.2 & 1.2 & 4.5 & $+4.2$ \\
 & \SPa{} shared & 5 & 0.312 & 0.310 & 1.01 & 2.05 & $+0.04$ (4) & 2.4 & 1.6 & 0.0 & 0.1 & $0.0$ \\
\addlinespace[2pt]
Orch-Workers 4B & \IPa{} parallel & 3 & 0.198 & 0.236 & 0.84 & 1.63 & $+0.25$ (1) & 9.7 & 2.9 & 7.9 & 7.9 & $+6.0$ \\
 & \SPa{} shared & 5 & 0.315 & 0.310 & 1.02 & 2.56 & $+0.08$ (2) & 6.7 & 2.9 & 2.6 & 2.6 & $+0.5$ \\
\midrule
\multicolumn{13}{@{}l}{\textit{Code}} \\
Voting 0.6B & \IPa{} parallel & 3 & 0.217 & 0.236 & 0.92 & 1.52 & $0.00$ (4) & 1.9 & 1.7 & 0.0 & 0.5 & $+0.2$ \\
 & \SPa{} shared & 4 & 0.270 & 0.276 & 0.98 & 1.89 & $+0.02$ (4) & 1.6 & 1.6 & $-1.2$ & $-0.3$ & $0.0$ \\
\addlinespace[2pt]
Voting 1.7B & \IPa{} parallel & 3 & 0.243 & 0.236 & 1.03 & 1.29 & -- & 0.7 & 0.9 & 0.0 & 0.0 & $-0.5$ \\
 & \SPa{} shared & 4 & 0.283 & 0.276 & 1.03 & 1.38 & -- & 0.5 & -- & $-1.3$ & $-1.3$ & $+0.4$ \\
\addlinespace[2pt]
Voting 4B & \IPa{} parallel & 3 & 0.214 & 0.236 & 0.91 & 4.30 & $+0.02$ (2) & 1.0 & 1.2 & $-1.2$ & $-1.2$ & $+1.6$ \\
 & \SPa{} shared & 4 & 0.192 & 0.276 & 0.70 & 3.51 & -- & 1.2 & 1.2 & 0.0 & 0.0 & $-0.4$ \\
\addlinespace[2pt]
Orch-Workers 0.6B & \IPa{} parallel & 3 & 0.214 & 0.236 & 0.91 & 1.41 & $+0.01$ (3) & 2.1 & 1.6 & 0.0 & 0.0 & $+0.7$ \\
 & \SPa{} shared & 5 & 0.301 & 0.310 & 0.97 & 1.97 & $-0.01$ (3) & 1.8 & 1.7 & 0.9 & 0.9 & $+0.1$ \\
\addlinespace[2pt]
Orch-Workers 1.7B & \IPa{} parallel & 3 & 0.245 & 0.236 & 1.04 & 1.91 & $+0.01$ (4) & 0.6 & 0.9 & 1.7 & 0.5 & $-0.9$ \\
 & \SPa{} shared & 5 & 0.323 & 0.310 & 1.04 & 2.56 & $+0.02$ (1) & 0.5 & 0.6 & 0.1 & 0.1 & $+0.9$ \\
\addlinespace[2pt]
Orch-Workers 4B & \IPa{} parallel & 3 & 0.234 & 0.236 & 0.99 & 1.75 & $-0.01$ (2) & 0.6 & 1.0 & $-0.2$ & $-0.2$ & $+1.1$ \\
 & \SPa{} shared & 5 & 0.327 & 0.310 & 1.06 & 2.56 & $0.00$ (1) & 0.6 & 0.6 & $-0.7$ & $-0.7$ & $-3.4$ \\
\midrule
\multicolumn{13}{@{}l}{\textit{Single-agent RL}} \\
Math 0.6B & generator & 1 & 0.110 & 0.128 & 0.86 & -- & $+0.06$ (4) & 1.5 & 1.4 & 1.3 & 1.3 & $0.0$ \\
Math 1.7B & generator & 1 & 0.110 & 0.128 & 0.86 & -- & $+0.16$ (4) & 1.3 & 1.4 & 0.1 & 0.5 & $-0.1$ \\
Math 4B & generator & 1 & 0.099 & 0.128 & 0.77 & -- & $+0.01$ (4) & 1.0 & 1.1 & 0.0 & 0.0 & $+1.4$ \\
Code 0.6B & generator & 1 & 0.135 & 0.128 & 1.06 & -- & $+0.08$ (4) & 1.1 & 1.3 & $-0.1$ & $-0.9$ & $+0.2$ \\
Code 1.7B & generator & 1 & 0.113 & 0.128 & 0.88 & -- & $+0.03$ (4) & 1.9 & 1.7 & 0.3 & 0.3 & $+0.3$ \\
Code 4B & generator & 1 & 0.112 & 0.128 & 0.88 & -- & $+0.04$ (4) & 2.7 & 2.3 & $-1.1$ & $-1.1$ & $+0.2$ \\
\bottomrule
\end{tabular}%
}
\end{table}

\subsection{Control Runs}
\label{sec:appendix:intervention}

Table~\ref{tab:controls} gives the controls of \S\ref{sec:results:landscape}, \S\ref{sec:results:ip_vs_sp} and \S\ref{sec:results:ip_decline}, and Table~\ref{tab:appendix:intervention_controls} their validation series. Each control changes the reference, \texttt{Voting-IP-1.7B-Math}, in one or more of these parts: the reward, the workflow, the policy sharing or the number of optimizer steps per training step. Together they test which of these parts the \IPa decline requires. Every control is trained once. The lr column of Table~\ref{tab:controls} is the first training step at which every logged learning rate equals the constant rate, and every warmup ends before the setting's peak.

\textbf{Per-voter reward.} This control differs from the reference only in its reward: it scores each generator instance on its own answer and the aggregator on the final answer.

\textbf{Single role.} This control removes the Voting workflow and the aggregator. Each problem receives three groups of eight samples from the generator prompt, each scored on its own answer and all trained on one adapter, which thus takes as many optimizer steps per training step on as many samples per problem as the reference's parallel generator adapter. It is validated as a single agent.

\textbf{Shared policy and reweighted shared policy.} The shared reference, \texttt{Voting-SP-1.7B-Math}, updates the generator's parameters with the aggregator's gradient too and does not decline over its validated horizon. The reweighted shared policy scales each sequence's loss by its role's length (\S\ref{sec:appendix:gradient_measurements}). From its peak to its last checkpoint its voters' truncation rate rises from 24.9\% to 55.7\% and their mean length from 3153 to 4081 tokens (1412 problems), as in the declining isolated runs. The reweighted shared policy declines in every held-out split.

\textbf{One generator step per training step.} The merged-step run is the reference with each mini-batch holding all of a training step's sequences. Its generator adapter therefore takes one optimizer step per training step on the reference's 24 samples per problem, and its aggregator adapter takes one, as in the reference. Over its validated horizon, which is longer than the reference's, the run does not decline (Table~\ref{tab:controls}). The control thus shows that the per-instance steps bring the decline forward; it is interpreted as delaying the decline, not as removing it.

\textbf{Single-agent RL.} Every single-agent RL run ends close to its peak over its full validated horizon (Table~\ref{tab:appendix:intervention_controls}, Panel~B).

\begin{table}[t]
\centering
\footnotesize
\caption{Controls on Voting at 1.7B on Math, and single-agent RL on Math. $N$: optimizer steps of the generator's adapter per training step ($8N$ samples per problem); lr: first step at the constant learning rate. Peak: highest validation accuracy (\%; step); change: to the last checkpoint, in points (paired exact McNemar: the four falls $p<10^{-8}$, the one-step run $p=0.06$, the others $p>0.3$). $\chi^{2}$: late ratio (\S\ref{sec:appendix:measurements}). $^{a}$Aggregator on the final answer. $^{b}$24 samples per problem in one step. $^{c}$Logged $\chi^{2}$ not comparable; late training-perplexity ratio: reweighted shared policy 7.9$^{e}$, one-step run 1.5 (reference 12.9; shared policy 1.3$^{e}$). $^{d}$Peak at the last checkpoint. $^{e}$Pooled over roles.}
\label{tab:controls}
\setlength{\tabcolsep}{3pt}
\resizebox{\textwidth}{!}{%
\begin{tabular}{@{}lccccccc@{}}
\toprule
Setting & Reward & $N$ & \shortstack[c]{Adapter opt.\ steps\\at peak / last} & \shortstack[c]{lr\\from} & \shortstack[c]{Peak\\(step)} & \shortstack[c]{Change\\after peak} & \shortstack[c]{Generator $\chi^{2}$\\(entropy proxy, logged)} \\
\midrule
Reference (\texttt{Voting-IP-1.7B-Math}) & broadcast & $3$ & 570 / 900 & 1 & $50.9$ (190) & $-10.2$ & $22.4$ \\
Per-voter reward & own answer$^{a}$ & $3$ & 570 / 825 & 1 & $56.0$ (190) & $-8.3$ & $20.7$ \\
Single role: no workflow, no aggregator & own answer & $3$ & 510 / 900 & 16 & $50.2$ (170) & $-7.4$ & $17.7$ \\
One generator step per training step$^{b}$ & broadcast & $1$ & 470 / 500 & 1 & $60.9$ (470) & $-2.0$ & \textemdash$^{c}$ \\
Shared policy (\texttt{Voting-SP-1.7B-Math}) & broadcast & $4$ & 1600 / 1800 & 1 & $48.6$ (400) & $-0.9$ & $2.7^{e}$ \\
Shared policy, loss reweighted by role length & broadcast & $4$ & 760 / 1200 & 1 & $49.2$ (190) & $-14.5$ & \textemdash$^{c}$ \\
\addlinespace[2pt]
SA-RL, 0.6B & own answer & $1$ & 290 / 300 & 1 & $31.4$ (290) & $-1.3$ & $1.5$ \\
SA-RL, 1.7B & own answer & $1$ & 390 / 430 & 1 & $53.9$ (390) & $-0.5$ & $1.3$ \\
SA-RL, 4B & own answer & $1$ & 300 / 300 & 1 & $66.6$ (300) & none$^{d}$ & $1.0$ \\
\bottomrule
\end{tabular}%
}
\end{table}

\begin{table}[t]
\centering
\caption{Validation series of the settings in Table~\ref{tab:controls}. Panel~A: validation accuracy (\%) at every validated training step; --: not validated. Panel~B: the single-agent RL runs of Table~\ref{tab:main_results} over their full validated horizon; change is last minus peak.}
\label{tab:appendix:intervention_controls}
\small
\renewcommand{\arraystretch}{0.9}

\textbf{Panel A.} \texttt{Voting-IP-1.7B-Math} and its controls; \texttt{Voting-SP-1.7B-Math} and its control with the loss reweighted by role length.
\smallskip

{\scriptsize
\setlength{\tabcolsep}{3pt}
\resizebox{\textwidth}{!}{%
\begin{tabular}{@{}rcccccc|rcccccc@{}}
\toprule
 &  & Per-voter & Single & One generator & Shared & Shared policy, & &  & Per-voter & Single & One generator & Shared & Shared policy, \\
Step & Reference & reward & role & step & policy & reweighted & Step & Reference & reward & role & step & policy & reweighted \\
\midrule
10 & 33.1 & -- & -- & 30.3 & 32.9 & 32.5 & 300 & 40.7 & -- & 42.8 & 54.2 & 46.5 & 34.7 \\
20 & 33.1 & -- & -- & 31.8 & 34.1 & 33.2 & 310 & -- & -- & -- & 54.5 & 44.6 & -- \\
30 & 34.7 & 40.2 & 32.1 & 33.6 & 34.8 & 36.0 & 320 & -- & -- & -- & 53.7 & 45.2 & -- \\
40 & 35.4 & -- & -- & 34.8 & 34.3 & 35.3 & 330 & -- & -- & -- & 54.7 & 45.1 & -- \\
50 & 36.2 & -- & -- & 36.8 & 33.6 & 36.4 & 340 & -- & -- & -- & 54.4 & 47.8 & -- \\
60 & 39.2 & 46.2 & 38.6 & 37.0 & 36.5 & 37.1 & 345 & -- & -- & -- & 55.9 & -- & -- \\
70 & 40.2 & -- & -- & 36.9 & 36.0 & 36.7 & 350 & -- & -- & -- & 56.8 & 46.2 & -- \\
80 & 41.4 & -- & -- & 39.7 & 33.6 & 37.2 & 355 & -- & -- & -- & 57.2 & -- & -- \\
90 & 42.7 & 48.9 & 43.1 & 38.7 & 35.6 & 39.0 & 360 & -- & -- & -- & 56.4 & 45.9 & -- \\
100 & 42.9 & -- & 45.4 & 41.9 & 34.9 & 39.9 & 365 & -- & -- & -- & 56.8 & -- & -- \\
110 & 44.5 & -- & 44.6 & 44.3 & 36.3 & 41.4 & 370 & -- & -- & -- & 56.9 & 46.7 & -- \\
120 & 46.1 & 52.5 & 46.2 & 44.8 & 38.0 & 43.4 & 375 & -- & -- & -- & 57.7 & -- & -- \\
130 & 47.7 & 51.7 & 47.7 & 45.5 & 38.5 & 44.1 & 380 & -- & -- & -- & 57.5 & 46.2 & -- \\
140 & 47.2 & 53.5 & 46.9 & 49.6 & 38.3 & 44.8 & 385 & -- & -- & -- & 56.1 & -- & -- \\
150 & 50.2 & 53.0 & 49.7 & 49.3 & 39.2 & 44.5 & 388 & -- & -- & -- & 58.0 & -- & -- \\
160 & 49.4 & 54.6 & 49.8 & 50.3 & 40.1 & 46.6 & 390 & -- & -- & -- & 56.0 & 47.1 & -- \\
170 & 50.3 & 55.8 & 50.2 & 50.8 & 40.4 & 47.7 & 400 & -- & -- & -- & 56.3 & 48.6 & -- \\
180 & 50.6 & 54.0 & 48.9 & 50.0 & 40.6 & 48.7 & 410 & -- & -- & -- & 58.1 & 47.9 & -- \\
190 & 50.9 & 56.0 & 49.5 & 51.1 & 41.1 & 49.2 & 415 & -- & -- & -- & -- & 48.2 & -- \\
200 & 50.2 & 53.4 & 49.6 & 49.7 & 42.4 & 48.5 & 420 & -- & -- & -- & 57.8 & 48.4 & -- \\
210 & 49.6 & 53.5 & 46.4 & 48.6 & 42.4 & 46.3 & 425 & -- & -- & -- & -- & 46.9 & -- \\
220 & 48.3 & 51.6 & 45.0 & 48.3 & 42.7 & 44.8 & 430 & -- & -- & -- & 57.8 & 47.9 & -- \\
230 & 47.9 & 51.0 & 45.0 & 51.4 & 43.8 & 43.4 & 440 & -- & -- & -- & 58.8 & 48.2 & -- \\
240 & 46.1 & 48.3 & 44.3 & 52.8 & 42.4 & 42.6 & 445 & -- & -- & -- & -- & 46.8 & -- \\
250 & 47.0 & 50.4 & 45.3 & 52.0 & 44.8 & 44.0 & 450 & -- & -- & -- & 58.1 & 47.7 & -- \\
260 & 43.0 & 49.6 & 44.3 & 52.3 & 43.6 & 41.1 & 460 & -- & -- & -- & 58.3 & -- & -- \\
270 & 44.2 & 48.5 & 42.6 & 51.2 & 43.1 & 37.8 & 470 & -- & -- & -- & 60.9 & -- & -- \\
275 & -- & 47.7 & -- & -- & -- & -- & 480 & -- & -- & -- & 58.3 & -- & -- \\
280 & 42.7 & -- & 43.1 & 52.5 & 44.5 & 36.5 & 490 & -- & -- & -- & 58.9 & -- & -- \\
290 & 40.9 & -- & 42.7 & 52.2 & 45.0 & 35.3 & 500 & -- & -- & -- & 58.9 & -- & -- \\
\bottomrule
\end{tabular}%
}
}

\medskip

\textbf{Panel B.} Single-agent RL, full validated horizon.
\smallskip

\begin{tabular}{@{}llccc@{}}
\toprule
Task & Scale & Peak (step) & Last (step) & Change \\
\midrule
Math & 0.6B & 31.4 (290) & 30.1 (300) & $-1.3$ \\
 & 1.7B & 53.9 (390) & 53.4 (430) & $-0.5$ \\
 & 4B & 66.6 (300) & 66.6 (300) & $0.0$ \\
\addlinespace[2pt]
Code & 0.6B & 13.8 (430) & 13.0 (501) & $-0.8$ \\
 & 1.7B & 19.4 (290) & 18.2 (301) & $-1.2$ \\
 & 4B & 34.2 (280) & 32.9 (301) & $-1.3$ \\
\bottomrule
\end{tabular}
\end{table}

\section{Relation to AT-GRPO and Dr.~MAS}
\label{sec:appendix:prior_methods}

AT-GRPO~\citep{zhao2025stronger} and Dr.~MAS~\citep{feng2026drmas} each propose a training rule for multi-agent LLM systems and evaluate it on their own workflows. This paper holds one multi-agent GRPO training rule fixed and studies how its per-role updates behave across a grid of workflow topology, task, model scale, and policy sharing. Table~\ref{tab:prior_methods} compares the three training rules component by component. Our advantages use the agent-wise normalization of Dr.~MAS (\S\ref{sec:appendix:rewards}).

\begin{table}[t]
\centering
\caption{Training rules of AT-GRPO, Dr.~MAS, and this paper. Entries marked $^{\dagger}$ describe the authors' released implementation rather than the paper. A training step is one rollout-and-update cycle over a batch of problems; an optimizer step is one parameter update on one mini-batch.}
\label{tab:prior_methods}
\footnotesize
\setlength{\tabcolsep}{4pt}
\begin{tabular}{@{}p{2.3cm}p{3.55cm}p{3.55cm}p{3.55cm}@{}}
\toprule
 & AT-GRPO & Dr.~MAS & This paper \\
\midrule
Advantage group & problem, role, and turn & problem and agent, pooling all of the agent's calls & problem and role, pooling all of a role's turns; on Voting on mathematics, one group per generator instance \\
\addlinespace
Normalization statistics & group mean and standard deviation & group mean and standard deviation & group mean and standard deviation \\
\addlinespace
Reward & team reward plus a role-local reward (e.g., execution of a tool agent's code, generated unit tests) & shared outcome reward on every agent's outputs, plus an invalid-action penalty$^{\dagger}$ & shared outcome reward on every role's sequences \\
\addlinespace
Rollout sampling & per turn, several candidates from identical parent states, all trained on; the episode continues from the highest-reward candidate & independent trajectories per problem & $n=8$ independent full-workflow rollouts per problem \\
\addlinespace
Policy assignment & one shared policy, one LoRA adapter per role on a shared base, or one model per role & one shared model or a separate full model per agent & one shared LoRA adapter (\SPa) or one LoRA adapter per role type (\IPa) \\
\addlinespace
Optimizer steps per policy per training step & grows with the policy's number of sequences (mini-batch size fixed in sequences)$^{\dagger}$ & fixed (mini-batch scaled to each model's sample count; one step on mathematics)$^{\dagger}$ & grows with the role's number of sequences (mini-batch size fixed in sequences); under \IPa on Voting, three for the generator adapter and one for the aggregator adapter \\
\addlinespace
Evaluated topologies & two-agent workflows with one instance per role (games, planning, code, mathematics); no parallel same-role instances & solver and verifier (mathematics); a verifier that hands off to a search or an answer agent (search); no parallel same-role instances & Eval-Opt, Voting, Orch-Workers; three parallel same-role instances in Voting and Orch-Workers \\
\addlinespace
KL term & none$^{\dagger}$ & none$^{\dagger}$ & none \\
\bottomrule
\end{tabular}
\end{table}

\textbf{Optimizer steps per training step.} The difference that bears on the parallel role's drift under \IPa (\S\ref{sec:results:ip_decline}, \S\ref{sec:appendix:decline_parameters}) is the number of optimizer steps each policy takes per training step, a choice neither paper analyzes. With the mini-batch size fixed in sequences, as in our trainer and in AT-GRPO's released implementation, a policy's step count grows with its number of sequences, so under \IPa the Voting generator adapter takes three optimizer steps per training step and the aggregator adapter one. Dr.~MAS's released implementation instead scales each model's mini-batch to its sample count, so every model takes a fixed number of optimizer steps per training step, one on mathematics, as the parallel role does in the merged-step control of Table~\ref{tab:controls}.

\end{document}